\documentclass[times]{article}

\usepackage{graphicx}

\usepackage{amssymb}
 \usepackage{amsthm}
\usepackage{algorithm}
\usepackage{algorithmic}

\begin{document}

\title{Active Learning of Inverse Models with Intrinsically Motivated Goal Exploration in Robots
\footnote{Baranes, A., Oudeyer, P-Y. (2012) Active Learning of Inverse Models with Intrinsically Motivated Goal Exploration in Robots, \textit{Robotics and Autonomous Systems,  61(1), pp. 49-73. http://dx.doi.org/10.1016/j.robot.2012.05.008}
}
}
\author{Adrien Baranes and Pierre-Yves Oudeyer \\
INRIA and Ensta-ParisTech, France}

\maketitle

\textbf{Highlights}: \\ \\
1) SAGG-RIAC is an architecture for active learning of inverse models in high-dimensional redundant spaces \\
2) This allows a robot to learn efficiently distributions of parameterized motor policies that solve a corresponding distribution of parameterized tasks \\
3) Active sampling of parameterized tasks, called active goal exploration, can be significantly faster than direct active sampling of parameterized policies \\
4) Active developmental exploration, based on competence progress, autonomously drives the system to progressively explore tasks of increasing learning complexity. \\

\begin{abstract}
We introduce the Self-Adaptive Goal Generation - Robust Intelligent Adaptive Curiosity (SAGG-RIAC) architecture as an intrinsically motivated goal exploration mechanism which allows active learning of inverse models in high-dimensional redundant robots. This allows a robot to efficiently and actively learn distributions of parameterized motor skills/policies that solve a corresponding distribution of parameterized tasks/goals. The architecture makes the robot sample actively novel parameterized tasks in the task space, based on a measure of competence progress, each of which triggers low-level goal-directed learning of the motor policy parameters that allow to solve it. For both learning and generalization, the system leverages regression techniques which allow to infer the motor policy parameters corresponding to a given novel parameterized task, and based on the previously learnt correspondences between policy and task parameters. 

We present experiments with high-dimensional continuous sensorimotor spaces in three different robotic setups: 1) learning the  inverse kinematics in a highly-redundant robotic arm, 2) learning omnidirectional locomotion with motor primitives in a quadruped robot, 3) an arm learning to control a fishing rod with a flexible wire. We show that 1) exploration in the task space can be a lot faster than exploration in the actuator space for learning inverse models in redundant robots; 2) selecting goals maximizing \textit{competence progress} creates developmental trajectories driving the robot to progressively focus on tasks of increasing complexity and is statistically significantly more efficient than selecting tasks randomly, as well as more efficient than different standard active motor babbling methods; 3) this architecture allows the robot to actively discover which parts of its task space it can learn to reach and which part it cannot.


\end{abstract}



\section{Motor Learning and Exploration of Forward and Inverse Models}

To operate robustly and adaptively in the real world, robots need to know how to predict the consequences of their actions (called here forward models, mapping typically $X=(S,\pi_{\theta})$, where $S$ is the state of a robot and $\pi_{\theta}:S\rightarrow A$ is a parameterized action policy, to the space of effect, or task space, $Y$). Reversely, they need to be able to compute the action policies that can generate given effects (called here inverse models, $(S,Y)\rightarrow \pi_\theta$). These models can be quite varied, for example mapping joint angles to hand position in the visual field, oscillation of the legs to body translation, movement of the hand in the visual field to movement of the end point of a tool, or properties of a hand tap an object to the sound it produces. Some of these models can be analytically elaborated by an engineer and provided to a robot (e.g. forward and inverse kinematics of a rigid body robot). But in many cases, this is impossible either because the physical properties of the body itself cannot be easily modeled (e.g. compliant bodies with soft materials), or because it is impossible to anticipate all possible objects the robot might interact with, and thus the properties of objects. More generally, it is impossible to model a priori all the possible effects a robot can produce on its environment, especially when robots are targeted to interact with in everyday human environments, such as in assistive robotics. As a consequence, learning these models through experience becomes necessary. Yet, this poses highly difficult technical challenges, due in particular to the combination of the following facts: 1) these models are often high-dimensional, continuous and highly non-stationary spatially, and sometimes temporally; 2) learning examples have to be collected autonomously and incrementally by robots; 3) learning, as we will detail below, can happen either through self-experimentation or observation, and both of these takes significant physical time in the real world. Thus, the number of training examples that can be collected in a life-time is strongly limited with regards to the size and complexity of the spaces. Advanced statistical learning techniques dedicated to incremental high-dimensional regression have been elaborated recently, such as \cite{Sigaud11,Peters11}. Yet, these regression mechanisms are efficient only if the quality and quantity of data is high enough, which is not the case when using unconstrained exploration such as random exploration.  Fundamental complementary mechanisms for guiding and constraining autonomous exploration and data collection for learning are needed. 

In this article, we present a particular approach to address constrained exploration and learning of inverse models in robots,  based on an active learning process inspired by mechanisms of intrinsically motivated learning and exploration in humans. As we will explain, the approach studies the combination of two principles for learning efficiently inverse models in high-dimensional redundant continuous spaces: 
\begin{itemize}
\item \textbf{Active goal/task exploration in a parameterized task space}: The architecture makes the robot sample actively novel parameterized tasks in the task space, each of which triggers low-level goal-directed learning of the motor policy parameters that allow to solve it. This allows to leverage the redundancies of the sensorimotor mapping, leading the system to explore densely only subregions of the space of action policies that are enough to achieve all possible effects. Thus, it does not need to learn a complete forward model and contrasts with approaches that directly sample action policy parameters and observe their effects in the task space. The system also leverages regression techniques which allow to infer the motor policy parameters corresponding to a given novel parameterized task, and based on the previously learnt correspondences between policy and task parameters. 
\item \textbf{Interestingness as empirically evaluated competence progress:} The measure of interestingness for a given goal/task is based on competence progress empirically evaluated, i.e. how previous attempts of low-level optimization directed at similar goals allowed to improve the capability of the robot to reach these goals.
\end{itemize}

In the rest of the section, we review various related approaches to constraining exploration for motor learning.


\subsection{Constraining the Exploration}
A common way to carry out exploration is to use a set of constraints on guiding mechanisms and maximally reduce the size and/or dimensionality of explored spaces. Social guidance is an important source of such constraints, widely studied in robot learning by demonstration/imitation where an external human demonstrator assists the robot in its learning process \cite{Abbeel04, Lopes07, Billard08,Calinon09, Lopes09, Cederborg10, Lopes10}. Typically, a robot teacher manually interacts with the robot by showing it a few behaviors corresponding to a desired movement or goal that it will then have to reproduce. This strategy prevents the robot from performing any autonomous exploration of its space and requires an attentive demonstrator. Some other techniques allow more freedom to the human teacher and the robot by allowing the robot to explore. This is typically what happens in the reinforcement learning (RL) framework where no demonstration is originally required and only a goal has to be fixed (as a reward) by the engineer who conceives the system \cite{Sutton98, Riedmiller09, Theodorou07}. Nevertheless, when the robot evolves in high-dimensional and large spaces, the exploration still has to be constrained. For instance, studies presented in \cite{Peters03} combine RL with the framework of learning by demonstration. In their experiments, an engineer has to first define a specific goal in a task space as a handcrafted reward function, then, a human demonstrator provides a few examples of successful motor policies to reach that goal, which is then used to initialize an optimization procedure. The Shifting Setpoint Algorithm (SSA) introduced by Schaal and Atkeson \cite{Schaal94} proposes another way to constrain the exploration process. Once a goal fixed in an handcrafted manner, a progressive exploration process is proposed: the system explores the world gradually from the start position and toward the goal by creating a local model around the current position and shifting in direction of the goal once this model is reliable enough, and so on. These kinds of techniques therefore restrain the exploration to narrow tubes of data targeted at learning specific tasks/goals that have to be defined by a human, either the programmer or a non-engineer demonstrator. 

These methods are efficient and useful in many cases. Nevertheless, in a framework where one would like a robot to learn a variety of tasks inside unprepared spaces like in developmental robotics \cite{Weng01,Oudeyer07, Weng04, Asada09}, or more simply full inverse models (i.e. having a robot learn to generate in a controlled manner many effects rather than only a single goal),  it is not conceivable that a human being interacts with a robot at each instant or that an engineer designs and tunes a specific reward function for each novel task to be learned. For this reason, it is necessary to introduce mechanisms driving the learning and exploration of robots in an autonomous manner.

\subsection{Driving Autonomous Exploration}

Active learning algorithms can be considered as organized and constrained self-exploration processes \cite{Fedorov72, Cohn96, Roy01, Settles09, Lopes10}. In the regression setting, they are used to learn a regression mapping between an input space $X$ and an output space $Y$ while minimizing the sample complexity, i.e. with a minimal number of examples necessary to reach a given performance level.  These methods, typically beginning by random and sparse exploration, build meta-models of performances of the motor learning mechanisms and concurrently drive the exploration in various sub-spaces for which a notion of \textit{interest} is defined, often consisting in variants of expected informational gain. A large diversity of criteria can be used to evaluate the utility of given sampling candidates, such as the maximization of prediction errors \cite{Thrun95}, the local density of already queried points \cite{Whitehead91}, the maximization of the decrease of global model variance \cite{Cohn96}, expected improvement \cite{Jones98}, or maximal uncertainty of the model \cite{Thrun92c} among others. There have been active-extensions to most of the existent learning methods, e.g.   logistic regression \cite{Schein07}, support vector machines \cite{Schohn00}, gaussian processes \cite{Kapoor07, Krause07,Krause08}. Only very recently have these approaches been applied to robotic problems, and even more recently if we consider examples with \textit{real} robots. Nevertheless examples that consider robotic problems already exist for a large variety of problems: building environment maps \cite{Martinez09,Thrun95}, reinforcement learning \cite{Barto04}, body schema learning \cite{Lopes10b}, imitation \cite{Lopes09b,Chernova09}, exploration of objects and body properties \cite{Oudeyer07}, manipulation \cite{Hart08b}, among many others. 

Another approach to exploration came from an initially different problem, that of understanding how robots could achieve cumulative and open-ended learning autonomously. This raised the question of the task-independent mechanisms that may allow a robot to get interested in practicing skills and learn new tasks that were not specified at design time. Two communities of researchers, the first one in reinforcement learning \cite{Sutton90,Schmidhuber91,Barto04,Schmidhuber06}, the second one in developmental robotics \cite{Huang02,Blank05,Oudeyer05,Oudeyer07,Schembri07}, formalized, implemented and experimented several mechanisms based on the concept of intrinsic motivation (sometimes called curiosity-driven learning) grounded into theories of motivation, spontaneous exploration, free play and development in humans \cite{White59, Deci85, Berlyne60} as well as in recent findings in the neuroscience of motivation \cite{Schultz97, Kakade02, Redgrave06}.

As argumented in \cite{Schmidhuber91, Barto04, Baranes09,Oudeyer10b}, architectures based on intrinsically motivated learning can be conceptualized as active learning mechanisms which, in addition to allowing for the self-organized formation of behavioral and developmental complexity, can also also allow an agent to efficiently learn a model of the world by parsimoniously designing its own experiments/queries. Yet, in spite of these similarities between work in active learning and intrinsic motivation, these two strands of approaches often differ in their underlying assumptions and constraints, leading to sometimes very different active learning algorithms. In many active learning models, one often assumes that it is possible to learn a model of the complete world within lifetime, and/or that the world is learnable everywhere, and/or where noise is homogeneous everywhere. Given those assumptions, heuristics based on the exploration of parts of the space where the learned model has maximal uncertainties or where its prediction are maximally wrong are often very efficient. Yet, these assumptions typically do not hold in real world robots in an unconstrained environment: the sensorimotor spaces, including the body dynamics and its interactions with the external world, are simply much too large to be learned entirely in a life time; there are typically subspaces which are unlearnable due to inadequate learning biases or unobservable variables; noise can be strongly homogeneous. Thus, different authors claimed that typical criteria used in traditional active learning approaches, such as the search for maximal uncertainty or prediction errors, might get trapped or become inefficient in situations that are common in open-ended robotic environments \cite{Schmidhuber06,Oudeyer07,Baranes09,Schmidhuber10}. This is the reason why new active learning heuristics have been proposed in developmental robotics, such as those based on the psychological concept of intrinsic motivations \cite{Ryan00, Deci85, Berlyne66} which relate to mechanisms that drive a learning agent to perform different activities for their own sake, without requiring any external reward \cite{Schmidhuber91,Barto04, Barto06, Barto10, Marshall04, Merrick09, Schembri07b, Schmidhuber91c, Huang02, Baldassare11, Oudeyer07b, Luciw11, Fasel10}. Different criteria were elaborated, such as the search for maximal reduction in empirically evaluated prediction error, maximal compression progress, or maximal competence progress \cite{Schmidhuber91,Schmidhuber06,Oudeyer07}. For instance, the architecture called Robust-Intelligent Adaptive Curiosity (RIAC) \cite{Baranes09}, which is a refinement of the IAC architecture which was elaborated for open-ended learning of affordances and skills in real robots \cite{Oudeyer07}, defines the interestingness of a sensorimotor subspace by the velocity of the decrease of the errors made by the robot when predicting the consequences of its actions, given a context, within this subspace. As shown in \cite{Oudeyer07, Baranes09}, it biases the system to explore subspaces of progressively increasing complexity.

Nevertheless, RIAC and similar "knowledge based" approaches (see \cite{Oudeyer08}) have some limitations: first, while they can deal with the spatial or temporal non-stationarity of the model to be learned, they face the curse-of-dimensionality and can only be efficient when considering a moderate number of control dimensions (e.g. up to 9/10). Indeed, as many other active learning methods, RIAC needs a certain level of sampling density in order to extract and compare the interest of different areas of the space. Also, because performing these measure costs time, this approach becomes more and more inefficient as the dimensionality of the control space grows \cite{Bishop07}. Second, they focus on the active choice of motor commands and measures of their consequences, which allows learning forward models that can be re-used as a side effect for achieving goals/tasks through online inversion: this approach is sub-optimal in many cases since it explores in the high-dimensional space of motor commands and consider the achievement of tasks only indirectly. 

A more efficient approach consists in directly actively exploring task spaces, which are also often much lower-dimensional, by actively self-generating goals within those task spaces, and then learn associated local coupled forward/inverse models that are useful to achieve those goals. Yet, as we will see, the process is not as straightforward as learning the forward model, since because of the space redundancy it is not possible to learn directly the inverse model (and this is the reason why learning the forward model and then only inversing it has often been achieved). In fact, exploring the task space will be used to learn a sub-part of the forward model that is enough for reaching most of reachable parts in the task space through local inversion and regression, leveraging techniques for generalizing policy parameters corresponding to novel task parameters based on previously learnt correspondences, such as in \cite{Bitzer09, Baranes09, Kober-RSS-10,Barto12}.


\subsection{Driving the Exploration at a Higher Level}

In a framework where a system should be able to learn to perform a maximal amount of different tasks (here this means achieving many goals/tasks in a parameterized task space) before focusing on different ways to perform the same tasks (here this means finding several alternative actions to achieve the same goal), knowledge-based exploration techniques like RIAC cannot be efficient in robots with redundant forward models. Indeed, they typically direct a robotic system to spend copious amounts of time exploring variations of action policies that produce the same effect, at the disadvantage of exploring other actions that might produce different outcomes, useful to achieve more tasks. An example of this is learning 10 ways to push a ball forward instead of learning to push a ball in 10 different directions. One way to address this issue is to take inspiration infant's motor exploration/babbling behavior, which has been argued to be teleological via introducing goals explicitly inside a task space and driving exploration at the level of these goals \cite{Hofsten04,Hofsten94,Meer95,Meer97}. Once a goal/task is chosen, the system would then try to reach it with a lower-level goal-reaching architecture typically based on coupled inverse and forward models, which might include a lower-level goal-directed active exploration mechanism. 

Two other developmental constraints, playing an important role in infant motor development, and presented in the experimentations of this paper, can also play an important role when considering such a task-level exploration process. First, we use motor synergies which have been shown as simplifying motor learning by reducing the number of dimensions for control (nevertheless, even with motor synergies, the dimensionality of the control space can easily go over several dozens, and exploration still needs to be organized). These motor synergies are often encoded using Central Pattern Generators (CPG) \cite{Ijspeert08, Delcomyn80, Avella03, Lee84, Berniker09} or as more traditional innate low-level control loops which are part of the innate structure allowing a robot to bootstrap the learning of new skills, as for example in \cite{Oudeyer07, Hart08} where it is combined with intrinsically motivated learning. Second, we will use a heuristic inspired by observations of infants who sometimes prepare their reaching movements by starting from a same rest position \cite{Berthier99}, by resetting the robot to such a rest position, which allows reducing the set of starting states used to perform a task.

In this paper, we propose an approach which allows us to transpose some of the basic ideas of IAC and RIAC architectures, combined with ideas from the SSA algorithm, into a multi-level active learning architecture called \textbf{Self-Adaptive Goal Generation RIAC algorithm (SAGG-RIAC)} (an outline and initial evaluation of this architecture was presented in \cite{Baranes10b}). Unlike RIAC which was made for active learning of forward models mapping action policy parametes to effects in a task space, we show that this new algorithm allows for efficient learning of inverse models mapping parameters of tasks to parameters of action policies that allow to achieve these tasks in redundant robots. This is achieved through active sampling of novel parameterized tasks in the task space, based on a measure of competence progress, each of which triggers low-level goal-directed learning of the motor policy parameters that allow to solve it. This takes advantage of both the typical redundancy of the mapping and of the fact that very often the dimensionality of the task space considered is much smaller than the dimensionality of motor primitives/action parameter space. Such an architecture also leverages both techniques for optimizing action policy parameters for a single predefined tasks (e.g. \cite{Peters08c,Stulp12}), as well as  regression techniques allowing to infer the motor policy parameters corresponding to a given novel parameterized task, and based on the previously learnt correspondences between policy and task parameters (e.g. \cite{Bitzer09, Baranes09, Kober-RSS-10,Barto12}). While approaches such as \cite{Peters08c,Stulp12} or  \cite{Bitzer09, Kober-RSS-10,Barto12} do not consider the problem of autonomous life-long exploration of novel parameterized tasks, they are very complemetary to the present work as they could be used as the low-level techniques for low-level learning of action parameter policies for self-generated tasks in the SAGG-RIAC architecture.

SAGG-RIAC can be considered as an active learning algorithm carrying out the concept of \textit{competence based intrinsically motivated learning} \cite{Oudeyer08} and is in line with concepts of mastery motivation, \textit{Flow}, \textit{Optimal Level theories} and \textit{zone of Proximal Development} introduced in psychology \cite{Dichter97, Csik96, Redding88, Arkes82, Vygotsky78}. 
In a \textit{competence based} active exploration mechanism, according to the definition \cite{Oudeyer08}, the robot is pushed to perform an active exploration in the goal/operational space as opposed to motor babbling in the actuator space. 

Several strands of previous research have began exploration various aspects of this family of mechanisms. First, algorithms achieving competence based exploration and allowing general computer programs to actively and adaptively self-generate abstract computational problems, or goals, of increasing complexity were studied in a theoretical computer science perspective \cite{Schmidhuber99, Schmidhuber02, Schmidhuber11}. While the high expressivity of these formalisms allows in principle to tackle a wide diversity of problems, they were not designed nor experimented for the particular family of problems of learning high-dimensional continuous models in robotics. While SAGG-RIAC also actively and adaptively self-generates goals, this is achieved with a formalism based on applied mathematics and dedicated to the problem of learning inverse models in continuous redundant spaces. 

Measures of interestingness based on a measure of  \textit{competence} to perform a skill were studied in \cite{Bakker04}, as well as in \cite{Schembri07b} where a \textit{selector} chooses to perform different skills depending on the temporal difference error to reach each skill. The study proposed in \cite{Stout10} is based on the \textit{competence progress}, which they use to select goals in a pre-specified set of skills considered in a discrete world. As we will show, SAGG-RIAC also uses competence progress, but targets learning in high-dimensional continuous robot spaces.

A mechanism for passive exploration in the task space for learning inverse models in high-dimensional continuous robotics spaces was presented in \cite{Rolf10a, Rolf11}, where a robot has to learn its arm inverse kinematics while trying to reach in a preset order goals put on a pre-specified grid informing the robot about the limits of its reachable space. In SAGG-RIAC  exploration is actively driven in the task space, allowing the learning process to minimize its sample complexity, and as we will show, to reach a high-level of performances in generalization and to discover automatically its own limits of reachability. 

%



%

In the following sections we introduce the global architecture and formalization of the Self-Adaptive Goal-Generation SAGG-RIAC architecture. Then, we study experimentally its capabilities to allow a robot efficiently and actively learn distributions of parameterized motor skills/policies that solve a corresponding distribution of parameterized tasks/goals, and in the context of three experimental setups: 1) learning the  inverse kinematics in a highly-redundant robotic arm, 2) learning omnidirectional locomotion with motor primitives in a quadruped robot, 3) an arm learning to control a fishing rod with a flexible wire. More precisely, we focus on the following aspects and contributions of the architecture:
\begin{itemize}
\item SAGG-RIAC creates developmental trajectories driving the robot to progressively focus on tasks of increasing complexity of learnability;
\item Drives the learning of a high variety of parameterized tasks (i.e. capability to reach various regions of the goal/task space) instead of numerous ways to perform the same task;
\item Allows learning fields of tasks in high-dimensional high-volume control spaces as long as the task space is low-dimensional (it can be high-volume);
\item Allows learning in task-spaces where only small and initially unknown subparts are reachable;
\item Drives the learning of inverse models of highly-redundant robots with different body schemas;
\item Guides the self-discovery of the limits of what the robot can achieve in its task space;
\item Allows improving significantly the quality of learned inverse models in terms of speed of learning and generalization performance to reach goals in the task space, compared to different methods proposed in the literature;
\end{itemize}

%
%
%
%



\section{Competence Based Intrinsic Motivation: The Self-Adaptive Goal Generation RIAC Architecture}

\subsection{Global Architecture}
\begin{figure}[h!]
\centerline{\includegraphics[width=\linewidth]{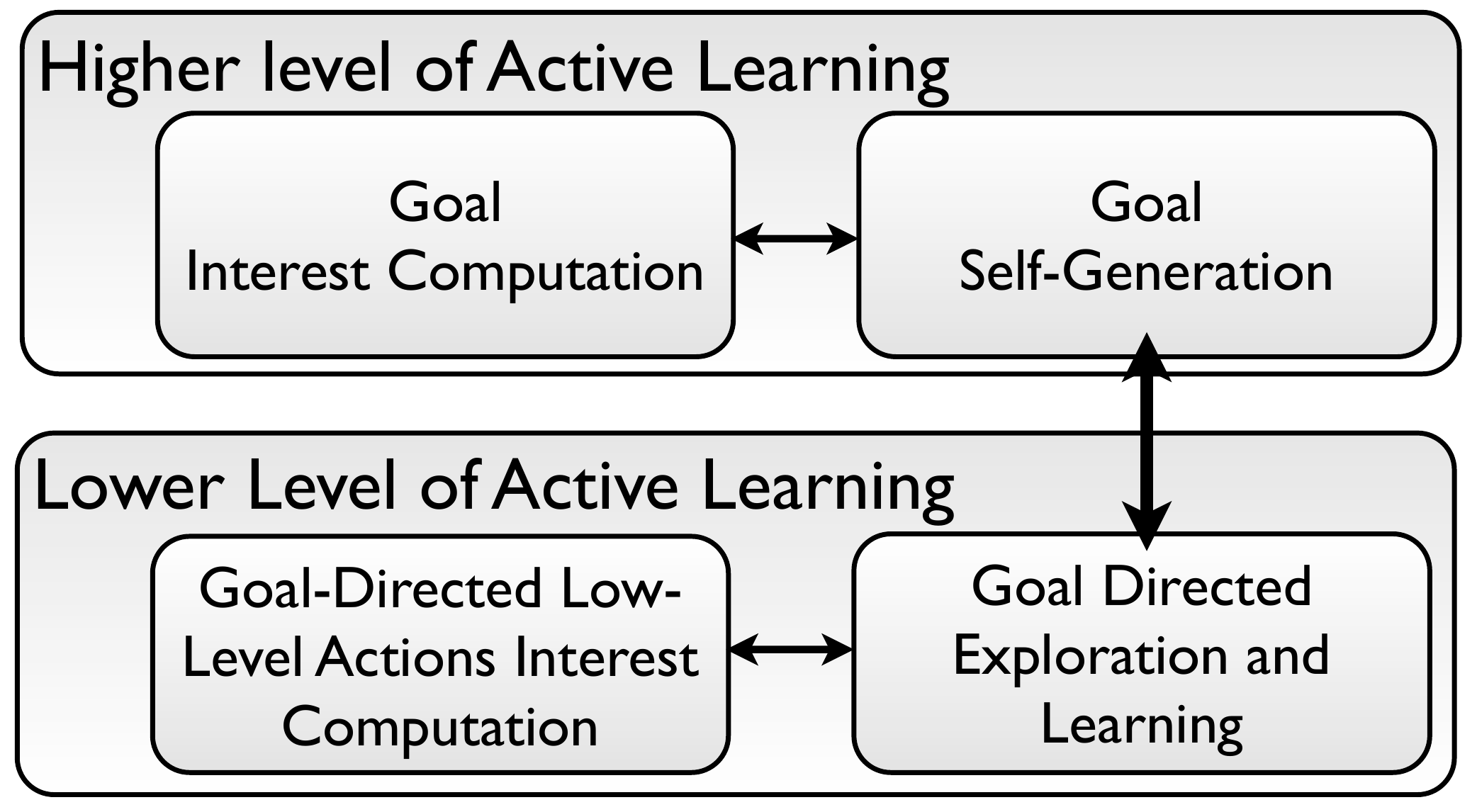}}
\caption{Global Architecture of the SAGG-RIAC architecture. The structure is comprised of two parts defining two levels of active learning: a higher level which considers the active self-generation and self-selection of goals, and a lower level, which considers the goal-directed active choice and active exploration of low-level actions, in order to reach the goals selected at the higher level.}
\label{Global}
\end{figure}

Let us consider the definition of competence based models outlined in \cite{Oudeyer08}, and extract from it two different levels for active learning defined at different time scales (Fig. \ref{Global}):
\begin{enumerate}
\item The higher level of active learning (higher time scale) takes care of the \textit{active self-generation and self-selection of goals/tasks} in a parameterized task space, depending on a measure of interest based on the level of competences to reach previously generated goals (e.g. competence progress);
\item The lower level of active learning (lower time scale) considers the \textit{goal-directed active choice and active exploration} of lower-level actions to be taken to reach the goals selected at the higher level, and depending on local measures of interest related to the evolution of the quality of learned inverse and/or forward models;
\end{enumerate}

\subsection{Model Formalization}
\label{formalization}
Let us consider a robotic system described in both a state/context space $S$, and a task space $Y$ which is a field of parameterized tasks/goals that can be viewed as defining a field of parameterized reinforcement learning problems. For a given context $s \in S$, a sequence of actions $a = \{a_1, a_2, ..., a_n\} \in A$, potentially generated by a parameterized motor synergy $\pi_{\theta}:S\rightarrow A$ (alternatively called an ``option'' and including a self-termination mechanism), allows a transition toward the new states $y \in Y$ such that $(s, a) \rightarrow y$, also written $(s,\pi_{\theta}) \rightarrow y$. For instance, in the first experiment introduced in the following sections where we use a robotic manipulator, $S$ represents its actuator/joint space, $Y$ the operational space corresponding to the cartesian position of its end-effector, and $A$ relates to velocity commands in the joints. Also, in the second experiment involving a quadruped where we use motor synergies, the context $s$ is always reset to a same state and has thus no influence on the learning, $A$ relates to the 24 dimensional parameters of a motor synergy which considers the frequency and amplitude of sinusoids controlling the position of each joints over time, and $Y$ relates to the position and orientation of the robot after the execution of the synergy during a fixed amount of time.


SAGG-RIAC drives the exploration and learning of how to reach goals given starting contexts/states. Starting states are formalized as configurations $s \in S$ and goals as a desired $y_g \in Y$. All states are considered to be potential starting states; therefore, once a goal has been generated, the low-level goal directed exploration and learning mechanism always tries to reach it by starting from the current state of the system as formalized and explained below.

When the initiation position $s_{start}$, the goal $y_g$ and constraints $\rho$ (e.g. linked with the spent energy) are chosen, it generates a motor policy $\pi_{\theta(Data)} {(s_{start}, y_g, \rho)}$ parameterized by $s_{start}$, $y_g$ and $\rho$ as well as parameters $\theta$ of internal forward and inverse models already learned with previously acquired data $\mathit{Data}$. Also, it is important to notice that  $\pi_{\theta(Data)} {(s_{start}, y_g, \rho)}$ can be computed on the fly, as in the experiments below, with regression techniques allowing to infer the motor policy parameters corresponding to a given novel parameterized task, and based on the previously learnt correspondences between policy and task parameters, such as in \cite{Bitzer09, Baranes09, Kober-RSS-10,Barto12}. 


We can make an analogy of this formalization with the \textit{Semi-Markov Option} framework introduced by Sutton \cite{Sutton99}. In the case of SAGG-RIAC, when considering an option $\langle I, \pi, \beta \rangle$, we can first define the initiation set ${I} : S \rightarrow [0;1]$, where ${I}$ is true everywhere, because, as presented before, every state can here be considered as a starting state. Also, goals are related to the terminal condition $\beta$ and $\beta = 1$ when the goal is reached, and the policy $\pi$ encodes the skill learned through the process induced by the lower-level of active learning and shall be indexed by the goal $y_g$, i.e. $\pi_{y_g}$. More formally, as induced by the use of semi-markov options, we define policies and termination conditions as dependent on all events between the initiation of the option, and the current instant. This means that the policy $\pi$, and $\beta$ are depending on the \textit{history} $h_{t\tau} = \{s_t, a_t, s_{t+1}, a_{t+1}..., s_\tau \}$ where $t$ is the initiation time of the option, and $\tau$, the time of the latest event. Denoting the set of all histories by $\Omega$, the policy and termination condition become defined by $\pi: \Omega \times {A} \rightarrow [0;1]$ and $\beta : \Omega \rightarrow [0;1]$. 

Moreover, because we have to consider cases where goals are not reachable (either because of physical impossibility or because the robot is not capable of doing it at that point of its development), we need to define a \textit{timeout} $t_{max}$ which can stop a goal reaching attempt once a maximal number of actions has been executed. $h_{t\tau}$ is thus needed to stop $\pi$, (i.e. the low-level active learning process), if $\tau > t_{max}$. 

Eventually, using the framework of options, we can define the process of goal self-generation, as the self-generation and self-selection of parameterized options, and a goal reaching attempt corresponding to the learning of a particular option. Therefore, the global SAGG-RIAC process can be described as exploring and learning \textit{fields of options}.

\subsection{Lower Time Scale: \\ Active Goal Directed Exploration and Learning}

In SAGG-RIAC, once a goal has been actively chosen at the high-level, the goal directed exploration and learning mechanism at the lower can be carried out in numerous ways: the architecture makes only little assumptions about them, and thus is compatible with many methods such as those described below (this is the reason why SAGG-RIAC is an architecture defining a family of algorithms). Its main idea is to guide the system toward the goal by executing low-level actions which allow a progressive exploration of the world toward this specific goal and that updates at the same time the local corresponding forward and inverse models, leveraging previously learnt correspondences with regression. The main assumptions about the methods that can be used for this lower level are:
\begin{itemize}
\item \textbf{Incremental learning and generalization}: based on the data collected incrementally, the method must be able to build incrementally local forward and inverse models that can be reused later on, in particular when considering other goals, such as the task-space regression techniques presented in \cite{Bitzer09, Baranes09, Kober-RSS-10,Barto12};
\item \textbf{Goal-directed optimization}: when a goal is set, an optimization procedure can improve the parameters of the action policy to reach the goal, such as policy gradient methods \cite{Peters08d,Stulp12} or stochastic optimization \cite{Hansen01};
\end{itemize}
A optional feature, which is a variant of the second assumption above, is:
\begin{itemize}
\item \textbf{Active optimization}: goal-directed optimization of the parameters of the action policy for reaching a self-generated goal. A learning feedback mechanism has to be added such that the exploration is active, and the selection of new actions depends on local measures about the quality of the learned model.
\end{itemize}

In the following experiments that will be introduced, we will use two different methods: one mechanism where optimization is inspired by the SSA algorithm \cite{Schaal94}, coupled with memory-based local forward and inverse regression models using local Moore-Penrose pseudo-inverses, and a more generic optimization algorithm mixing stochastic optimization with memory-based regression models using pseudo-inverse.
Other kinds of techniques could be used. For the optimization part, algorithms such as natural actor-critic architectures in model based reinforcement learning \cite{Peters08d}, algorithms of convex optimization \cite{Dattorro11}, algorithms of stochastic optimization like CMA (e.g. \cite{Hansen01}), or path-integral methods (e.g. \cite{Stulp11, Stulp12}). 

For the regression part, we are here using a memory-based approach, which if combined with efficient data storage and access structures \cite{Arya98,Muja09}, scales well from a computational point of view. Yet, if memory limits would be a limited resource, and as little assumption about the low-level regression algorithms are made in the SAGG-RIAC architecture, parameterized models allowing to control memory requirements such as Neural networks, Support Vector Regression, Gaussian Process Regression could instead be considered \cite{Sigaud11}, such as in \cite{Bitzer09, Baranes09, Kober-RSS-10,Barto12}.

\subsection{Higher Time Scale: \\ Goal Self-Generation and Self-Selection}
The Goal Self-Generation and Self-Selection process relies on a feedback defined using the concept of competence, and more precisely on the competence improvement in given regions (or subspaces) of the task space where goals are chosen. The measure of competence can be computed at different instants of the learning process. First, it can be estimated once a reaching attempt in direction of a goal has been declared as terminated. Second, for robotic setups which are compatible with this option, competence can be computed during low-level reaching attempts. In the following sections, we detail these two different cases:
\subsubsection{Measure of Competence for a Terminated Reaching Attempt}
A reaching attempt for a goal is considered terminated according to two conditions:
\begin{itemize}
\item A timeout related to a maximum number of iterations allowed by the low-level of active learning has been exceeded.
\item The goal has effectively been reached.
\end{itemize}
We introduce a measure of competence for a given goal reaching attempt as dependent on two metrics: the similarity between the point in the task space $y_f$ attained when the reaching attempt has terminated, and the actual goal $y_g$; and the respect of constraints $\rho$. These conditions are represented by a cost, or competence, function $C$ defined in $[-\infty; 0]$, such that higher $C(y_g,y_f, \rho)$ will be, the more a reaching attempt will be considered as efficient. From this definition, we set a measure of competence $\Gamma_{y_g}$ directly linked with the value of $C(y_g,y_f, \rho)$:
\begin{eqnarray}
\Gamma_{y_g}= \left\{
\begin{array}{ll}
 C(y_g,y_f, \rho) & \mbox{if} \ C(y_g,y_f, \rho) \le \varepsilon_{sim} < 0\\
  0   & \mbox{otherwise}  \nonumber
 \end{array}
 \right.
\end{eqnarray}

where $\varepsilon_{sim}$ is a tolerance factor such that $C(y_g,y_f, \rho) > \varepsilon_{sim}$ corresponds to a goal reached. We note that a high value of $\Gamma_{y_g}$ (i.e. close to $0$) represents a system that is competent to reach the goal $y_g$ while respecting constraints $\rho$. A typical instantiation of $C$, without constraints $\rho$, is defined as $C(y_g,y_f, \emptyset)= - \|y_g - y_f\|^2$, and is the direct transposition of prediction error in RIAC \cite{Oudeyer07, Baranes09} to the task space in SAGG-RIAC. Yet, this competence measure might take some other forms in the SAGG-RIAC architecture, such as the variants explored in the experiments below. 

\subsubsection{Measure of Competence During a Reaching Attempt or During Goal-Directed Optimization}
\label{subgoals}
When the system exploits its previously learnt models to reach a goal $y_g$, using a computed $\pi_\theta$ through adequate local regression, or when it is using the low-level goal-directed optimization to optimize the best current $\pi_\theta$ to reach a self-generated goal $y_g$, it does not only collect data allowing to measure its competence to reach $y_g$, but since the computed $\pi_\theta$ might lead to a different effect $y_e \neq y_g$, it also allows to collect new data for improving the inverse model and the measure of competence to reach other goals in the locality of $y_e$. This allows to use all experiments of the robot to update the model of competences over the space of paremeterized goals.

\subsubsection{Definition of Local Competence Progress}
The active goal self-generation and self-selection relies on a feedback linked with the notion of competence introduced above, and more precisely on the monitoring of the progress of local competences. We first need to define this notion of local competence. Let us consider a subspace called a region ${R} \subset Y$. Then, let us consider different measures of competence $\Gamma_{y_i}$ computed for different attempted goals $y_i \in R$, in a time window consisting of the $\zeta$ last attempted goals.  For the region $R$, we can compute a measure of competence $\Gamma$ that we call a \textit{local measure} such that:
\begin{eqnarray}
 \Gamma &=& \left(\frac{\sum_{{y_j} \in {R}}(\Gamma_{y_j})}{|{R}|} \right) 
 \end{eqnarray}
with $|{R}|$, cardinal of ${R}$.



\begin{figure}
\center
\includegraphics[width=\linewidth]{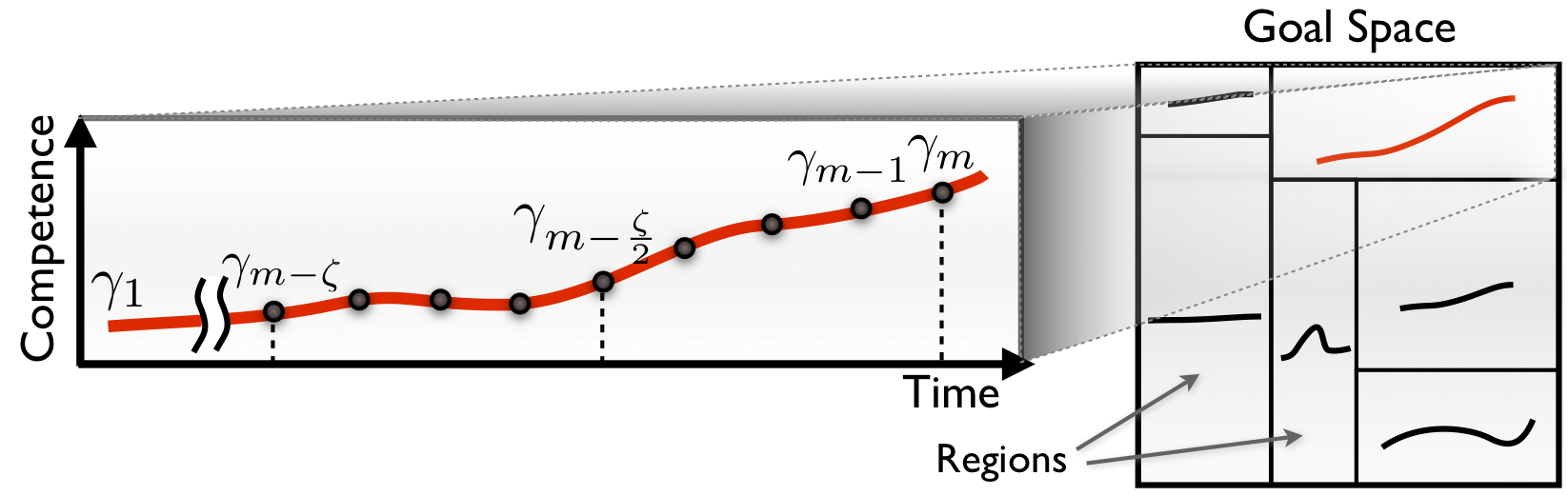}
\caption{Task space and example of regions and subregions split during the learning process according to the competence level. Each region displays its competence level over time, measure which is used for the computation of the interest according to equation \ref{interest}.}
\label{competenceTime}
\end{figure}

Let us now consider different regions ${R}_i$ of $Y$ such that ${R}_i \subset Y$, $\bigcup_i {R}_i = Y$ 
(initially, there is only one region which is then progressively and recursively split; see below and see Fig. \ref{competenceTime}). 
Each ${R}_i$ contains attempted goals $\{y_{i_1,t_1}, y_{i_2,t_2}, ..., y_{i_k,t_k}\}_{{R}_i}$ and corresponding competences obtained $\{\Gamma_{y_{i_1,t_1}}, \Gamma_{y_{i_2,t_2}}, ..., \Gamma_{y_{i_k,t_k}}\}_{{R}_i}$, indexed by their relative time order of experimentation $t_1< t_2< ...< t_k | t_{n+1} = t_{n}+1 $ inside this precise subspace ${R}_i$ ($t_i$ are not the absolute time, but integer indexes of relative order in the given region). 

An estimation of interest is computed for each region $R_i$. The interest $interest_i$ of a region $R_i$ is described as \textit{the absolute value of the derivative of local competences inside ${R}_i$, hence the amplitude of local competence progress, over a sliding time window of the $\mathbf{\zeta}$ more recent goals attempted inside ${R}_i$} (equation \ref{interest}):
\begin{center}

\begin{eqnarray}
interest_i =  \frac{\left| \left(\displaystyle \sum_{j=| {R}_i|-\zeta}^{|{R}_i|-\frac{\zeta}{2}} \Gamma_{y_j} \right) - \left(\displaystyle \sum_{j=|{R}_i|-\frac{\zeta}{2}}^{|{R}_i|} \Gamma_{y_j} \right) \right|}{\zeta}
\label{interest}
\end{eqnarray}
\end{center}
By using a derivative, the interest considers the \textit{variation of competences}, and by using an absolute value, it considers cases of \textit{increasing and decreasing competences}. In SAGG-RIAC, we will use the term \textit{competence progress} with its general meaning to denote this increase and decrease of competences.

An increasing competence signifies that the expected competence gain in $R_i$ is important. Therefore, potentially, selecting new goals in regions of high competence progress could bring both a high information gain for the learned model, and also drive the reaching of not previously achieved goals. 

Depending on the starting position and potential evolution of the environment or of the body (e.g. breaking of a body part), a decrease of competences inside already well-reached regions can arise. In this case, the system should be able to focus again in these regions in order to at least verify the possibility to re-establish a high level of competence inside. This explains the usefulness to consider the absolute value of the competence progress as shown in equation \ref{interest}. 

Using a sliding window in order to compute the value of interest prevents the system from keeping each measure of competence in its memory, and thus limits the storage resource needed by the core of the SAGG-RIAC architecture.

\subsubsection{Goal Self-Generation Using the Measure of Interest}
\label{splitSection}
Using the previous description of interest, the goal self-generation and self-selection mechanism carries out two different processes:
\begin{enumerate}
\item Splitting of the space $Y$ where goals are chosen, into subspaces, according to heuristics that allows to maximally discriminate areas according to their levels of interest.
\item Selecting the next goal to perform.
\end{enumerate}
Such a mechanism has been described in the RIAC algorithm introduced in \cite{Baranes09}, but was previously applied to the actuator space $S$ rather than to the goal/task space $Y$ as is done in SAGG-RIAC. 
Here, we use the same kind of methods such as a recursive split of the space, each split being triggered once a predefined maximum number of goals $g_{max}$ has been attempted inside. Each split is performed such that it maximizes the difference of the interest measure described above in the two resulting subspaces. This allows the easy separation of areas of differing interest and therefore of differing reaching difficulty. More precisely, here the split of a region $R_n$ into $R_{n+1}$ and $R_{n+2}$ is done by selecting among $m$ randomly generated splits, a split dimension $j \in |Y|$ and then a position $v_j$ such that:
\begin{itemize}
\item All the $y_{i}$ of $R_{n+1}$ have a $j^{th}$ component smaller than $v_j$;
\item All the $y_{i}$ of $R_{n+2}$ have a $j^{th}$ component higher than $v_j$;
\item The quantity $Qual(j,v_j) = card(R_{n+1}). card(R_{n+2}).\| interest_{R_{n+1}} - interest_{R_{n+2}} \|$ is maximal;
\end{itemize}

Finally, as soon as at least two regions exist after an initial random exploration of the whole space, goals are chosen according to the following heuristics, selected according to probabilistic distributions:

\textbf{1.} $mode(1)$:  in $p_1$\% percent (typically $p_1=70$\%)  of goal selections, a random goal is chosen along a uniform distribution inside a region which is selected with a probability proportional to its interest value: 
\begin{eqnarray}
P_n = \frac{ interest_n - \textbf{min}(interest_i)}{\sum_{i=1}^{|R_n|}interest_i - \textbf{min}(interest_i)}
\end{eqnarray}
Where $P_n$ is the selection probability of the region ${R}_n$, and $interest_i$ corresponds to the current $interest$ of the region ${R}_i$. \\
\textbf{2.} $mode(2)$:  in $p_2$\%  (typically $p_2=20$\% of cases), a random goal is chosen inside the whole space $Y$. \\
\textbf{3.} $mode(3)$:  in $p_3$\% (typically $p_3=10$\%), a region is first selected according to the interest value (like in $mode(1)$) and then a new goal is generated close to the already experimented one which received the lowest competence estimation.

\subsubsection{Reduction of the Initiation Set}
In order to improve the quality of the learned inverse model, we add a heuristic inspired by two observations on infant motor exploration and learning. The first one, proposed by Berthier et al. \cite{Berthier99} is that  infant's reaching attempts are often preceded by movements that either elevate their hand or move their hand back to their side. And the second one, noticed in \cite{Rolf10a}, is that infants do not try to reach for objects forever but sometimes relax their muscles and rest. Practically, these two characteristics allow them to reduce the number of initiation positions that they use to reach an object, which simplifies the reaching problem by letting them learn a reduced number of reaching movements.

Such mechanism can be transposed in robotics to motor learning of arm reaching tasks as well as  other kind of skills such as locomotion or fishing as shown in experiments below. In such a framework, it directly allows a highly-redundant robotic system to reduce the space of initiation states used to learn to reach goals, and also typically prevent it from experimenting with too complex actuator configurations. We add such a process in SAGG-RIAC, by specifying a rest position $(s_{rest}, y_{rest})$ reachable without any need of planning from the system, that is set for each $r$ subsequent reaching attempts (we call $r$ the reset value, with $r > 0$). 

\subsection{New Challenges of Unknown Limits of the Task Space}
\label{additionSubgoals}
In traditional active learning methods and especially knowledge-based intrinsically motivated exploration \cite{Barto05, Oudeyer07, Marshall04, Schmidhuber91, Blank05}, the system is typically designed to select actions to perform inside a set of values inside an already known interval (for instance, the range of angles that can be taken by a motor, or the phases and amplitudes of CPGs which can be easily identified). In these cases, the challenge is to select which areas would potentially give the most information to the system, to improve its knowledge, inside this fixed range of possibilities. As argued earlier, a limit of these approaches is that they become less and less efficient as the dimensionality of the control space increases. Competence based approaches allow to address this issue when a low-dimensional  task space can be identified. Nevertheless, in that case, a new problem arises when considering unbounded learning: the space where goals are reachable can be extremely large and it is generally very difficult to predict its limits and undesirable to ask the engineer to identify them. Therefore, when carried out in large spaces where the reachable area is only a small part of it, the algorithm could necessitate numerous random goal self-generations to be able to estimate interests of different subregions. In order to reduce this number, and help the system to converge easily toward regions where competence can be improved, we emphasize two different mechanisms that can be used in SAGG-RIAC, during a reaching attempt:

\begin{enumerate}
\item Conservation of every point reached inside the task space even if they do not correspond to the attempted goal (see section \ref{subgoals}): when the robot performs a reaching attempt toward a goal $y$, and, instead of reaching it, terminates at another state $y'$, we consider $y'$ as a goal reached with a value of competence depending on constraints $\rho$. In cases where no constraints are studied, we can consider the $y'$ as another goal reached with the highest level of competence. 
\item Addition of subgoals: in robotic setups where the process of goal reaching can be subdivided and described using subgoals which could be fixed on the pathway toward the goal, we artificially add states $y_1, y_2, ..., y_n$ that have to be reached before $y$ while also respecting the constraints $\rho$, and estimate a competence measure for each one. 
\end{enumerate}
The consideration of these two heuristics has important advantages: first, they can significantly increase the number of estimations of competence, and thus the quantity of feedback returned to the goal self-generation mechanism. This reduces the number of goals that have to be self-generated to bootstrap the system, and thus the number of low-level iteration required to extract first interesting subspaces. Also, by creating areas of different competence values around already reached states, they influence the discovery of reachable areas. Finally, they result in an interesting emergent phenomena: they create a growing area of increasing competence around the first discovered reachable areas. Indeed, by obtaining values of competences inside reachable areas, the algorithm is able to split the space first in these regions, and compute values of interest. These values of interest will typically be high in already reached areas and influence the goal self-generation process to create new goals in its proximity. Once the level of competence becomes important and stabilized in efficiently reached areas, the interest becomes null, then, new areas of interest close to these ones will be discovered, and so on.

\subsection{PseudoCode}
Pseudo-code \ref{alg:PseudoCode2} and algorithm \ref{alg:PseudoCode1} present the flow of operations in the SAGG-RIAC architecture. Algorithms \ref{alg:PseudoCode3} and \ref{alg:PseudoCode4} are simple alternative examples of low-level goal-directed optimization algorithms that are used in the experimental section, but they could be replaced by other algorithms like $PI^{2}-CMA$ \cite{Stulp12}, $CMA$ \cite{Hansen01}, or those presented in \cite{Peters08c}. The function \textbf{Inefficient} can also be built in numerous manners and will not be described in details in the pseudo-code (examples will be described then for each experimentation). Its function is to judge if the current model has been efficient enough to reach or come closer to the decided goal, or if the model has to be improved in order to reach it. 

In the following sections, we will present two different kinds of experiments. The first one is a reaching experiment where a robotic arm has to learn its inverse kinematics to reach self-generated end-effector positions. It uses an evolving context $s \in S$, also called setpoint in SSA, representing its current joint configuration. Therefore, it can be described by the relationship $(s,a) \rightarrow y$ where $s, a$ and $y$ can evolve. It is thus possible to use a goal-directed optimization algorithm very similar to SSA in this experiment, like the one in algorithm \ref{alg:PseudoCode3}.

In the two other experiments, in contrast, we control the robots using parameterized motor synergies and consider a fixed context (a rest position)  $s \in S$ where the robot is reset before each action: we will first consider a quadruped learning omnidirectional locomotion, and then an arm controlling a flexible fishing rod learning to put the float in precise self-generated positions on top of the water. Thus, these systems can be described by the relationship $(s,\pi_\theta) \rightarrow y$, where $s$ will here be fixed and $\theta$ will be the parameters of the motor synergy used to control the robots. Thus, a variation of setpoint being prevented here, a variant of SSA will be proposed for such experiments (similar to a more traditional optimization algorithm), where the context will not evolve and always be reset, like in algorithm \ref{alg:PseudoCode4}.

\begin{algorithm}[lt!]
 \caption{The SAGG-RIAC Architecture}
 \label{alg:PseudoCode2}
\begin{algorithmic}
\STATE $S$: State/Context space
\STATE $\Pi$: Space of paremeterized action policies $\pi_\theta$
\STATE $Y$: Space of parameterized tasks $y_i$
\STATE $\mathbf{M}$: regression model of the forward mapping $(S,\Pi) \rightarrow Y$ 
\STATE $\mathbf{M^{-1}}$: regression model of the inverse mapping $(S,Y) \rightarrow \Pi$ 
\STATE $\mathbf{R}$: set of regions $\mathbf{R}_i \subset Y$ and corresponding measures $interest_i$; 

\STATE {\bfseries input:} thresholds $\varepsilon_{C}$; $\varepsilon_{max}$; $timeout$
\STATE {\bfseries input:} rest position $s_{rest} \in S$; reset value: $r$
\STATE {\bfseries input:} starting position $s_{start} \in S$
\STATE {\bfseries input:} number of explorative movements $q \in \mathbb{N}$
\STATE {\bfseries input:} starting time: $t$
\STATE {\bfseries input:} $q$ budget of physical experiments for goal-directed optimization
\LOOP
   \STATE{Reset the system in the resting state ($s_{start}$ = $s_{rest}$) every $r$ iteration of the loop;}
    \STATE{\textbf{\underline{Active Goal Self-Generation (high-level):}}}
     \STATE{Self-generate a goal $y_g \in Y$ using the $mode(m \in {[1;2;3]})$ with probability $p_m$ (see Section 2.4.4.)}
        \STATE{\textbf{\underline{Active Goal-Directed Exploration and Learning (low-level):}}}
        \STATE{Let $s_{t}$ represent the current context of the system}
        \IF{Made possible by the sensorimotor space}
        \STATE{Compute a set of subgoals $\{y_1, y_2, ... , y_n\} \in Y$ on the pathway toward $y_g;$ (e.g. with a planning algorithm that takes $s$, $\mathbf{M}$, $\mathbf{M^{-1}}$ and $y_g$ into account);}
        \ELSE 
        \STATE{$\{y_1, y_2, ... , y_n\} = \emptyset;$}
        \ENDIF
	 \FOR{each $y_j$ in  $\{y_1, y_2, ... , y_n\} \cup y_g$}
    \WHILE{ $\Gamma_{y_{j}} \le \varepsilon_{C}$ \& $timeout$ not exceeded}
        \STATE{Compute and execute an action/synergy $\pi_{\theta_j} \in \Pi$ using $\mathbf{M}^{-1}$ such that it targets $y_{j}$, e.g. using techniques such as in \cite{Bitzer09, Baranes09, Kober-RSS-10,Barto12};}
        \STATE{Get the resulting actually performed $\widetilde{y_{j}}$ and update $\mathbf{M}$ and $\mathbf{M}^{-1}$ with new data $(s_t, \theta_j, \widetilde{y_{j}})$}
        \STATE{Compute the competence $\Gamma_{\widetilde{y_{j}}}$  (see section 2.4.1.)}
        \STATE{\textbf{UpdateRegions}($\mathbf{R}, \widetilde{y_{j}}, \Gamma_{\widetilde{y_{j}}}$);}
        
        \IF{experiment with evolving context}
        \STATE Goal-directed optimization of $\theta_j$ to reach $y_j$, with SSA like algorithm such as Algorithm \ref{alg:PseudoCode3}, and given a budget of $q$ allowed physical experiments;
        \ELSE 
        \STATE Goal-directed optimization of $\theta_j$ to reach $y_j$ such as algorithm \ref{alg:PseudoCode4}, or alternatively algorithms such as \cite{Stulp12, Hansen01, Peters08c},  and given a budget of $q$ allowed physical experiments;
                \ENDIF
       \ENDWHILE

       \STATE{Compute the competence $\Gamma_{y_j}$  (see section 2.4.1.)}
       \STATE{\textbf{UpdateRegions}($\mathbf{R}$, $y_j$, $\Gamma_{y_j}$);}
    
 \ENDFOR

\ENDLOOP
\end{algorithmic}
\end{algorithm}

\begin{algorithm}[t!]
 \caption{Pseudo-Code of UpdateRegions}
 \label{alg:PseudoCode1}
\begin{algorithmic}
\STATE {\bfseries input:} $\mathbf{R}$: : set of regions $\mathbf{R}_i \subset Y$ and corresponding measures $interest_i$; 
\STATE {\bfseries input:} $y_t$: current goal
\STATE {\bfseries input:} $\Gamma_{y_t}$: competence measure for $y_t$
\STATE Let $g_{Max}$ be the maximal number of elements inside a region
\STATE Let $\zeta$ be a time window used to compute the interest
\STATE{Find the region $R_n$ in $\mathbf{R}$ such that $y_t \in R_n$;}
\STATE{Let $k = card(R_n)$}
\STATE{Add $\Gamma_{y_{t},k}$ in $R_n$, where $k$ is an indice indicating the ordinal order in which $\Gamma_{y_t}$ was added in the region as compared to other measures of competences in $R_n$ ;}
\STATE{Compute the new value of $interest_n$ of $R_n$ according to each $\Gamma_{y_{i},l} \in R_n$ such that:} 
\begin{center}
\STATE{$interest_n =  \frac{\left| \left(\displaystyle \sum_{l=| {R}_n|-\zeta}^{|{R}_n|-\frac{\zeta}{2}} \Gamma_{y_{i},l} \right) - \left(\displaystyle \sum_{l=|{R}_n|-\frac{\zeta}{2}}^{|{R}_n|} \Gamma_{y_{i},l} \right) \right|}{\zeta}$}
\end{center}
\IF{ $card(R_n) > g_{max}$}
\STATE{Split $R_n$; (see text, section \ref{splitSection})}
\ENDIF
\end{algorithmic}
\end{algorithm}

\begin{algorithm}[t!]
 \caption{Example of Pseudo-Code for the Low-Level Goal-Directed Exploration with Evolving Context (used in the experimentation introduced section \ref{reachingTaskExp})}
 \label{alg:PseudoCode3}
\begin{algorithmic}
        \STATE {\bfseries input:} $q$ is the budget of physical experiment allowed to the robot for local optimization;
	\STATE \textbf{Update} the current context $s_t = s_j$; \COMMENT{where $s_j$ is the context after having performed $\pi_{\theta_j}$}
        \IF{\textbf{Inefficient}($\mathbf{M^{-1}}$, $\widetilde{y_{j}}$, $y_{j}$)}
        \STATE{\textbf{Local Exploration Phase:}}       
 	\FOR{$i = 1$ to $q$}
        \STATE{Perform action policy $\pi_{\theta_i}$ with $\theta_i$ drawn randomly in the vicinity of $\theta_j$;}
        \STATE{Measure the resulting $y_{i}$ and $s_i$;}
        \STATE{Update $\mathbf{M}$ and $\mathbf{M^{-1}}$ with $(s_t, \theta_i, y_{i})$;}
        \STATE{\textbf{Update} the context $s_t = s_i$;} 
        \STATE{Compute the competence $\Gamma_{y_i}$;}
        \STATE{\textbf{UpdateRegions}($\mathbf{R}$, $y_i$, $\Gamma_{y_i}$);}
        \ENDFOR
        \ENDIF
        \end{algorithmic}
\end{algorithm}

\begin{algorithm}
 \caption{Example of Pseudo-Code for the Low-Level Goal-Directed Exploration with a Fixed or Resettable Context (used in the experiments introduced sections \ref{LabelExp2} and \ref{LabelExp3})}
 \label{alg:PseudoCode4}
\begin{algorithmic}
        \STATE {\bfseries input:} $q$ is the budget of physical experiment allowed to the robot for local optimization;
	\STATE \textbf{Reset} the current context: $s_t = s_{rest}$;
        \IF{\textbf{Inefficient}($\mathbf{M^{-1}}$, $\widetilde{y_{j}}$, $y_{j}$)}
        \STATE{\textbf{Local Exploration Phase:}}   
        \STATE{Initialize $\theta_k = \theta_j$ and $y_k = y_{j}$ and $\Gamma_{y_k}=\Gamma_{y_j}$}
	\FOR{$i = 1$ to $q$}
       \STATE{Perform $\pi_{\theta_i}$ where $\theta_i$ is drawn randomly in the vicinity of $\theta_k$;} 
        \STATE{Observe the resulting $y_i$;}
        \STATE{Update $\mathbf{M}$ and $\mathbf{M^{-1}}$ with the resulting $(s_t, \theta_i, y_{i})$;}
        \STATE{\textbf{Reset} the current context: $s_t = s_{rest}$;}
        \STATE{Compute the competence $\Gamma_{y_i}$;}
        \STATE{\textbf{UpdateRegions}($\mathbf{R}$, $y_i$, $\Gamma_{y_i}$);}
        \IF{$\Gamma_{y_i} > \Gamma_{y_k}$}
              \STATE{$\theta_k = \theta_i$}
              \STATE{$y_k = y_i$}
        \ENDIF
        \ENDFOR      
          \ENDIF
\end{algorithmic}

\end{algorithm}



\section{Experimental Setup 1: Learning Inverse Kinematics with a Redundant Arm}

In this section, we propose an experiment carried out with a robotic arm which has to explore and learn its forward and inverse kinematics. Also, before discussing the details of our active exploration approach in this first experimentation case, we firstly define the representations of the models and control paradigms involved in this experiment. Here, we focus on robotic systems whose actuators are settable by positions and velocities, and restrict our analysis to discrete time models. 

Allowing robots to be self-adaptive to environmental conditions and changes in their own geometry is an important challenge of machine learning. These changes in the robot geometry directly have an impact on its Inverse Kinematics \textbf{IK}, relating workspace coordinates (where tasks are usually specified), to actuators coordinates (like joint position, velocity, or torque used to command the robot). Learning inverse kinematics is useful in numerous machine learning cases, such as when no accurate kinematic model of a robot is available or when an online calibration is needed due to sensor or motor imprecision. Moreover, in developmental robotics studies, the a priori knowledge of a precise model of the body is often avoided, because of its implausibility from the biological point of view. In the following experiment, we assume that the inverse kinematics of our system is totally unknown, and we are interested in studying how SAGG-RIAC can efficiently guide the discovery and learning of its inverse kinematics. 
\subsection{Control Paradigms for Learning Inverse Kinematics}

Let us mathematically formulate forward and inverse kinematics relations. We define the intrinsic coordinates (joint/actuator positions) of a manipulator as the $n$-dimensional vector $S = \alpha \in \mathbb{R}^n$, and the position and orientation of the manipulator's end-effector as the $m$-dimensional vector $y \in \mathbb{R}^m$. Relative to this formalization, actions of the robot corresponds to speed commands parameterized by a vector $\theta = \dot\alpha \in \mathbb{R}^n$  which controls the instantaneous speed of each of the $n$ joints of the arm. The forward kinematic function of this system is generally written as $y = f(\alpha)$, and inverse kinematics relationship is defined as $\alpha = f^{-1}(y)$.

When a redundant manipulator is considered ($n > m$), or when $m = n$, solutions to the inverse relationship are generally non-unique \cite{Handbook08}. The problem posed to inverse learning algorithms is thus to determine particular solutions to $\alpha = f^{-1}(y)$, when multiple solutions exists. A typical approach used for solving this problem considers local methods, which learn relationships linking small changes $\Delta \alpha$ and $\Delta y$ :
\begin{eqnarray}
\dot y &=& J(\alpha) \dot \alpha
\end{eqnarray}
where $J(\alpha)$ is the Jacobian matrix.

Then, using the Jacobian matrix and inverting it to get a single solution $\dot \alpha$ corresponding to a desired $\dot y$ raises the problem of the non-convexity property of this last equation. A solution to this non-convex problem has then been proposed by Bullock in \cite{Bullock93} who converted it into a convex problem, by only considering the learning task within the spatial vicinity $\widehat{\dot \alpha}$ of a particular $\alpha$ : 
\begin{eqnarray}
  \dot y &=& J(\alpha) \widehat{\dot \alpha}
\end{eqnarray}

\subsection{Representation of Forward and Inverse Models to be Learnt}

\begin{figure}
\centerline{\includegraphics[width=\linewidth]{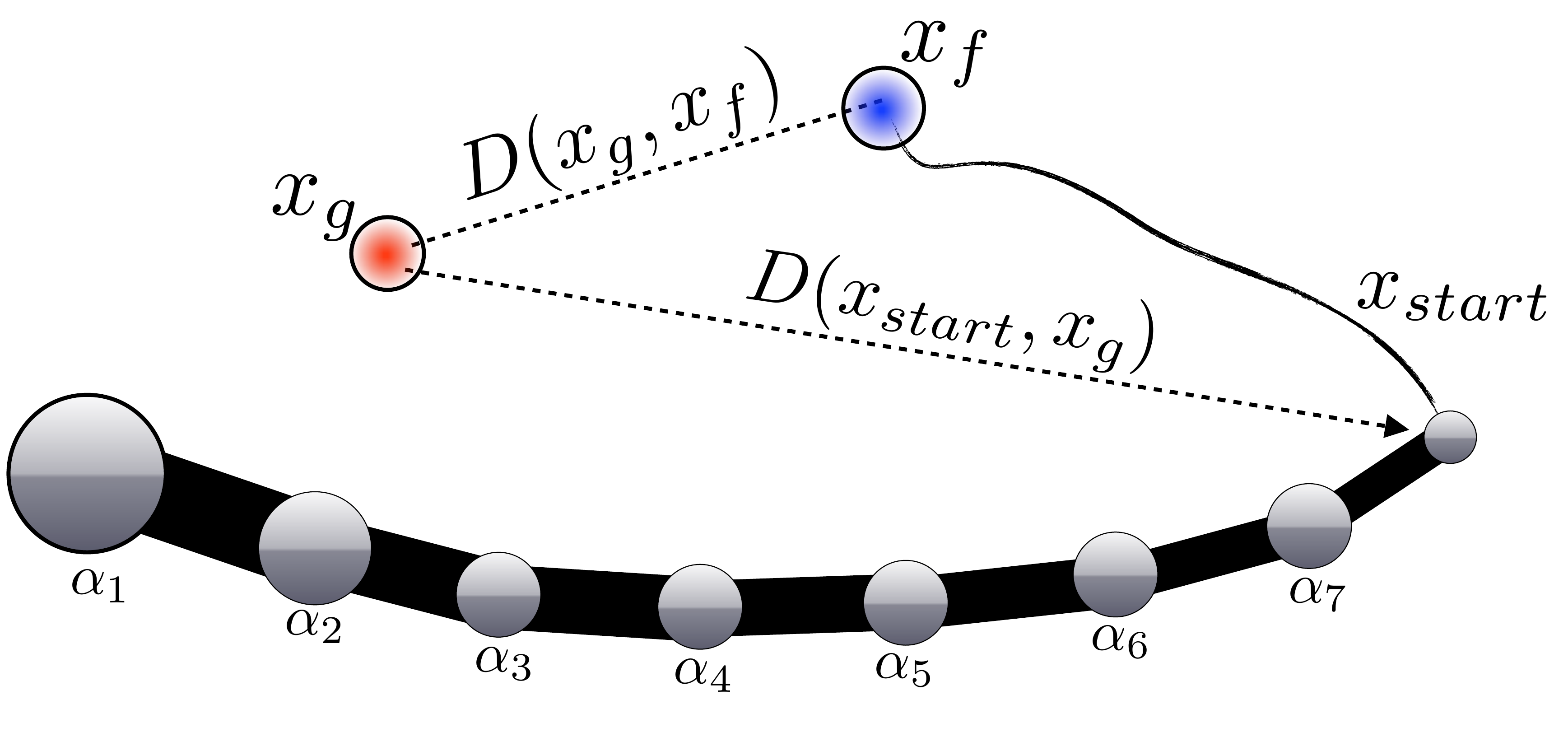}}
\caption{Values used to compute the competence $\Gamma_{y_g}$, considering a manipulator of 7 degrees-of-freedom, in a 2 dimensions operational/task space. Here, the arm is set in a position called \textit{rest position} which is not straight and slightly bent. $(\alpha_{rest}, y_{rest})$.}
\label{FigArm}
\end{figure}

We use here \textit{non-parametric} models which typically determine local models in the vicinity of a current datapoint. By computing a model using parameterized functions on datapoints restrained to a locality, they have been proposed as useful for real time queries, and incremental learning. Learning inverse kinematics typically deals with these kind of constraints, and these local methods have thus been proposed as an efficient approach to IK learning \cite{Vijayakumar05, Sigaud11}. In the following study, we use an incremental version of the Approximate Nearest Neighbors algorithm (ANN) \cite{Muja09}, based on a tree split using the k-means process, to determine the vicinity of the current $\alpha$. Also, in the environments that we use to introduce our contribution, we do not need highly robust, and computationally very complex regression methods. Using the pseudo-inverse of Moore-Penrose \cite{Albert72} to compute the pseudo-inverse $J^+(\alpha) $ of the Jacobian  $J(\alpha)$ in a vicinity $\widehat{\dot \alpha}$ is thus sufficient. Possible problems happening due to singularities \cite{Handbook08, Chiaverini97, Salaun10} being bypassed by adding noise in the joint configurations (see \cite{Rolf11} for a study about this problem).

Also, in the following equation, we use this method to deduce the change $\Delta \alpha$ corresponding to a $\Delta x$, for a given joint position $\alpha$:
\begin{eqnarray}
      \dot \alpha &=& J^+(\alpha)  \dot y
      \label{jacobianPseudo}
\end{eqnarray}

\subsection{Robotic Setup}
\label{reachingTaskExp}

In the following experiments, we consider a $n$-dimensional manipulator controlled in position and speed (as many of today's robots), updated at discrete time values. The vector $\alpha \in \mathbb{R}^n$ which represents joint angles corresponds to the context/state space $S$ and the vector $y \in \mathbb{R}^m$ which is the position of the manipulator's end-effector in $m$ dimensions in the Euclidian space $\mathbb{R}^m$ corresponds to the task space $Y$ (see Fig. \ref{FigArm} where $n = 7$ and $m=2$). We evaluate how the SAGG-RIAC architecture can be used by a robot to learn how to reach all reachable points in the environment $Y$ with this arm's end-effector. Learning the inverse kinematics is here an online process that arises each time a micro-action $\theta = \Delta \alpha \in A$ is executed by the manipulator: by doing each micro-action, the robot stores measures $(\alpha, \Delta \alpha, \Delta x)$ in its memory and creates a database $Data$ which contains elements $(\alpha_i,\Delta \alpha_i, \Delta y_i)$ representing the discovered change  $\Delta y_i$ corresponding to a given $\Delta \alpha_i$ in the configuration $\alpha_i$ (this learning entity can be called a schema according to the terminology of Drescher \cite{Drescher91}). These measures are then reused online to compute the Jacobian $J(\alpha) = \Delta y /  \Delta \alpha$ locally to move the end-effector in a desired direction $\Delta y_{desired}$ fixed toward the self-generated goal. Therefore, we consider a learning problem of $2n$ dimensions, the relationship that the system has to learn being $(\alpha, \Delta \alpha) \rightarrow \Delta y$. Also, in this experiment, where we suppose $Y$ Euclidian, and do not consider obstacles, the direction to a goal can be defined as following a straight line between the current end-effector's position and the goal. 

\subsection{Evaluation of Competence}
In this experiment, in order to clearly illustrate the main contribution of our algorithm, we do not consider constraints $\rho$ and only focus on the reaching of goal positions $y_g$. It is nevertheless important to notice that a constraint $\rho$ has a direct influence on the low-level of active learning of SAGG-RIAC, and thus an indirect influence on the higher level. As using a constraint can require a more complex exploration process guided at the low-level, a more important number of iterations at this level can be required to reach a goal, which could have an influence on the global evolution of the performances of the learning process used by the higher-level of SAGG-RIAC.

We define here the competence function $C$ with the Euclidian distance $D(y_g,y_f)$, between the goal position and the final reached position $y_f$, which is normalized by the starting distance $D(y_{start}, y_g)$, where $y_{start}$ is the end-effector's starting position. This allows, for instance, to give a same competence level when considering a goal at $1 cm$ from the origin position, which the robot approaches at $0.5 cm$ and a goal at $1 mm$, which the robot approaches at $0.5 mm$.

\begin{eqnarray}
C(y_g, y_f, y_{start}) &=& - \frac{D(y_g,y_f)}{D(y_{start}, y_g)}
\end{eqnarray}
where $C(y_g, y_f, y_{start}) = 0$ if $D(y_{start}, y_g) < \varepsilon_{C}$ (the goal is too close from the start position) and $C(y_g, y_f, y_{start}) = -1$ if $D(y_g,y_f) > D(y_{start}, y_g)$ (the end-effector moved away from the goal).

\subsection{Addition of subgoals} Computing local competence progress in subspaces/regions typically requires the reaching of numerous goals. Because reaching a goal can necessitate several micro-actions, and thus time, obtaining competence measures can be long. Also, without biasing the learning process and as already explained in section \ref{additionSubgoals}, we improve this mechanism by taking advantage of the Euclidian nature of $Y$: we increase the number of goals artificially, by adding subgoals on the pathway between the starting position and the goal, where competences are computed. Therefore, considering a starting state $y_{start}$ in $Y$, and a self-generated goal $y_g$, we define the set of $l$ subgoals $\{y_1, y_2, ..., y_l\}$ where $y_i = (i/l) \times (y_g-y_{start})$, that have to be reached before attempting to reach the terminal goal $y_g$.

We also consider another way to increase the number of competence measures which is to take into consideration each experimented position of the end-effector as a goal reached with a maximal competence value. This will typically help the system to distinguish which regions are efficiently covered, and to discover new regions of interest.

\subsection{Active Goal Directed Exploration and Learning}


Here we propose a method inspired by the SSA algorithm to guide the system to learn on the pathway toward the selected goal position $y_g$. This instantiation of the SAGG-RIAC architecture uses algorithm \ref{alg:PseudoCode3} and considers evolving contexts, as explained below.

\subsubsection{Reaching Phase}
\label{sec:reaching}
The reaching phase deals with creating a pathway to the current goal position $y_{g}$. This phase consists of determining, from the current position $y_c$, an optimal micro-action which would guide the end-effector toward $y_{g}$. For this purpose, the system computes the needed end-effector's displacement $\Delta y_{next} = v. \frac{y_c - y_{g}}{\| y_c - y_{g}\|}$ (where $v$ is the velocity bounded by $v_{max}$ and $ \frac{y_c - y_{g}}{\| y_c - y_{g}\|}$ a normalized vector in direction of the goal), and performs the action $\Delta \alpha_{next} = J^+ . \Delta y_{next}$, with $J^+$, pseudo-inverse of the Jacobian estimated in the close vicinity of $\alpha$ and given the data collected by the robot so far. After each action $\Delta y_{next}$, we compute the error $\varepsilon = \| \widetilde{\Delta y}_{next}  - \Delta y_{next} \|$, and trigger the exploration phase in cases of a too high value $\varepsilon > \varepsilon_{max} > 0$. $\varepsilon_{max}$ is thus a parameter which has to be set depending on the range of error $\varepsilon$ that can be experienced, and will be set depending on a tolerance that can be conceded to allow reaching goal positions with the current learned data. While a too high value of  $\varepsilon_{max}$ will prevent exploring and learning new data (the system spending potentially too important amounts of time exploring around a same configuration and get trapped in local minima), too low values of $\varepsilon_{max}$ will prevent an efficient local optimization.

\subsubsection{Exploration Phase}
This phase consists in performing $q \in \mathbb{N}$ small random explorative actions $\Delta \alpha_i$, around the current position $\alpha$, where the variations can be  derandomized such as in \cite{Hansen01}. This allows the learning system to improve its regression model of the relationship $(\alpha, \Delta \alpha) \rightarrow \Delta y$, in the close vicinity of $\alpha$, which is needed to compute the inverse kinematics model around $\alpha$. 
\label{sectionTimeout}
During both phases, a counter is incremented for each micro-action and reset for each new goal. The timeout used to define a goal as unreached and to stop a reaching attempt uses this counter. A maximal quantity of micro-actions is fixed for each goal as directly proportional to the number of micro-action it requires to be reached. In the next experiments, the system is allowed to perform up to $1.5$ times the distance between $y_{start}$ and $y_g$ before stopping the reaching attempt. 

\subsection{Qualitative Results for a 15 DOF Simulated Arm}

In the simulated experiment introduced in this section, we consider the robotic arm presented Fig. \ref{FigArm} with 15 DOF, each limb of the robot having the same length (considering a 15 DOF arm corresponds to a problem of 32 continuous dimensions, with 30 dimensions in the actuator/state space and 2 dimensions in the goal/task space). We set the dimensions of the task space $Y$ as bounded in intervals $y_g \in [0;150] \times [-150;150]$, where 50 units is the total length of the arm, which means that the arm covers less than $1/9$ of the space $Y$ where goals can be chosen (i.e. the majority of areas in the operational/task space are not reachable, which has to be self-discovered by the robot). We fix the number of subgoal per goal to 5, and the maximal number of elements inside a region before a split to $g_{max} = 50$. We also set the desired velocity $v = 2$ units/micro-action, and the number of explorative actions $q = 20$. Moreover, we reset the arm to the rest position $(\alpha_{rest}, y_{rest})$ (position displayed in Fig. \ref{FigArm}) every $r = 1$ reaching attempts. This allows reducing the initiation set and prevent the system from experimenting with too complex joint positions where the arm is folded, and where the Jacobian is more difficult to compute. Using a low value of $r$ is an important characteristic for the beginning of the learning process. A too high value of $r$ prevents learning rapidly how to achieve a maximal amount of goal position, due to the difficulty to reuse the previously learned data when the arm is folded in unknown positions.

The bent character of the rest position is also useful to avoid to begin a micro-action close to a singularity like when the arm is totally unfolded. Also, in this experiment, we consider each experimented position of the end-effector as if it was a goal reached with the maximal competence level (these numerous positions are not displayed in the following figures in order to not overload the illustrations).

\begin{figure}[h!]
\centerline{\includegraphics[width=\linewidth]{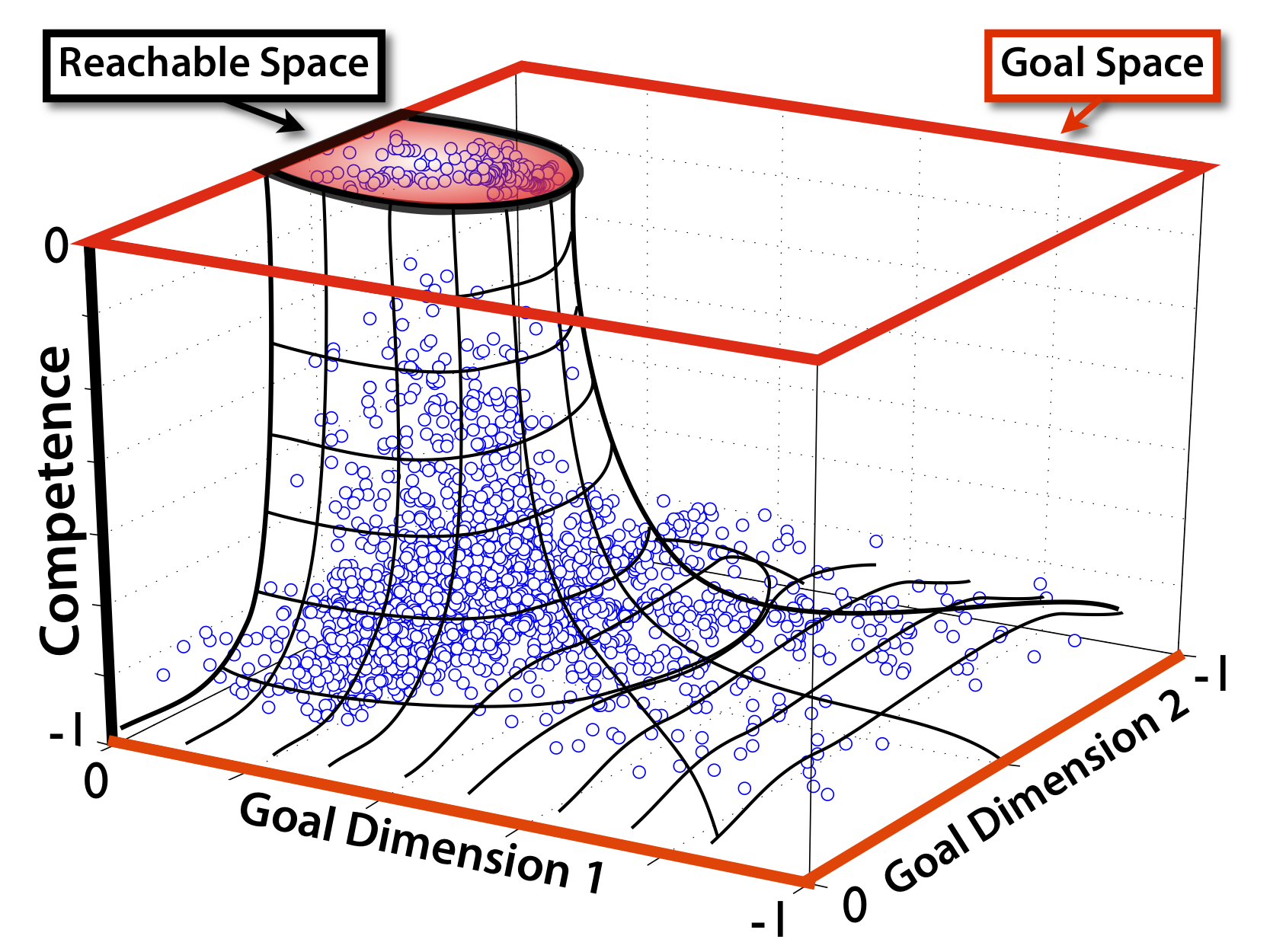}}
\caption{Competence values corresponding to the entire set of self-generated goals collected over an experiment of 30000 micro-actions on a 15 DOF arm. The heterogeneous set of competence values situated inside the reachable space illustrates the typical measures of competence that can be measured in this region over a whole experiment. For a visualization of the evolution of these competence values, see figure \ref{FigureCompetence}}
\label{FigureCompetenceAll}
\end{figure}

\begin{figure*}[t!]
\centerline{\includegraphics[width=0.7\textwidth]{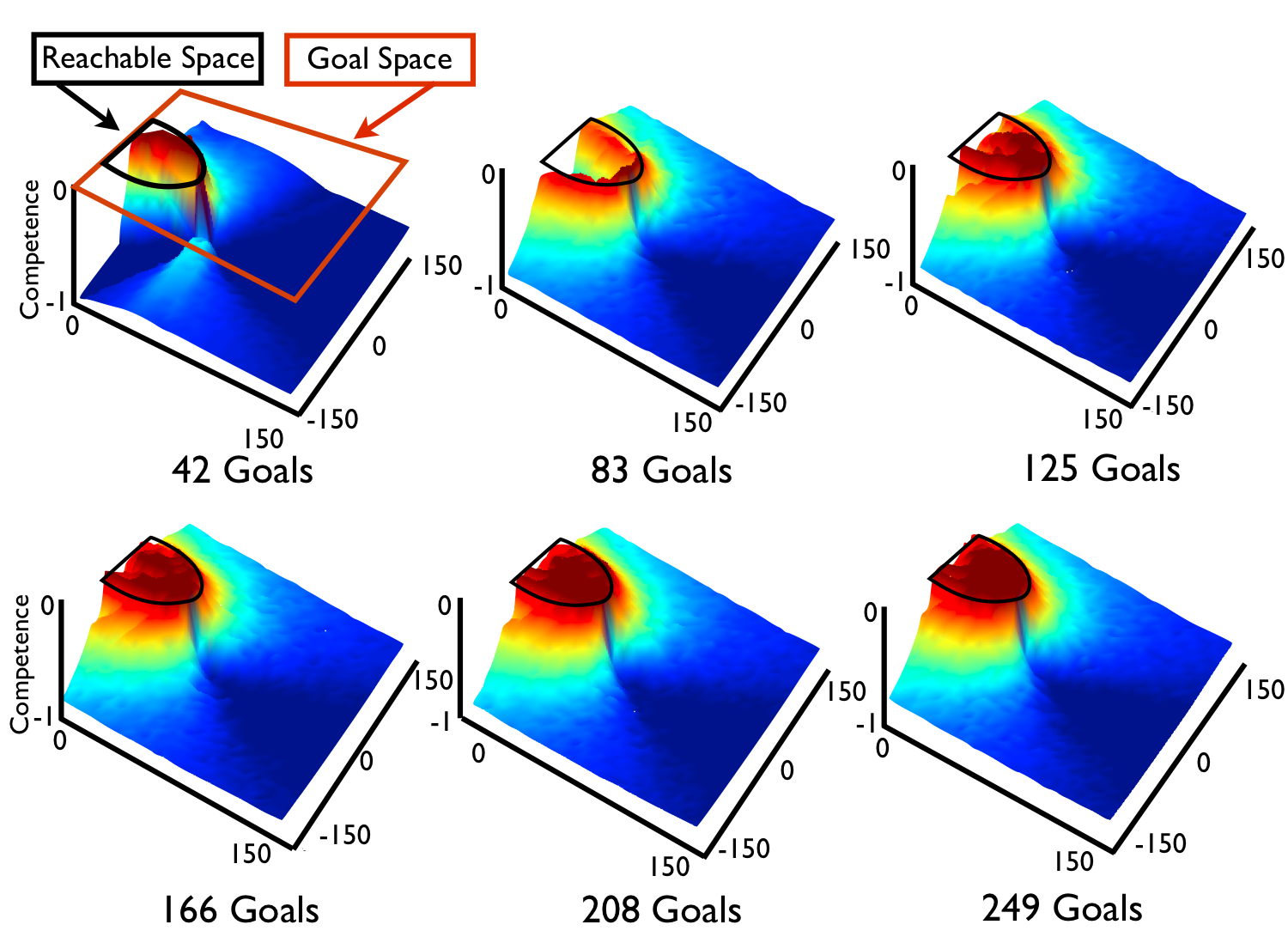}}
\caption{Evolution of competence values corresponding to self-generated goals collected during an experiment of 30000 micro-actions on a 15 DOF arm. Time is indexed by the number of self-generated goals. Higher values (dark red) corresponds to position that has been reached using learned data. (For interpretation of the references to color in this figure legend, the reader is referred to the web version of this article)}
\label{FigureCompetence}
\end{figure*}

\begin{figure*}[t!]
\centerline{\includegraphics[width=0.7\textwidth]{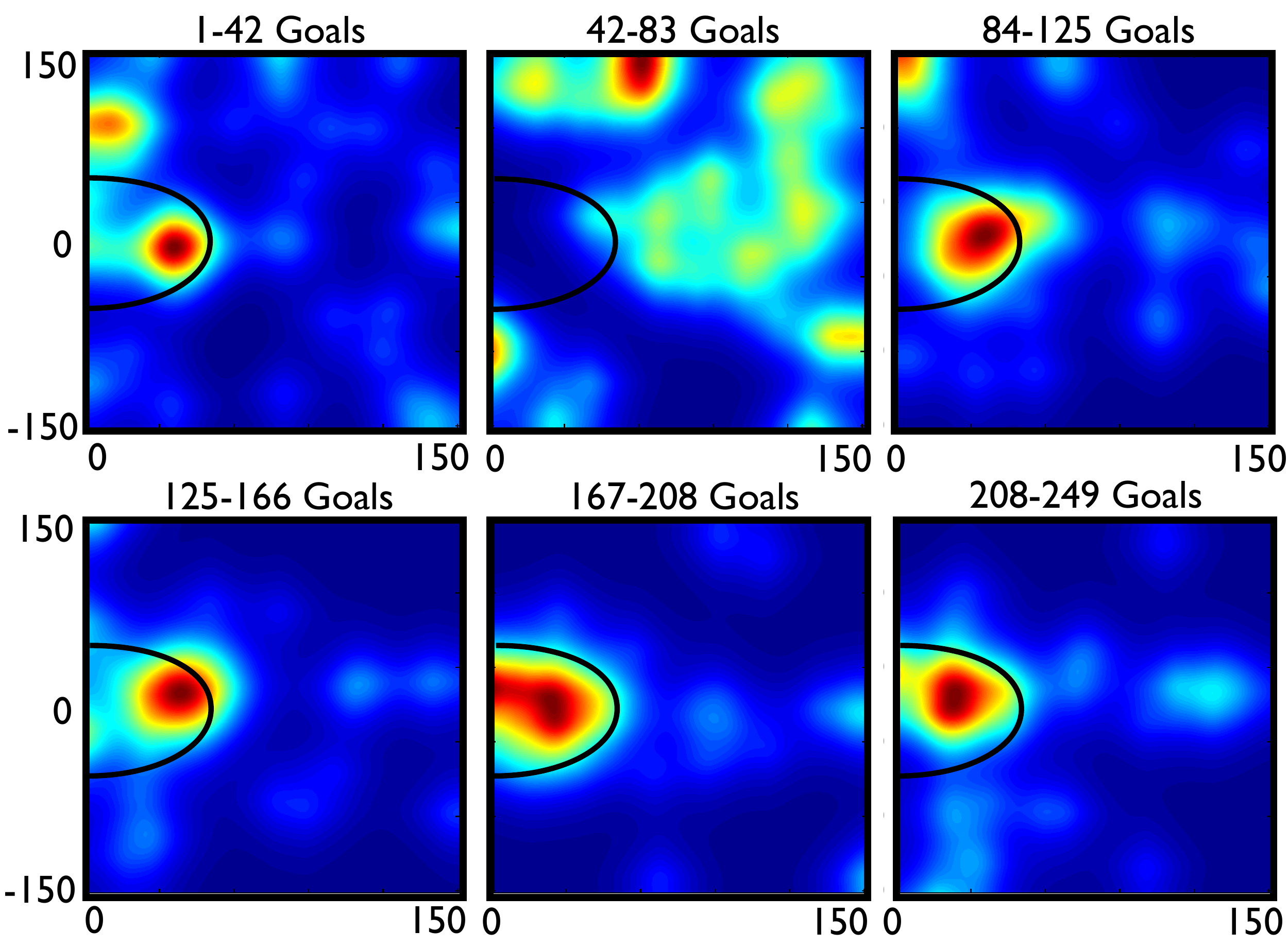}}
\caption{Evolution of the distribution of self-generated goals displayed over time windows indexed by the number of performed goals, for an experiment of 30000 micro-actions on a 15 DOF arm measuring 50 units. The black half-circle represents the contour of the area reachable by the arm. Higher values (dark red) corresponds to higher density of self-generated goals. (For interpretation of the references to color in this figure legend, the reader is referred to the web version of this article)}
\label{FigureEvolution}
\end{figure*}

\begin{figure*}[t]
\centerline{\includegraphics[width=0.7\textwidth]{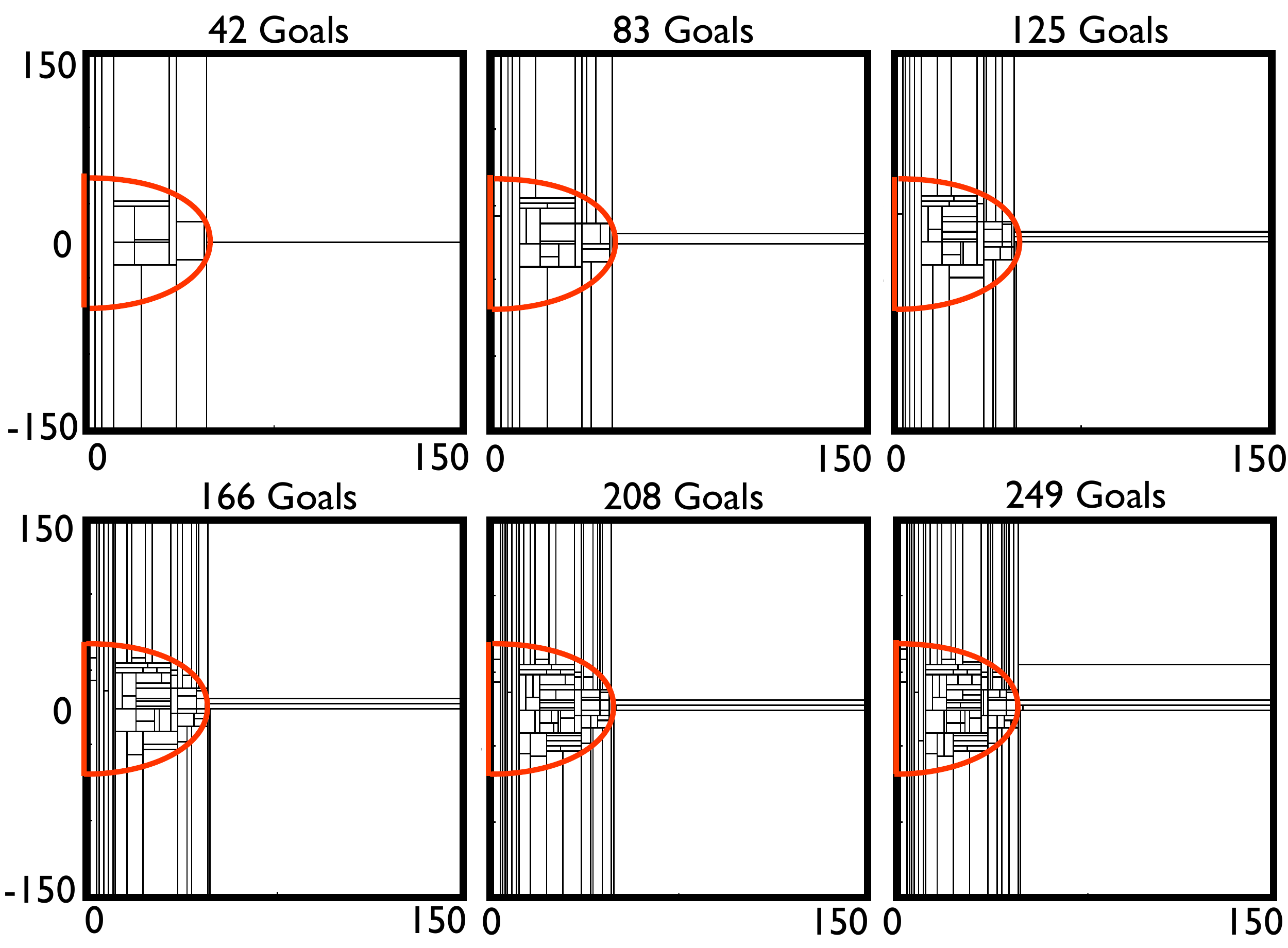}}
\caption{Evolution of the splitting of the task/goal space and creation of subregions indexed by the number of goals self-generated (without counting subgoals), for the experiment presented in Fig. \ref{FigureEvolution}.}
\label{GoalsExplorationRegions}
\end{figure*}

\begin{figure*}[t]
\centerline{\includegraphics[width=0.7\textwidth]{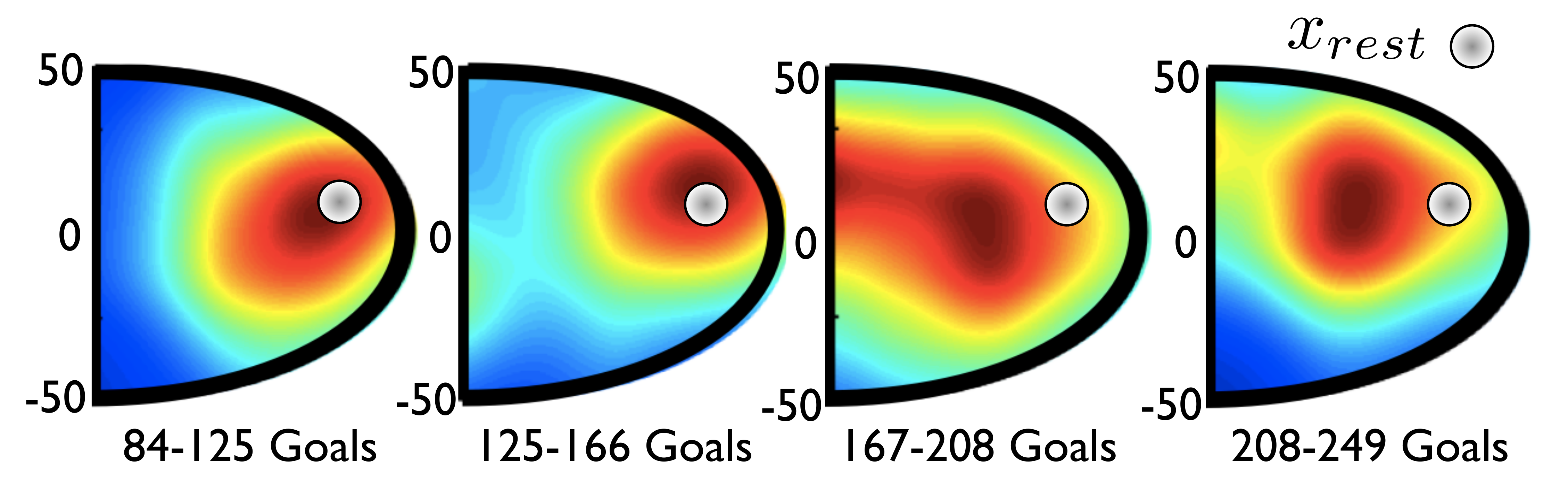}}
\caption{Details of the evolution of the distribution of self-generated goals inside the reachable area for the experiment presented in Fig. \ref{FigureEvolution}. Gray points represent the end-effector rest position $y_{rest}$.}
\label{FigureEvolutionZoom}
\end{figure*}

\subsubsection{Evolution of Competences over Time}
Fig. \ref{FigureCompetenceAll} represents the whole distribution of self-generated goals and sub-goals selected by the higher-level of active learning module, and their corresponding competences after the execution of 30000 micro-actions. The global shape of the distribution of points allows observing the large values of competence levels inside the reachable space and its close vicinity, and the global low competence inside the remaining space. 

The progressive increase of competences is displayed on Fig. \ref{FigureCompetence} where we evaluate over time (indexed here by the number of goals self-generated) the global competence of the system to reach positions situated on a grid which covers the entire task space. From these estimations of competence, we can extract two interesting phenomena: first of all, the two first subfigures, estimated after the self-generation of 42 and 83 goals, show that the system is, at the beginning of the exploration and learning process, competent to only attain areas situated close to the limits of the reachable space. Then, the 4 other subfigures show the progressive increase of competences inside the reachable space following an increasing radius whose the origin is situated around the end-effector rest position.

The first observation is due to the reaching mechanism in itself, which, when possessing a few data acquired, does not allow the robot to experiment complex joint movements, but only simple ones which typically leads to the limits of the arm. The second phenomenon is due to the coupling of the lower-level of active learning inspired by SSA with the heuristic of returning to $y_{rest}$ every subsequent goals. Indeed, the necessity to be confident in the local model of the arm to shift toward new positions makes the system progressively explore the space, and resetting it to its rest position makes it progressively explore the space by beginning close to $y_{rest}$. Finally, goal positions that are physically reachable but far from this radius typically present a low competence to be reached initially, before the radius spreads enough to reach them.

\subsubsection{Global Exploration over Time}
Fig. \ref{FigureEvolution} shows histograms of goal positions self-generated during the execution of the 30000 micro-actions (only goals, not subgoals for an easy reading of the figure). Each subfigure corresponds to a specified time window indexed by the number of generated goals: the first one (upper-left) shows that, at the onset of learning, the system already focuses in a small area around the end-effector's rest position, and thus discriminates differences between a subpart of the reachable area and the remaining space (the whole reachable zone being represented by the black half-circle on each subfigure of Fig. \ref{FigureEvolution}). In the second subfigure, the system is, inversely, focusing almost only on regions of the space which are not reachable by the arm. This is due to the imprecise split of the space at this level of the exploration, which left very small reachable areas (which have already been reached with a high competence), at the edge inside each large unreachable regions. This typically gives a high mean competence to each of these region when they are created. Then, due to the very large part of unreachable areas, in comparison to reachable ones, the mean competence decreases over time. This brings interest to the region, thanks to the mathematical definition of the interest level, which, by using an absolute value, pushes the robot toward areas where the competence is decreasing. This complex process which allows driving the exploration in these kind of heterogeneous regions then allows dividing efficiently the task space into reachable and unreachable regions. 

Then, considering a global observation of subfigures 3 to 6, we can conclude that the system effectively autonomously discovers its own limits by focusing the goal self-generation inside reachable areas during the largest part of the exploration period. The system is indeed discovering that only a subpart is reachable due to the interest value becoming null in totally unreachable areas where the competence value is low.

\subsubsection{Exploration over Time inside Reachable Areas}
A more precise observation of subfigures 3 to 6 is presented in Fig. \ref{FigureEvolutionZoom} where we can specifically observe the self-generated goals inside the reachable area. First, we can perceive that the system is originally focusing in an area around the end-effector's rest position $y_{rest}$ (shown by gray points in Fig. \ref{FigureEvolutionZoom}). 

Then, it increases the radius of its exploration around $y_{rest}$ and focuses on areas further afield to the end-effector's rest position. Subfigures 2 and 3 shows that the system explores new reachable parts corresponding to the right part close to its basis (subfigure 2), and then, the left part close to its basis (subfigures 3).

Also, comparing the two first subfigures, and the two last ones, we observe a shift of the maximum exploration peak toward the arm basis. This is first linked with the loss of interest of self-generating goals around the end-effector's rest position. Indeed, because the system becomes highly efficient inside this region, the competence level becomes high and stationary over time, which leads to low interest values. At the same time, this phenomenon is also linked with the increase of competences in new reachable positions far from the end-effector rest position $y_{rest}$, closer to its basis, which creates new regions of interest (see the four last subfigures of Fig. \ref{FigureCompetence}).

\subsubsection{Emergent Process}
The addition of subgoals and the consideration of each end-effector's position as a goal reached with the highest competence level have important influences on the learning process. If we look at traditional active learning algorithms which cannot deal with open-ended learning \cite{Cohn96, Freund97, Dasgupta04}, as well as RIAC-like algorithms different from SAGG-RIAC \cite{Oudeyer07, Marshall04, Schembri07b, Mugan09, Schmidhuber91, Baranes09}, we can notice that even if these techniques deal with avoiding excessive exploration in unlearnable or extremely complex areas, the learning process still has to begin by a period of random exploration of the whole space, to distinguish and extract which subparts are the most \textit{interesting} according to the used definition of interest. Thanks to the addition of sub-goals and/or the consideration of every end-effector's position in SAGG-RIAC, in addition to exploring in the task space, we reduce the number of needed random global exploration, and improve the capability of the system to deal with large (i.e. when the volume of reachable space is small as compared to the volume of the whole space) task spaces. Using subgoals indeed creates a concentration of goals around the current end-effector's position, which progressively grows according to new experimented positions. 

\label{feedback}
Furthermore, the consideration of each end-effector's position for the estimation of competence allows discovering progressively which positions are reachable with a high competence level, and gives a fast indication of first subregions where these high competences are situated. This increases the number of subregions close to the reachable areas and allows computing the interest values in the growing vicinity of the end-effector's experimented positions (see Fig. \ref{GoalsExplorationRegions} where the progressive split of subregions in reachable areas is displayed). 

Therefore, these additions of competence measures allow the system to discover and focus on areas where the competence is high in a very low number of goal self-generation, and tackle the typical problem of fast estimation and distinction of interesting areas. Nevertheless, this emergent process only helps to increase the number of feedbacks required by the goal self-generation mechanism to split the space, and do not influence the low-level active learning. Then, the timeout which defines a goal as unreached during a single reaching attempt becomes crucial when considering high-volume task spaces with large unreachable parts as introduced in the following section.

\begin{figure*}[t!]
\centerline{\includegraphics[width=0.7\textwidth]{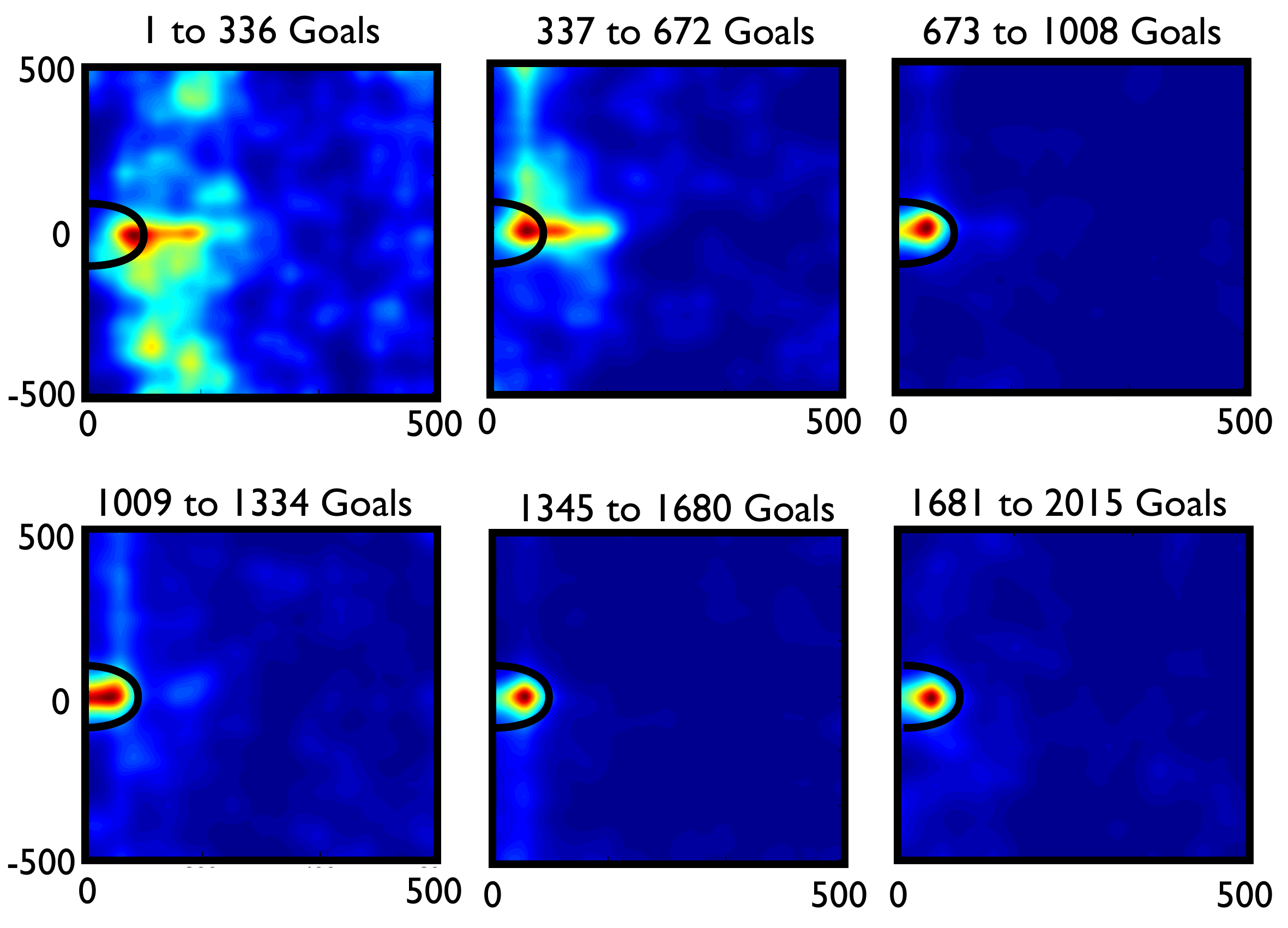}}
\caption{Histograms of self-generated goals displayed over time windows indexed by the number of performed goals, for an experiment of 30000 micro-actions on a 15 DOF arm, for a high-volume task space $S = [-500;500] \times [0;500]$, according to the reachable space contained in $[-50;50] \times [0;50]$ (the black half-circle represents the contour of the area reachable by the arm according to its length of 50 units).}
\label{RIAC10Times}
\end{figure*}

\subsubsection{Robustness in High-Volume Task Spaces}
\label{blocking}
in the previous experiment, the timeout which describes a goal as not reached and stops a reaching attempt is defined as directly proportional to the number of micro-actions required to reach each goal. Practically, as introduced section \ref{sectionTimeout}, we allowed the system to perform $1.5$ times the distance between $y_{start}$ and $y_g$ before declaring a goal as not reached (including explorative movements). 

This timeout is efficient enough to learn efficiently by discriminating regions of different complexities in the middle-size space $S ' = [0;150] \times [-150;150]$ considered in this experiment. Nevertheless, it can have an important influence on the SAGG learning process when considering extremely large task spaces with small underlying reachable areas. For instance, if we consider a task space $Y = [-500;500] \times [-0;500]$ where only $[-50;50] \times [0;50]$ is reachable, the low-level of active learning will spend an extremely large number of iterations trying to reach each unreachable goal if this kind of timeout is used.

Therefore, when considering such high-volume spaces, the definition of a new timeout becomes crucial. In Fig. \ref{RIAC10Times}, we demonstrate the high discriminating factor of SAGG-RIAC in such a task space ($Y = [-500;500] \times [-0;500]$) when using a timeout which is not only based on the distance to the goal. This one has also been designed to stop a reaching attempt according to the following blocking criteria: let us consider a self-generated goal $y_g$ that the low-level exploration and reaching mechanisms try to reach. Then, if the system is not coming closer to the goal even after some low-level explorations, the exploration toward this precise goal stops. In a practical way, when $w$ consecutive low-level explorations are triggered (typically $w \geq 2$) and thus no progress to the goal was made, we declare a goal as unreached, and compute the corresponding competence level. Using such a definition, the rapidity of discovering blocking situations will depend on both values of $w$ and number of explorative actions $q$. Minimal values of these two parameters allows the fastest discoveries, but decrease the quality of the low-level exploration mechanism when exploring reachable spaces (in the experiment presented in Fig. \ref{RIAC10Times} we use $q = 5$ and $w = 3$).

\subsubsection{Conclusion of Qualitative Results}
When considering low-level mechanisms allowing an efficient progressive learning, the SAGG-RIAC algorithm is capable to discriminate very efficiently reachable areas in such high-volume spaces. Then, it is also able to drive a progressive self-generation of goals through reachable subspaces of progressively growing complexities of reachability. 

In this experiment, the reachable region in the task space was convex and with no obstacles. Yet, as we will see in the fishing experiment below, SAGG-RIAC is capable of identifying correctly its zones of reachability, given a low-level optimization algorithm, even if there are ``holes'' or obstacles: goals initially generated in unreachable positions or in positions for which obstacles prevent their reaching provide a low level of competence progress, and thus the system stops trying to reach them. It is also possible to imagine that some given self-generated goals might be reachable only by an action policy going \textit{around} an obstacle. Such a capability is not a property of the SAGG-RIAC architecture by itself, but a property of the optimization algorithm, and action representation, that is used at the low-level goal-directed mechanism. In the present experiment, low-level optimization was a simple one only considering action policies going in a straight line to the goal. Yet, if one would have used more complex optimization leveraging continuous domain planning techniques (e.g. \cite{Toussaint06}), the zones of reachability would be increased if obstacles are introduced since the low-level system could learn to go around them.


\subsection{Quantitative Results for Experiments \\ with Task Spaces of Different Sizes}
 \label{maxSection}
In the following evaluation, we consider the same robotic system than previously described (15DOF arm of 50 units) and design different experiments. For each one, we estimate the efficiency of the inverse model learned by testing how it allows in average the robot to reach positions selected inside a test database of 100 reachable positions (uniformly distributed in the reachable area and independent from the exploration of the robot).
We will also compare SAGG-RIAC to three other types of exploration techniques: 
\begin{enumerate}
\item SAGG-RANDOM, where goals are chosen randomly (higher-level of active learning (RIAC) disabled)
\item ACTUATOR-RANDOM, where small random micro-actions $\Delta \alpha$ are executed. This method corresponds to classical random motor babbling.
\item ACTUATOR-RIAC, which corresponds to the original RIAC algorithm that uses the decrease in prediction errors $(\alpha, \Delta \alpha) \rightarrow \Delta x$ to compute an interest value and split the space $(\alpha, \Delta \alpha)$.
\end{enumerate}
Also, to be comparable to SAGG-RIAC, each ACTUATOR technique will have the position of the arm reset to the \textit{rest position} every $max$ micro-actions, $max$ being the number of micro-actions needed to reach the more distant reachable position. $max$ is proportional to the desired velocity which is here of $v = 2$ units/micro-action as well as the size of the task space (this will explain the different results of each ACTUATOR methods when used with task spaces of different sizes). In every graph, we present statistical results obtained after launching the same experiment with different random seeds 15 times.

\subsubsection{Exploration in the Reachable Space}
The first quantitative experiment is designed to compare the quality of inverse models learned using babbling in the task/operational space (i.e. using goals), instead of more traditional motor babbling heuristics executed in the configuration/actuator space. We still consider a $n$=15 DOF arm of 50 units, also, to be suited for the first study, dimensions of $Y$ will be bounded in intervals $y_g \in [0;50] \times [-50;50]$ which means that the arm can reach almost all the space $Y$ where goals can be chosen (the limits of reachability are thus almost given to the robot). In this experiment, we fix $q = 20$ for the SAGG methods and use a timeout only relative to the distance to the current goal (a end-effector movement of $1.5$ times the one needed is allowed). 

\begin{figure}[t]
\centerline{\includegraphics[width=0.9\linewidth]{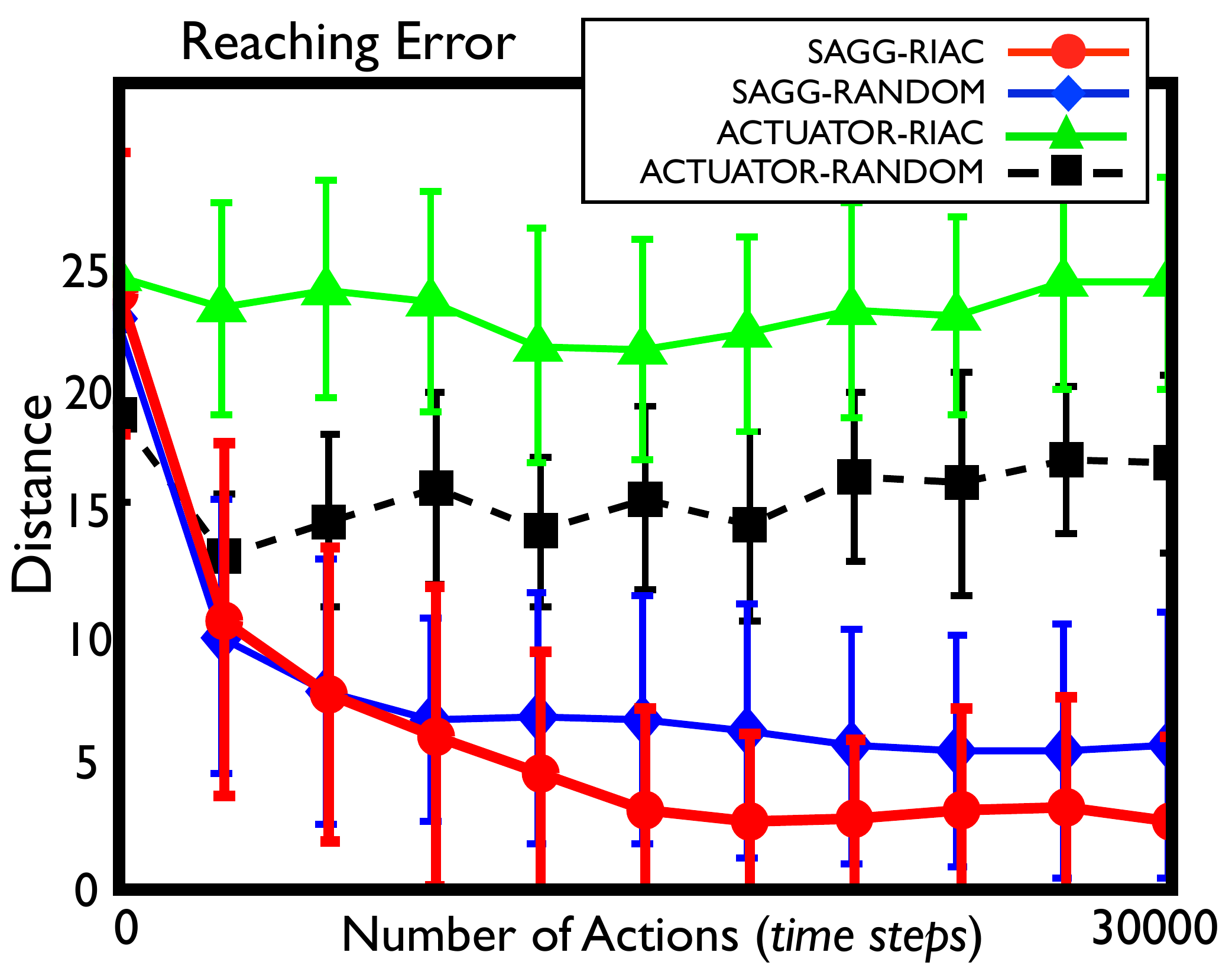}}
\caption{Evolution of mean distances between the goal and the end effector after reaching attempts over an independently randomly generated set of test goals. Here SAGG-RIAC and SAGG-RANDOM are only allowed to choose goals within $Y = [0;50] \times [-50;50]$ (i.e. most eligible goals are physically \textit{reachable}). Standard deviations are computed over 15 experiments at the same instants for each curve, and shifted in graphs for an easy reading.}
\label{FigCurvesMult1}
\end{figure}

\begin{figure}[t]
\centerline{\includegraphics[width=0.9\linewidth]{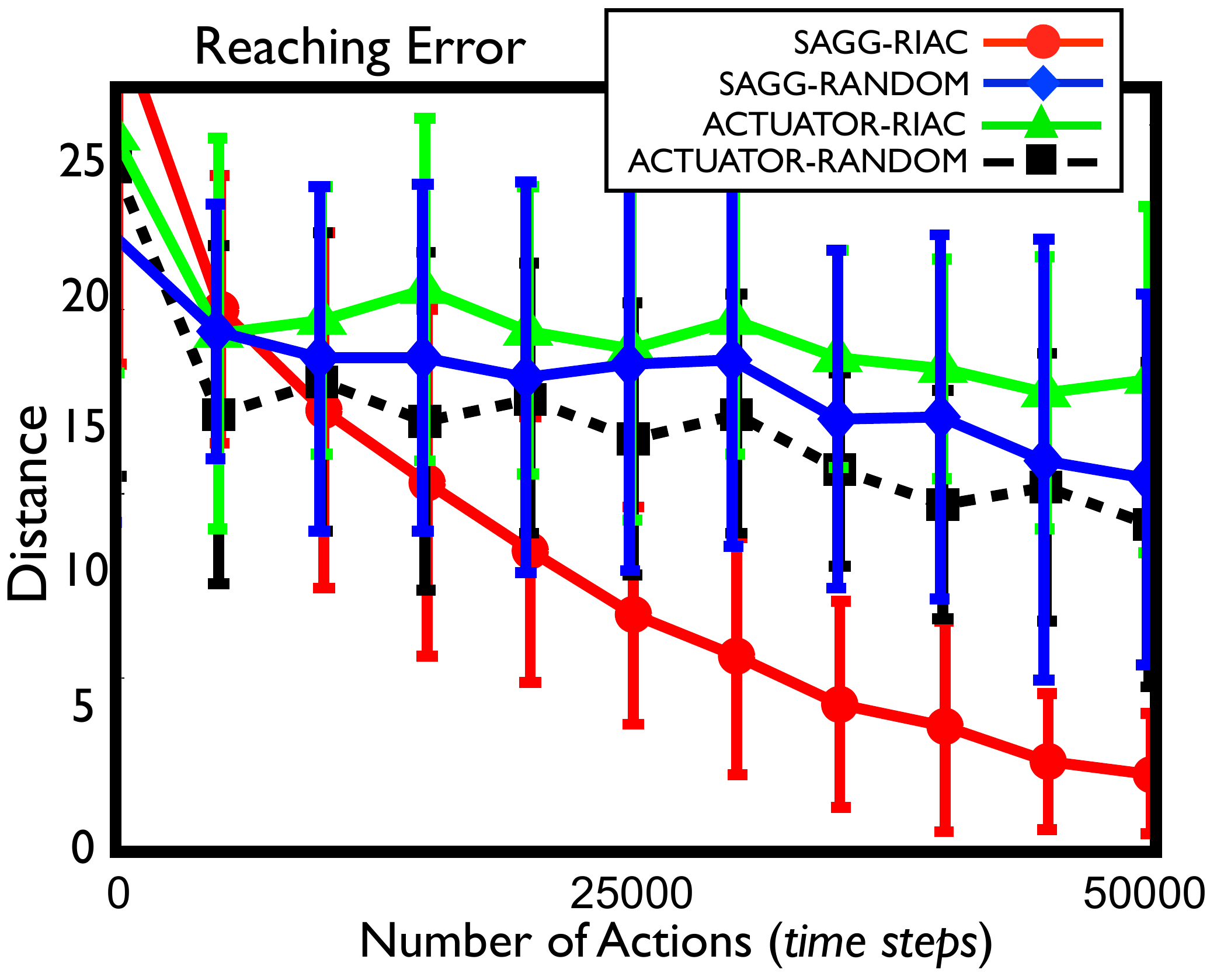}}
\caption{Evolution of mean distances between the goal and end effector after reaching attempts over an independently randomly generated set of test goals, averaged over 15 experiments. Here SAGG-RIAC and SAGG-RANDOM are allowed to choose goals within a large space corresponding to the one in Fig. \ref{RIAC10Times}, define as $Y = [0;500] \times [-500;500]$ (i.e. most eligible goals are physically \textit{unreachable}).}
\label{FigCurvesMult1bis}
\end{figure}

Fig. \ref{FigCurvesMult1} shows the evolution of the capability of the system to reach the 100 test goals using the inverse model learned by each technique, starting from the rest position. This capability is computed using the mean Euclidian distance between the goal and the final state of a reaching attempt.


Globally, these results show that in order to learn inverse kinematics of this highly-redundant arm, exploration in the goal/operational space is significantly more efficient than exploration in the actuator space using either random exploration or RIAC-like active learning. Moreover, better performances of ACTUATOR-RANDOM compared to ACTUATOR-RIAC emphasizes that the original version of RIAC has not been designed for the efficient learning of inverse models of highly-redundant systems (high-dimension in the actuator space).

Focusing on the evaluation of the two mechanisms which use SAGG, we can also make the important observation that SAGG-RIAC is here more efficient than SAGG-RANDOM when considering a system which already knows its own limits of reachability. More precisely, we observe both increase in learning speed and final  generalization performances (this results resonates with results from more classic active learning, see \cite{Cohn94}). These improvement signifies that SAGG-RIAC is efficiently able to progressively discriminate and focus on areas which bring the highest informational amount (i.e. areas which have not been visited enough). It brings to the learning system more useful data to create an efficient inverse model, contrarily to the SAGG-RANDOM approach which continues to select goals in already efficiently reached areas.

\subsubsection{Robustness in Large Task Spaces}
in the following experiment, we would like to test the capability of SAGG-RIAC to focus on reachable areas when facing high volume task spaces (will call this phenomenon the discrimination capability). Therefore, we will here consider a task space $Y = [0;500] \times [-500;500]$. Fig. \ref{FigCurvesMult1bis} shows the learning efficiency of SAGG-RIAC using the timeout with blocking criteria as described in the section \ref{blocking}. This allows to test the quantitative aspect of the discrimination capability of SAGG-RIAC and its comparison with the three other techniques when facing high volume task spaces where only small subparts are reachable. As Fig. \ref{FigCurvesMult1bis} shows, SAGG-RIAC is here the only method able to drive an efficient learning in such a space. SAGG-RANDOM actually spends the majority of the time trying to reach unreachable positions. Also, the size of the task space has an influence on the two ACTUATOR algorithms if we compare results in $Y = [0;50] \times [-50;50]$ introduced Fig. \ref{FigCurvesMult1} and in $Y = [0;500] \times [-500;500]$ introduced Fig. \ref{FigCurvesMult1bis}. This is due to the value $max$ of micro-actions performed by ACTUATOR methods which is proportional to the size of the task space as explained section \ref{maxSection}. Results considering the space $Y = [0;500] \times [-500;500]$ seems more efficient for these methods, where the value of $max$ is higher than in $Y = [0;50] \times [-50;50]$. An increase of $max$ thus allows these methods to explore more efficiently the reachable space whose exploration is limited when considering a too low value of $max$.

\subsubsection{Robustness in Very Large Task Spaces}
\begin{figure}[h!]
\centerline{\includegraphics[width=0.9\linewidth]{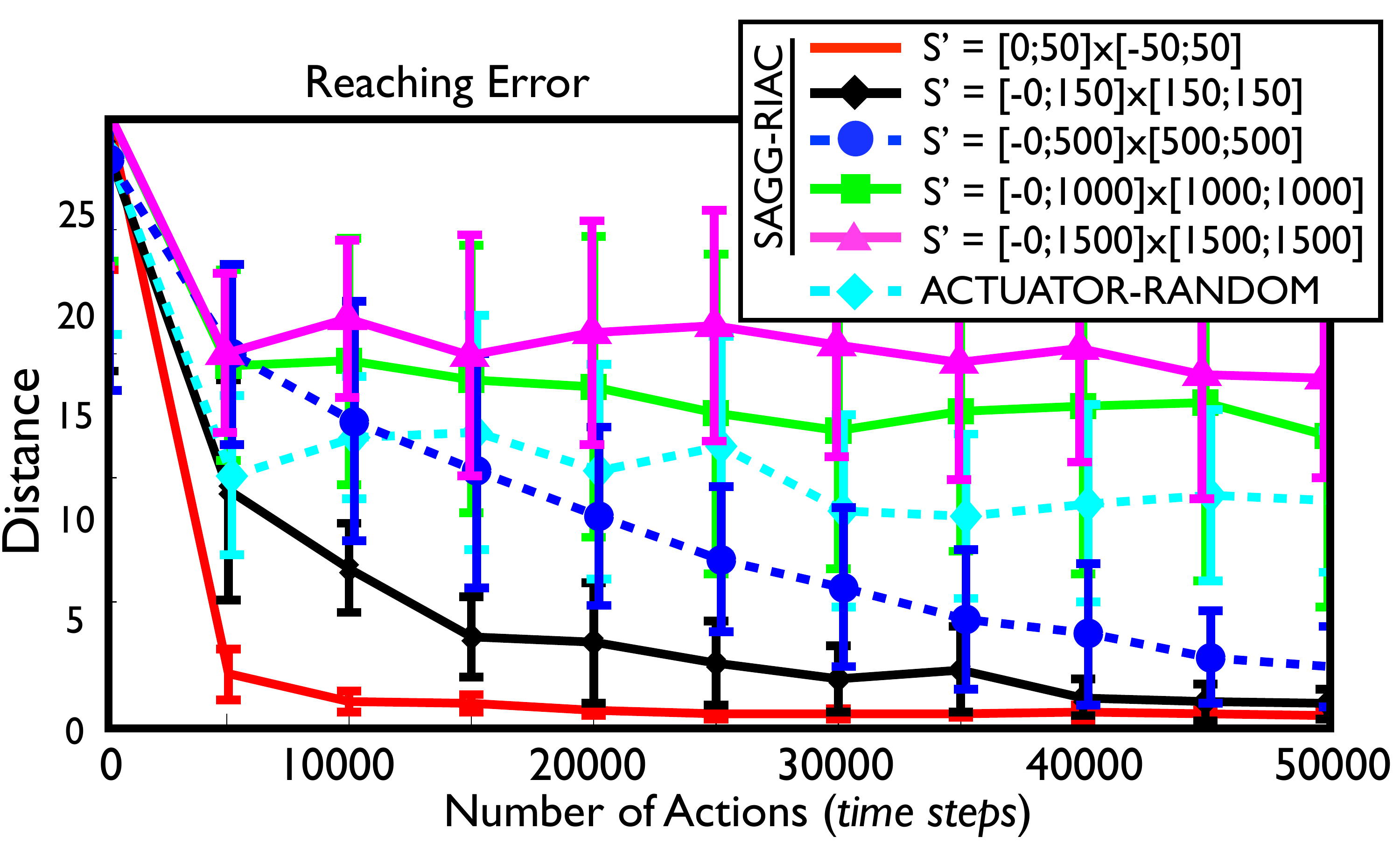}}
\caption{Quantitative results of SAGG-RIAC when used with task spaces of different sizes and comparison with ACTUATOR-RANDOM.}
\label{BeforeConstraints}
\end{figure}

Finally, we test the robustness of SAGG-RIAC in task spaces larger than in the previous section. Fig. \ref{BeforeConstraints} shows the behavior of SAGG-RIAC when used with task spaces of different sizes, from 1 to 900 times the size of the reachable space, and compare these results with a random exploration in the actuator space when the value of $max$ is fixed as when $Y = [0;500] \times [-500;500]$. We can notice here that, although the high discriminative capacity of SAGG-RIAC in large spaces such as $Y = [0;500] \times [-500;500]$, as shown previously, the performances of this technique decrease when the size of the considered task space increases. Therefore, we can observe that SAGG-RIAC obtains better results than ACTUATOR-RANDOM since 5000 micro-actions when considering spaces smaller than $Y = [0;500] \times [-500;500]$. Then, this method shows better results than ACTUATOR-RANDOM only after 10000 micro-actions when considering the space $Y = [0;500] \times [-500;500]$. And finally, this one becomes less efficient than ACTUATOR-RANDOM when the considered space increases in comparison to the reachable space, as shown by results when considering spaces $Y = [0;1000] \times [-1000;1000]$ and $Y = [0;1500] \times [-1500;1500]$. These results clearly show that SAGG-RIAC is robust in spaces up to 100 times larger than the reachable space, but has some difficulties to explore even larger spaces. Therefore, despite the fact that SAGG-RIAC is very efficient in large spaces, it seems that the challenge of autonomous exploration in un-prepared spaces can not be totally resolved by this algorithm, a human supervisor being still necessary to define a set of (even very approximate) limits for the task space. As it will be emphasized in the perspective of this work, some complementary techniques should be used in order to bring robustness to such spaces, such as mechanisms inspired by the notion of maturational constraints which are able to fix limits on the task space since the beginning of the exploration process.

\subsection{Quantitative Results for Experiments \\ with Arm of Different Number of DOF and Geometries}

In every experiment, we set the dimensions of $Y$ as bounded by the intervals $y_g \in [0;150] \times [-150;150]$, where 50 units is the total length of the arm, which means that the arm covers less than $1/9$ of the space $Y$ where goals can be chosen (i.e. the majority of areas in the operational/task space are not reachable, which has to be discovered by the robot). 

For each experiment, we set the desired velocity $v = 0.5$ units/micro-action, and the number of explorative actions $q = 20$. Moreover, we reset the arm to the rest position $(\alpha_{rest}, y_{rest})$ every $r = 2$ reaching attempts, which increases the complexity of the reaching process. 

\begin{figure*}[t!]
\centerline{\includegraphics[width=0.7\textwidth]{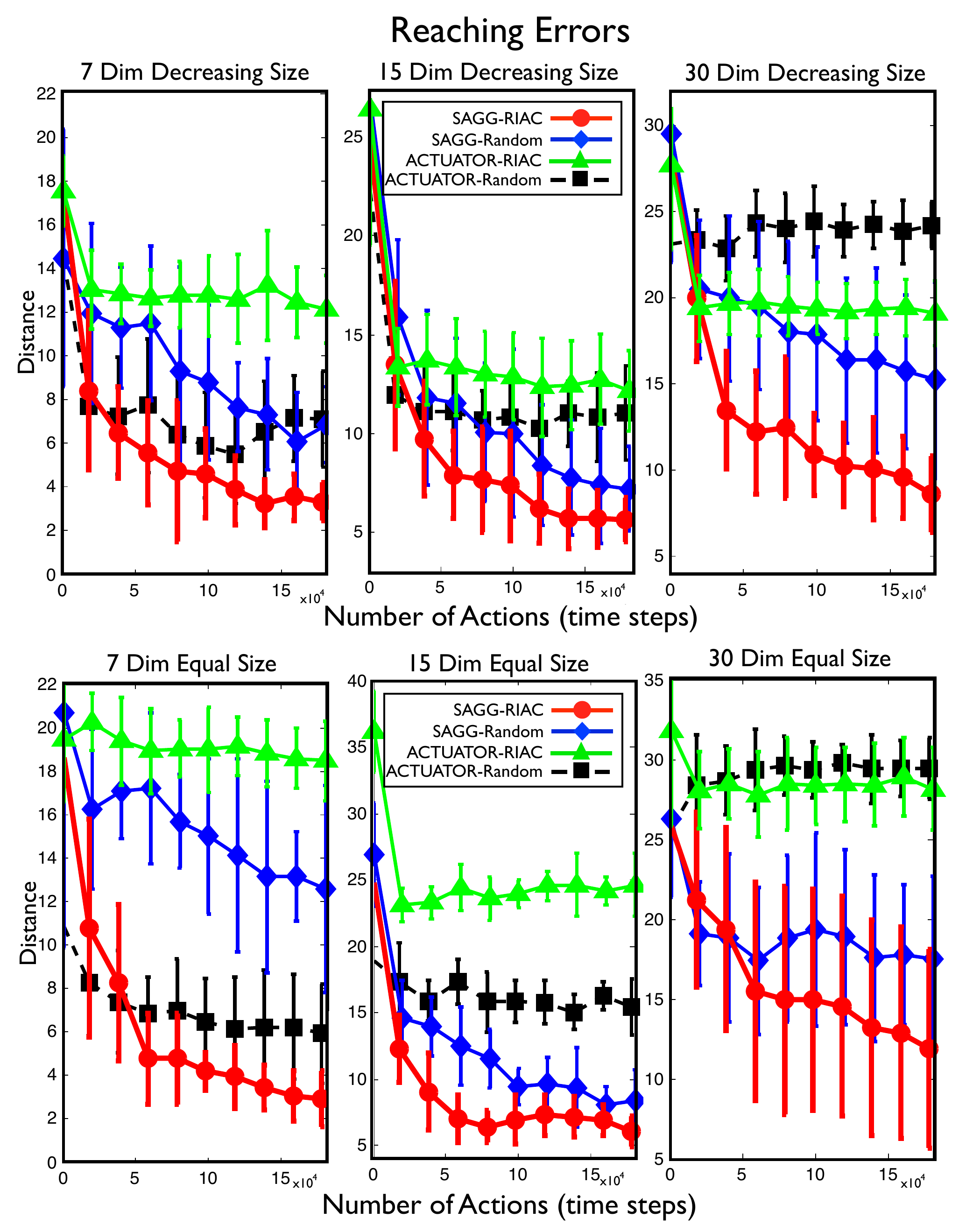}}
\caption{Evolution of mean distances between the goal and end effector after reaching attempts over an independently randomly generated set of test goals, averaged over 15 experimentations. Here SAGG-RIAC and SAGG-random are only allowed to choose goals within $Y = [0;150] \times [-150;150]$ (i.e. the set of reachable goals is only a small subset of eligible goals).}
\label{FigCurvesMult3}
\end{figure*} 

We present a series of experiments aiming to test the robustness of SAGG-RIAC in arm setups with different shapes and numbers of degrees-of-freedom. Performed tests used 7, 15, and 30 DOF arms whose each limb has either the same length or a decreasing length depending on its distance from the arm's base (we use the golden number to specify the relative size of each part, taking inspiration from the architecture of human limbs). These experiments permit testing the efficiency of the algorithm for highly redundant systems (considering a 30 DOF arm corresponds to a problem of 62 continuous dimensions, with 60 dimensions in the actuator/state space and 2 dimensions in the goal/task space), and different morphologies.

Also, to stress the capability of the system to make the robot self-discover its own limits, we remove the consideration of each end-effector position experimented as a goal reached with the highest level of competence (see \ref{feedback}). In these experiments, the competence level is therefore evaluated only for goals and subgoals. We fix $q = 100$, and compute tests of inverse models over 200000 micro-actions.

\subsubsection{Quantitative Results}
Fig. \ref{FigCurvesMult3} illustrates the performances of the learned inverse models when used to reach goals from an independent test database and evolving along with the number of experimented micro-actions. First, we can globally observe the slower decreasing velocity (over the number of micro-actions) of SAGG-RANDOM and SAGG-RIAC, compared to the previous experiment, which is due to the higher value of $q$ and the removed consideration of every end-effector position. Graphs on the first line of Fig. \ref{FigCurvesMult3} present the reaching errors of 7, 15 and 30 DOF arms with decreasing lengths. The first subfigure shows that when considering 7 DOF, which is a relatively low number of degrees of freedom, SAGG-RANDOM is not the second more efficient algorithm. Indeed, the ACTUATOR-RANDOM method is here more efficient than SAGG-RANDOM after 25000 micro-actions and is then stabilized, while SAGG-RANDOM is progressively decreasing, reaching the same level as ACTUATOR-RANDOM at the end of the experiment. This is due to the high focalization of SAGG-RANDOM outside the reachable area, which leads to numerous explorations toward unreachable positions. As shown also in this subfigure, adding the RIAC active component to SAGG efficiently improves the learning capabilities of the system; SAGG-RIAC reaching errors were indeed the lowest for this 7 DOF system. 

Experiments with 15 DOF and 30 DOF shows that both SAGG methods are here more efficient than actuator methods, SAGG-RIAC showing a significant improvement compared to every other algorithm (for 15DOF, the level of significance is $p = 0.002$ at the end of the experiment (200000 micro-actions)).

Experiments presented with 7, 15 and 30 DOF arms where each limb has the same length show the same kind of results. The 7 DOF experiment shows that ACTUATOR-RANDOM can be more efficient than SAGG-RANDOM, and that the addition of RIAC allows obtaining a significant improvement in this case, but also when considering 15 and 30 DOF. 

\subsubsection{Conclusion of Quantitative Results}
Globally, quantitative results presented here emphasize the high efficiency and robustness of SAGG-RIAC when carried out with highly redundant robotic setups of different morphologies, compared to more traditional approaches which explore in the actuator (input) space. They also showed that random exploration in the goal (output) space can be very efficient when used in high-dimensional systems, even when considering a task space more than 9 times larger than the reachable subspace. 
These results therefore indicate the high potential of \textbf{competence based motor learning} for IK learning in highly-redundant robots.

\subsection{Qualitative Results for a Real 8 DOF Arm}

\begin{figure*}[t!]
\centerline{\includegraphics[width=0.7\textwidth]{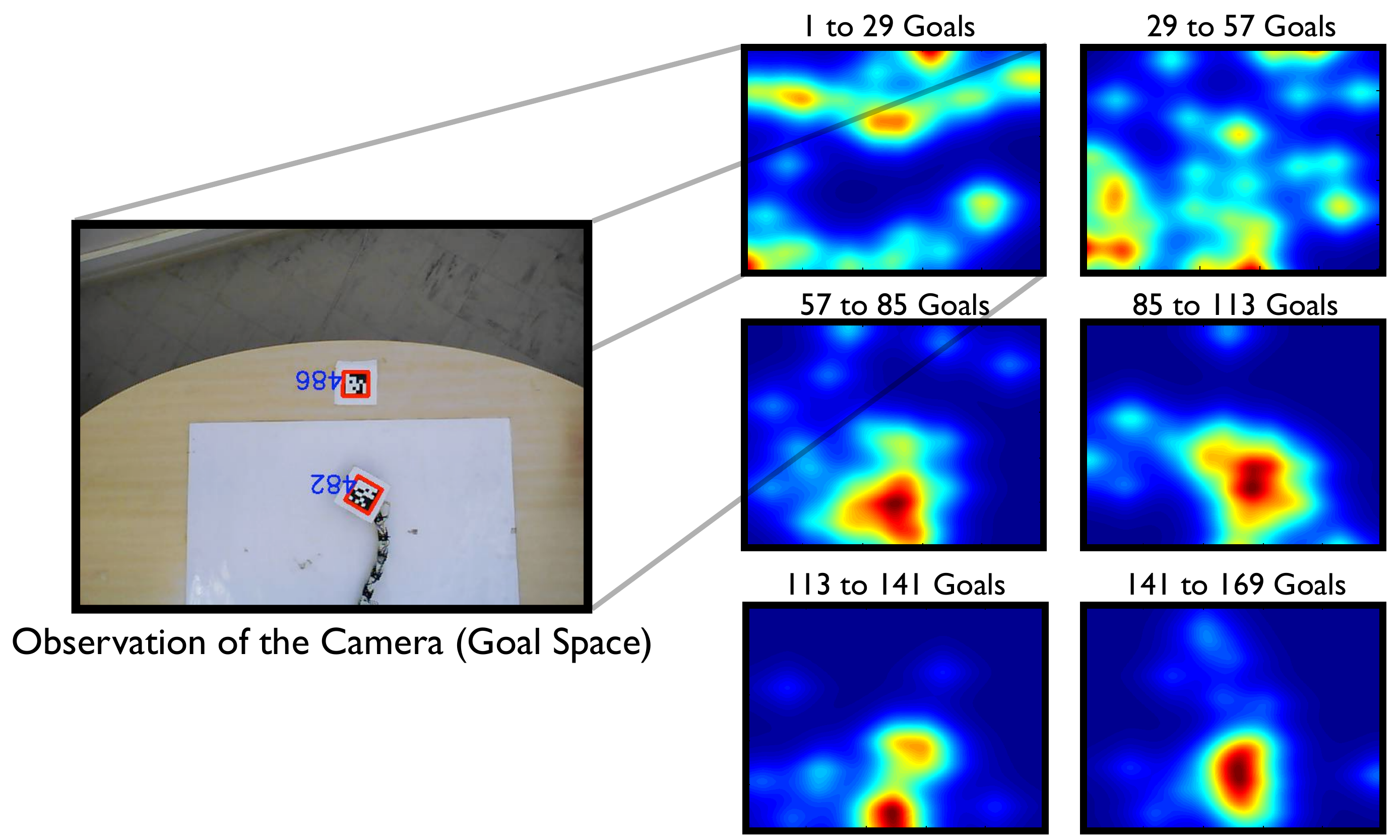}}
\caption{Histograms of self-generated goals displayed over time windows indexed by the number of performed goals, for an experiment of 10000 micro-actions on a real 8 DOF arm. Each histogram represents the surface covered by the camera, which here defines the task space.}
\label{ExplorationReal}
\end{figure*}

In this section, we test the robustness of the algorithm in a qualitative point of view when considering a real robotic setup (not simulated) which corresponds to the simulation presented above: we use a 8 DOF arm controlled in position. Also, helping to test the robustness of our method, we use low quality motors whose averaged noise is $20\%$ for each movement. The fixed task space corresponds to the whole surface observable by a camera fixed on top of the robot, which is more than three times larger than the reachable space (see the left part of Fig. \ref{ExplorationReal}). In order to allow the camera to distinguish the end-effector of the arm and to create a visual referent framework on the 2D surface, we used visual tags and the software ARToolKit Tracker \cite{Artoolkit00}.

Fig. \ref{ExplorationReal} (right part) shows histograms of self-generated goals displayed over sliding time windows indexed by the number of performed goals (without counting subgoals) for an experiment of 10000 micro-actions. We can observe that the algorithm manages to discover the limits of the reachable area and drives the exploration inside after the goal 57. Then, the system continues to focus on the reachable space until the end of the experimentation, alternating between different areas inside. More precisely, we can notice while comparing the bottom-left subfigure to the two positioned on the second line, that the system seems to concentrate only after some time on the areas situated close to its basis, and therefore more difficult to reach. The progressive increase of the complexity of positions explored which appeared in simulation therefore also happens here. Finally, the last subfigure shows that the system continues its exploration toward an area more central of the reachable part. This is due to the high level of noise of the motor control: while the system is originally not very robust in this part of the space, an improvement of the generalization capacity of the learning algorithm allows obtaining an increase of competences in already visited areas of the task space. 

This experiment shows the efficiency of the SAGG-RIAC architecture to drive the learning process in real noisy robotic setups with only a few iterations, as well as its capacity to still control the complexity of the exploration when considering highly-redundant systems. 

\section{Experimental Setup 2: Learning Omnidirectional Quadruped Locomotion with Motor Synergies}

\label{LabelExp2}
Sometimes stemming from pre-wired neuronal structures (e.g. central pattern generators \cite{Grillner85, Nishii94, Ijspeert08}), motor synergies are defined as the coherent activations (in space or time) of a group of muscles. They have been proposed as building blocks simplifying the scaffolding of motor behaviors because allowing the reduction of the number of parameters needed to represent complex movements \cite{Avella03, Lee84, Berniker09, Ting07}. 
Described as crucial for the development of motor abilities, they can be seen as encoding an unconscious continuous control of muscles which simplifies the complexity of the learning process: learning complex tasks using parameterized motor synergies (such as walking, or swimming) indeed corresponds to the tuning of relatively low-dimensional (but yet which can have a few dozen dimensions) high-level control parameters, compared to the important number of degrees of freedom which have to be controlled (thousand in the human body, see \cite{Bernstein67}).

\begin{figure}[h!]
\center
\includegraphics[width=0.7\linewidth]{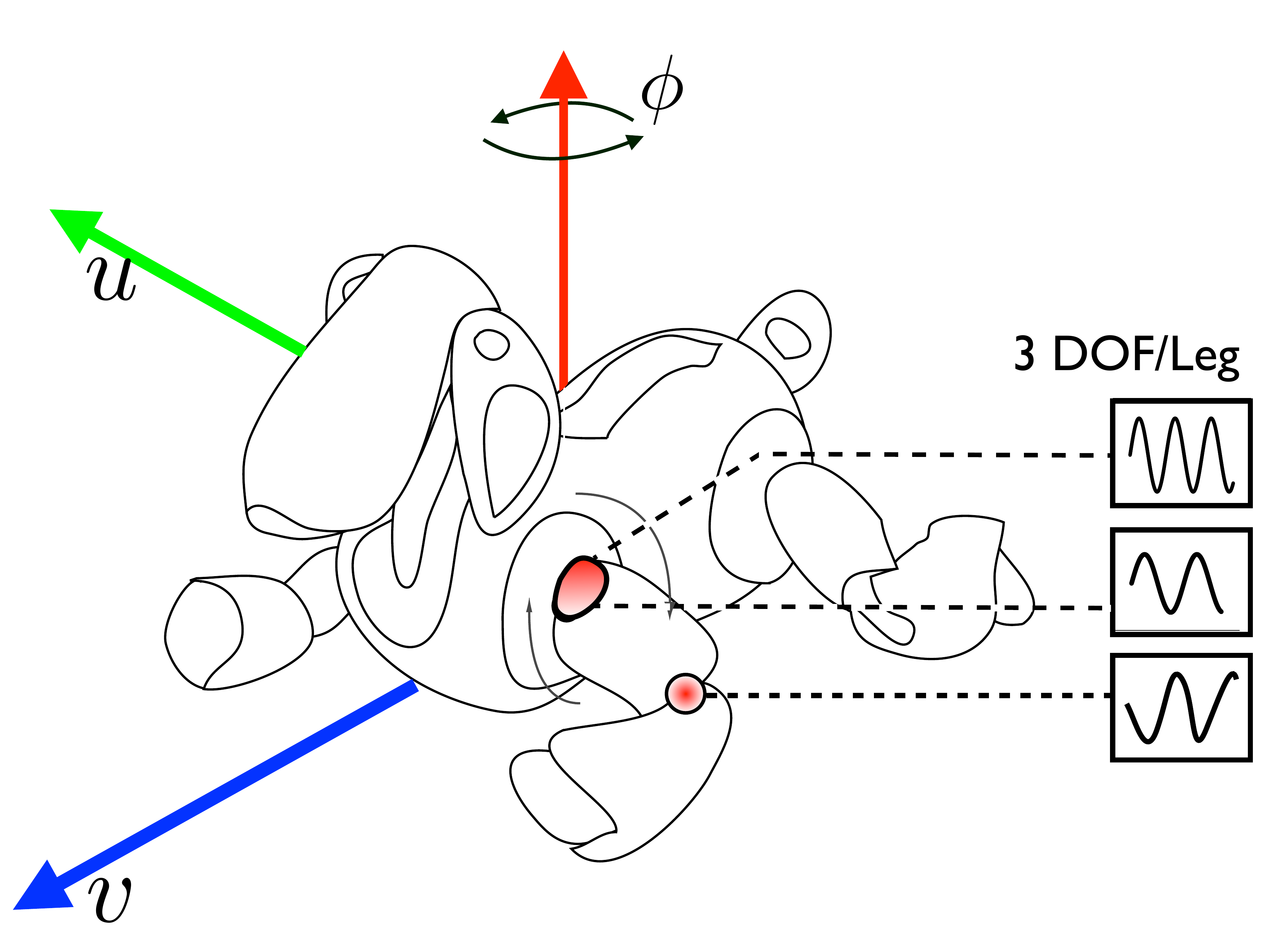}
\caption{12 degrees-of-freedom quadruped controlled using motor synergies parameterized with 24 values : 12 for the amplitudes and 12 others for the phases of a sinusoid tracked by each motor. Experiments consider a task space $u,v,\alpha$ which corresponds to the 2D position and orientation of the quadruped.}
\label{FigQuadruped1}
\end{figure}

\subsection{Formalization}
In the two following experiments, we simplify the learning process by using such parameterized motor synergies controlling amplitude, phase, and velocity of Central Pattern Generators (CPGs). Mathematically, using motor synergies simplifies the description of the considered robotic system.  In the framework introduced above (section \ref{formalization}) we defined our system as being represented by the relationship $(s, a) \rightarrow y$, where for a given configuration $s \in S$, a sequence of actions $a = \{a_1, a_2, ..., a_n\} \in A$ allows a transition toward $y \in Y$. 

In the current framework we consider the sequence of actions as being generated directly by parameterized motor synergies $\pi_\theta$, which means that the sequence of actions is directly encoded and controlled (using feedbacks internal to the synergy) by setting parameters $\theta$ specified at the beginning of an action. For instance, in the experiment described in this section, we define a synergy as a set of parameterized sinusoids (one on each joint) that a motor joint has to track with a low-level pre-programmed PID-like controller.
Eventually, motor synergies can be seen as a way to encapsulate the low-level generation of sequences of micro-actions, allowing the system to directly focus on the learning of models $(s, \pi_\theta) \rightarrow y$, with $s \in S$ fixed (the rest position of the robot) and $\theta$ a set of parameters controlling the synergy (we will remove the fixed context $s$ in the next notations for a easier reading and only write $\pi_\theta \rightarrow y$).



\subsection{Robotic Setup}

In the following experiment, we consider a quadruped robot simulated using the Breve simulator \cite{Breve} (physics simulation is based on ODE). Each of its leg is composed of 2 joints, the first (closest to the robot's body) is controlled by two rotational DOF, and the second, one rotation (1 DOF). Each leg therefore consists of 3 DOF, the robot having in its totality 12 DOF (See Fig. \ref{FigQuadruped1}). 

This robot is controlled using motor synergies $pi_\theta$ whose parameters $\theta \in \mathbb{R}^n$ directly specify the phase and amplitude of each sinusoid which controls the precise rotational value of each DOF over time. These synergies are parameterized using a set of 24 continuous values, 12 representing the phase $ph$ of each joint, and the 12 others, the amplitude $am$; $\theta = \{ph_{1,2, .., 12}; am_{1, 2, ..,12}\}$, where each joint $i$ receives the command $am \times sin(\omega t + ph)$, with $\omega$ a fixed frequency. Each experimentation consists of launching a motor synergy $\pi_\theta$ for a fixed amount of time, starting from a fixed position. After this time period, the resulting position $y_f$ of the robot is extracted into 3 dimensions: its position $(u,v)$, and its rotation $\phi$. The correspondence $\theta \rightarrow (u,v,\phi)$ is then kept in memory as a learning exemplar. 

\begin{figure*}[t!]
\center
\includegraphics[width=0.8\textwidth]{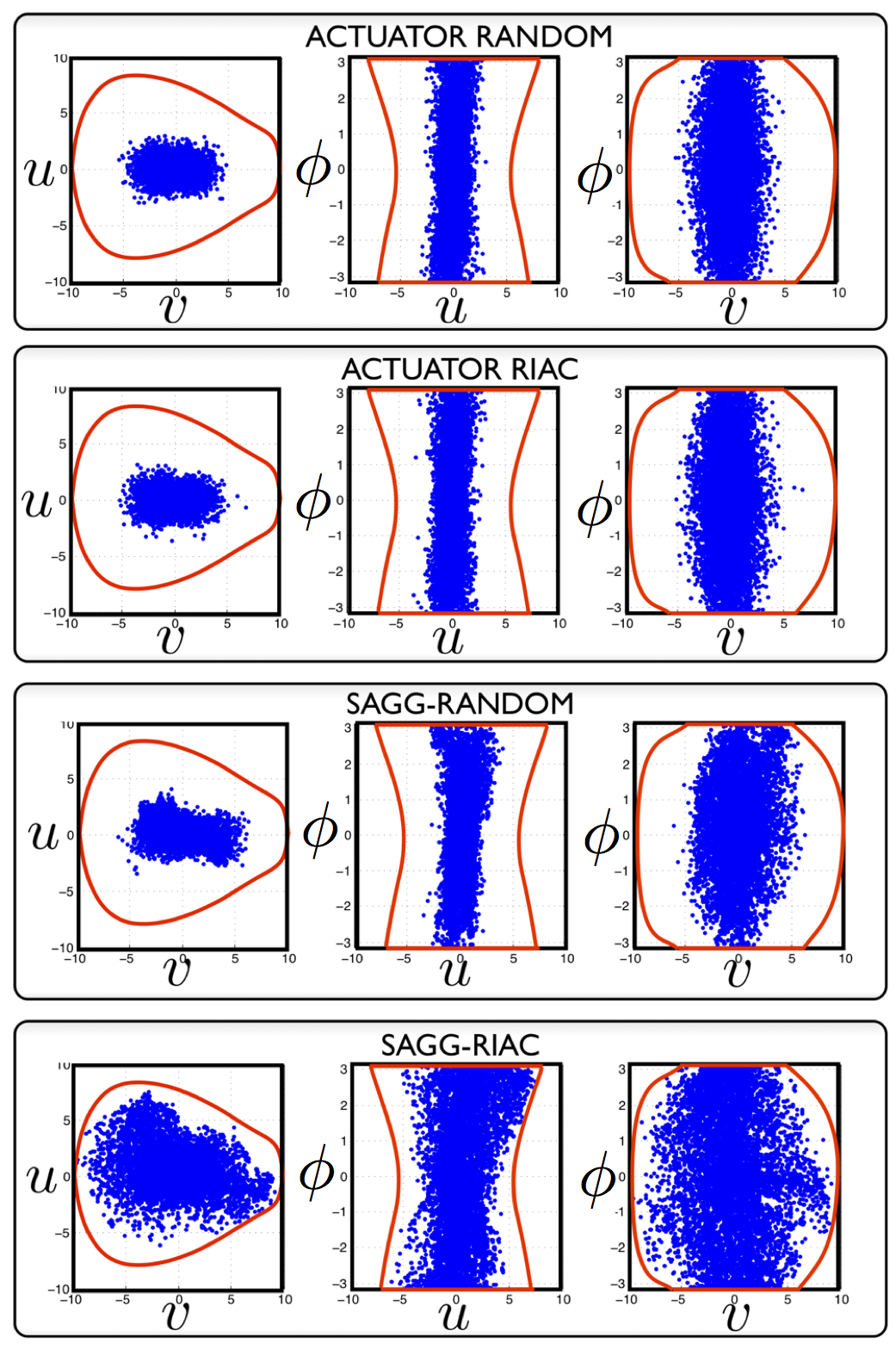}
\caption{Positions explored by the quadruped inside the task space $u,v,\phi$ after 10000 experiments (running a motor synergy during a fixed amount of time), using different exploration mechanisms. Red lines represents estimated limits of reachability. (For interpretation of the references to color in this figure legend, the reader is referred to the web version of this article)}
\label{ExplorationQuadruped}
\end{figure*}

The three dimensions $u,v,\phi$ are used to define the task space of the robot. Also, it is important to notice that precise areas reachable by the quadruped using these motor synergies cannot be estimated beforehand. In the following, we set the original dimensions of the task space to $[-45;45] \times [-45; 45] \times [-2\pi;2\pi]$ on axis $(u,v,\phi)$, which was a priori larger than the reachable space. Then, after having carried out numerous experimentations, it appeared that this task space was actually more than 25 times the size of the area accessible by the robot (see red contours in Fig. \ref{ExplorationQuadruped}). 

The implementation of our algorithm in such a robotic setup aims to test if the SAGG-RIAC driving method allows the robot to learn efficiently and accurately to attain a maximal amount of reachable positions, avoiding the selection of many goals inside regions which are unreachable, or that have previously been visited. 

\begin{figure}
\center
\includegraphics[width=0.5\linewidth]{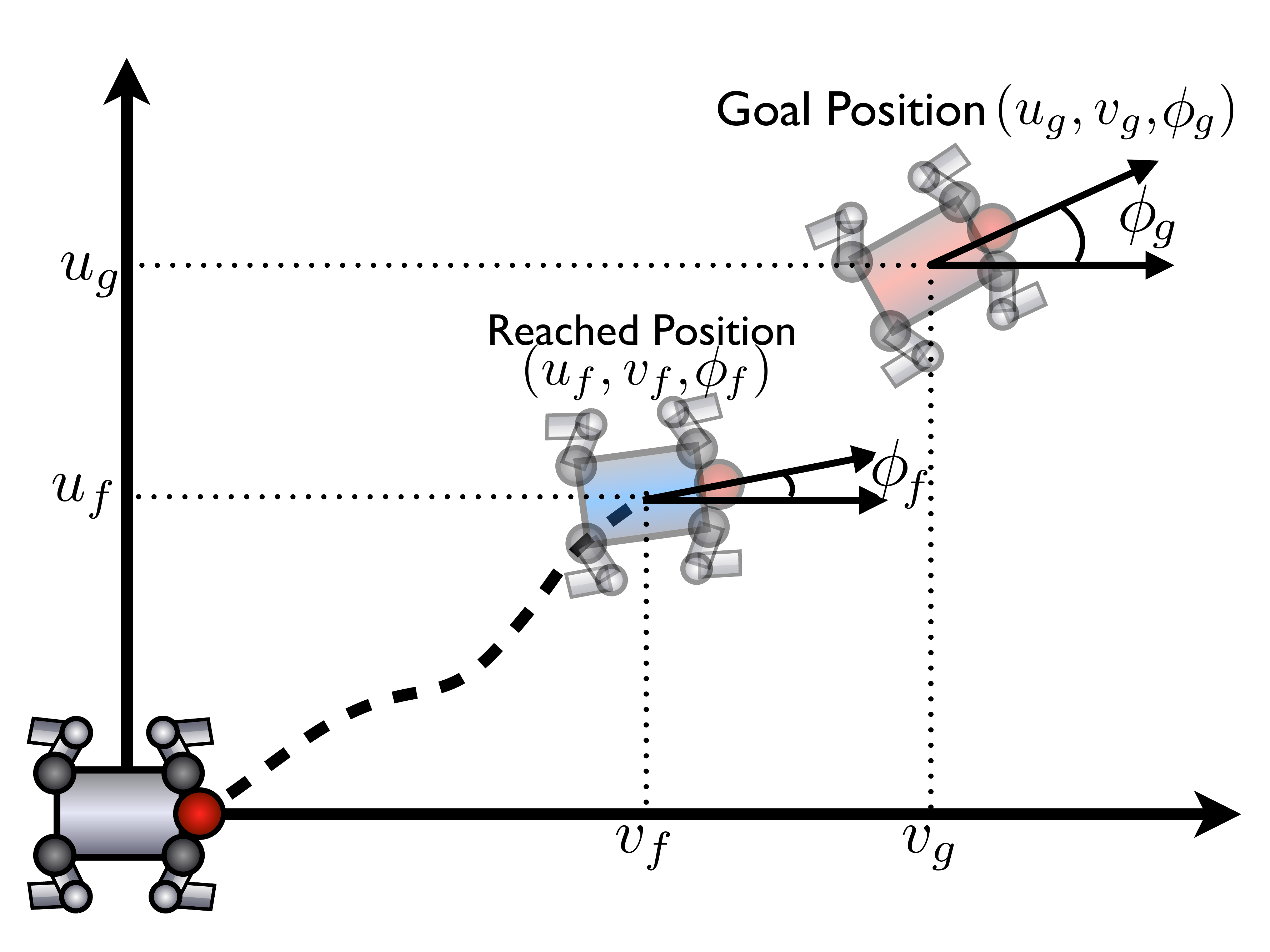}
\caption{Example of experimentation of the quadruped and illustration of beginning position, goal position $(u_g,v_g,\phi_g)$, and a corresponding reached position $(u_f,v_f,\phi_f)$ whose value are used to compute the measure of competence.}
\label{FigQuadruped2}
\end{figure}

\subsection{Measure of competence}

In this experiment, we do not consider constraints $\rho$ and only focus on reaching of the goal positions $y_g = (u_g, v_g, \phi_g)$. In every iteration the robot is reset to a same configuration called the origin position (see Fig. \ref{FigQuadruped2}). We define the competence function $C$ using the Euclidian distance  goal/robot's position $D(y_g,y_f)$ after a reaching attempt, which is normalized by the original distance between the origin position $y_{origin}$, and the goal $D(y_{origin}, y_g)$ (See Fig. \ref{FigQuadruped2}). 

In this measure of competence, we compute the Euclidian distance using $(u,v,\phi)$ where dimensions are rescaled in $[0;1]$. Each dimension therefore has the same weight in the estimation of competence (an angle error of $\phi = \frac{1}{2\pi}$ is as important as an error $u = \frac{1}{90}$ or $v = \frac{1}{90}$).

\begin{eqnarray}
C(y_g, y_f, y_{start}) &=& - \frac{D(y_g,y_f)}{D(y_{start}, y_g)}
\end{eqnarray}
where $C(y_g, y_f, y_{start}) = 0$ if $D(y_{start}, y_g) = 0$.

\subsection{Active Goal Directed Exploration and Learning}

Reaching a goal $y_g$ necessitates the estimation of a motor synergy $\pi_{\theta_{i}}$ leading to this chosen state $y_g$. Considering a single starting configuration for each experimentation, and motor synergies $\pi_\theta$, the forward model which defines this system can be written as the following: 
\begin{eqnarray}
\theta \rightarrow (u,v,\phi)
\end{eqnarray}
Here, we have a direct relationship which only considers the 24 dimensional parameter vector $\theta = \{ph_{1,2, .., 12}; am_{1, 2, ..,12}\}$ of the synergy as inputs of the system, and a position in $(u,v,\phi)$ as output. We thus have a fixed context and use here an instantiation of the SAGG-RIAC architecture with local optimization algorithm Alg. \ref{alg:PseudoCode4}, detailed below. 
 
\subsubsection{Reaching Phase}

The reaching phase deals with reusing the data already acquired and use local regression to compute an inverse model $((u, v, \phi) \rightarrow \theta )_L$ in the locality $L$ of the intended goal $y_g = (u_g, v_g, \phi_g)$. In order to create such a local inverse model (numerous other solutions exist, such as \cite{Bitzer09, Baranes09, Kober-RSS-10,Barto12}), we extract the potentially more reliable data using the following method:

We first extract from the learned data the set $L$ of the $l$ nearest neighbors of $(u_g, v_g, \phi_g)$ and then retrieve their corresponding motor synergies using an ANN method \cite{Muja09}:
\begin{eqnarray}
L &=&  \left\{ \{u,v, \phi, \theta \}_1,  \{u,v, \phi, \theta \}_2,   ... ,  \{u,v, \phi, \theta \}_l   \right\}
\end{eqnarray}
Then, we consider the set $M$ which contains $l$ sets of $m$ elements:
\begin{eqnarray}
M =
\left\{
\begin{array}{ccc}
M_1 : \left\{ \{u,v, \phi, \theta \}_1,  \{u,v, \phi, \theta \}_2,  ...,  \{u,v, \phi, \theta \}_m  \right\}_1 \\
M_2 : \left\{ \{u,v, \phi, \theta \}_1,  \{u,v, \phi, \theta \}_2,  ...,  \{u,v, \phi, \theta \}_m  \right\}_2 \\
... \\
M_l : \left\{ \{u,v, \phi, \theta \}_1,  \{u,v, \phi, \theta \}_2,  ...,  \{u,v, \phi, \theta \}_m  \right\}_l
\end{array}
\right\}
\end{eqnarray}
where each set $\left\{ \{u,v, \phi, \theta \}_1,  \{u,v, \phi, \theta \}_2,  ...,  \{u,v, \phi, \theta \}_m  \right\}_i$ corresponds to the $m$ nearest neighbors of each $\theta_{i}$, $i \in L$, and their corresponding resulting position $(u,v,\phi)$. 

For each set $\left\{ \{u,v, \phi, \theta \}_1,  \{u,v, \phi, \theta \}_2,  ...,  \{u,v, \phi, \theta \}_m  \right\}_{i}$, we estimate the standard deviation $\sigma$ of the parameters of their motor synergies $\theta$ :

\begin{eqnarray}
\sigma(M_j) &=& \sigma \left( \theta_{j} \in \{ \{u,v, \phi, \theta \}_{1,..., m}  \} \right) 
\end{eqnarray}

Finally, we select the set $M_k = \left\{ \{u,v, \phi, \theta \}_1,  \{u,v, \phi, \theta \}_2,  ...,  \{u,v, \phi, \theta \}_m  \right\}$ inside $M$ such that it minimizes the standard deviation of its synergies:
\begin{eqnarray}
M_k = argmin_{i} \ \sigma(M_i)
\end{eqnarray}

From $M_k$, we estimate a local linear inverse model $((u, v, \phi) \rightarrow \theta )$ by using a pseudo-inverse as introduced in the reaching experiment, and use it to estimate the motor synergy parameters $\theta_g$ which correspond to the desired goal $(u_g, v_g, \phi_g)$. 

\subsubsection{Exploration Phase}
The system here continuously estimates the distance between the goal $y_g$ and already reached position $y_c$ which is the closest from the goal. If the reaching phase does not manage to make the system come closer to $y_g$, i.e. $D(y_g, y_t) > D(y_g,y_c)$, with $y_t$ as last effectively reached point in an attempt toward $y_g$, the exploration phase is triggered.

In this phase the system first considers the nearest neighbor $y_c = (u_c, v_c, \phi_c)$ of the goal $(u_g, v_g, \phi_g)$ and gets the corresponding known synergy $\theta_c$. Then, it adds a random noise $rand(24)$ to the 24 parameters $\{ph_{1,2, .., 12}, am_{1, 2, ..,12}\}_c$ of this synergy $\theta_c$ which is proportional to the Euclidian distance $D(y_g,y_c)$. The next synergy $\theta_{t+1} = \{ph_{1,2, .., 12}, am_{1, 2, ..,12}\}_{t+1}$ to experiment can thus be described using the following equation:
\begin{eqnarray}
\theta_{t+1} &=& \left( 
\begin{array}{ccc}
\{ph_{1,2, .., 12}, am_{1, 2, ..,12}\}_c  \\
+ \  \lambda . rand(24) . D(y_g,y_c)
\end{array}
 \right)
\end{eqnarray}

where $rand(i)$ returns a vector of $i$ random values in $[-1;1]$, $\lambda > 0$ and $\{ph_{1,2, .., 12}, am_{1, 2, ..,12}\}_c$ the motor synergy which corresponds to $y_c$.

\subsection{Qualitative Results} 
Fig. \ref{ExplorationQuadruped} presents the positions explored by the quadruped inside the task space $u,v,\phi$ after 10000 experimentations (running of motor synergies during the same fixed amount of time) using the exploration mechanisms introduced previously. ACTUATOR-RANDOM and ACTUATOR-RIAC select parameters of motor synergies in this experiment, whereas SAGG-RANDOM and SAGG-RIAC self-generate goals $(u,v,\phi)$.

Comparing the two first exploration mechanisms (ACTUATOR-RANDOM and ACTUATOR-RIAC) we cannot distinguish any notable difference, the space explored appears similar and the extent of explored space on the $(u,v)$ axis is comprised in the interval $[-5;5]$ for $u$ and  $[-2.5; 2.5]$ for $v$ on both graphs. Moreover, we notice that the difference between $u$ and $v$ scales is due to the inherent structure of the robot, which simplifies the way to go forward and backward rather than shifting left or right.

Considering SAGG methods, it is important to note the difference between the reachable area and the task space. In Fig. \ref{ExplorationQuadruped}, red lines correspond to the estimated reachable area which is comprised of $[-10; 10] \times [-10; 10] \times [-\pi; \pi]$, whereas the task space is much larger: $[-45;45] \times [-45; 45] \times [-2\pi; 2\pi]$. We are also able to notice the asymmetric aspect of its repartition according to the $v$ axis, which is due to the decentered weight of the robot's head.

First, the SAGG-RANDOM method seems to slightly increase the space covered on the $u$ and $v$ axis compared to ACTUATOR methods, as shown by the higher concentration of positions explored in the interval $[-5; -3] \cup [3; 5]$ of $u$. However, this change does not seem very important when comparing SAGG-RANDOM to these two algorithm. 

Second, SAGG-RIAC, contrary to SAGG-RANDOM, shows a large exploration range: the surface in $u$ has almost twice as much coverage than using previous algorithms, and in $v$, up to three times; there is a maximum of $7.5$ in $v$ where the previous algorithms were at $2.5$. These last results emphasize the capability of SAGG-RIAC to drive the learning process inside reachable areas which are not easily accessible (hardly discovered by chance).
 
\subsection{Quantitative Results}
\begin{figure}
\center
\includegraphics[width=0.6\linewidth]{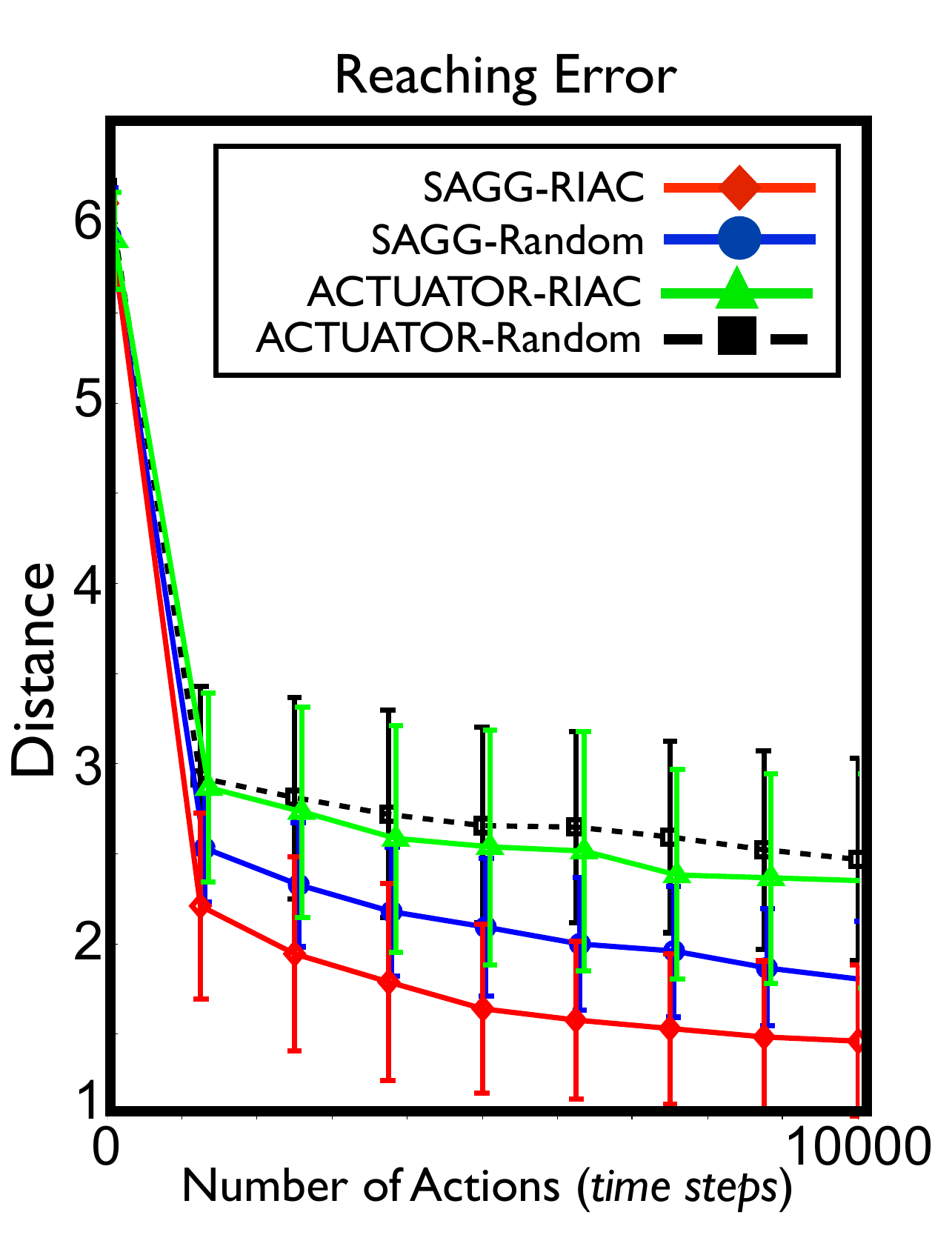}
\caption{Quantitative results for the quadruped measured using the reaching error over the number of experimentations.}
\label{ResultsQuad}
\end{figure}

In this section, we aim to test the efficiency of the learned forward/inverse models to guide the quadruped to reach a set of goal positions from an independently generated test database. Here we consider a test database of 100 goals, generated independantly and covering approximately uniformly the reachable part of the task space, and compute the distance between each goal attempted, and the reached position. Fig. \ref{ResultsQuad} shows performances of the 4 methods introduced previously. First of all, we can observe the higher efficiency of SAGG-RIAC compared to the other three methods which can be observed after only 1000 iterations. The high decreasing velocity of the reaching error (in the number of experimentations) is due to the consideration of regions limited to a small number of elements (30 in this experiment). It allows creating a very high number of regions within a small interval of time, which helps the system to discover and focus on reachable regions and its surrounding area. 

ACTUATOR-RIAC shows slightly more efficient performances than ACTUATOR-RANDOM. Also, even if SAGG-RANDOM is less efficient than SAGG-RIAC, we can observe its highly decreasing reaching errors compared to ACTUATOR methods, which allows it to be significantly more efficient than these method when considered at 10000 iterations. Again, as in the previous experiment, we can also observe that SAGG-RIAC does not only allow to learn faster how to master the sensorimotor space, but that the asymptotic performances also seem to be better \cite{Cohn96}.

\subsection{Conclusion of Results for the Quadruped Experiment}
These experiments first emphasize the high efficiency of methods which drives the exploration of motor synergies in terms of their effects in the task space. As illustrated by qualitative results, SAGG methods, and especially SAGG-RIAC, allows driving the exploration in order to explore large spaces containing areas hardly discovered by chance, when limits of reachability are very difficult to predict. Then, quantitative results showed the capability of SAGG-RANDOM and SAGG-RIAC methods to learn inverse models efficiently when considering highly-redundant robotic systems controlled with motor synergies.

\section{Experimental Setup 3: Learning to Control a Fishing Rod with Motor Synergies}
\label{LabelExp3}
\subsection{Robotic Setup}

\begin{figure}
\center
\includegraphics[width=0.7\linewidth]{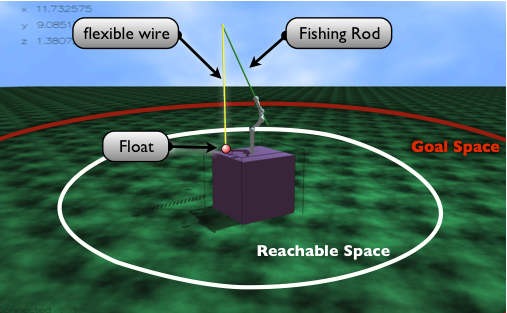}
\caption{4 degrees-of-freedom arm with a fixed fishing rod at its extremity. The arm is controlled using motor synergies which affect the velocity of each joint, and are parameterized by 4 values. Experiments consider a two-dimensional task space $x,y$ which corresponds to the position of the float when touching the water after performing a movement.}
\label{Fishing}
\end{figure}

This experiments consists of having a robot learning to control a fishing rod (with a flexible wire) in order to attain certain positions of the float when it touches the water. This setup is simulated using the Breve simulator, such as in the previous experiment.
The rod is fixed on a 4 DOF arm controlled with motor synergies which affect the velocity of each joint, and are parameterized by the values $\theta = (v_1, v_2, v_3, v_4)$, $v_i \in [0;1]$. More precisely, for each experimentation of the robot we use a low-level pre-programmed PID controller which tracks the desired velocity $v_i$ of each joint $i$ during a fixed short amount of time (2 seconds), starting from a fixed rest position, until suddenly stopping the movement. During the movement, as well as a few second after, we monitor the 3D position of the float in order to detect a potential contact with the water (a flat plane corresponding to the water level). If the water is touched, we extract the 2D coordinates $(x,y)$ of the float on the plane (if not, we do not consider this trial). These coordinates, as well as the parameters of the synergies will be used to describe the forward model of the system as $(v_1, v_2, v_3, v_4) \rightarrow (x,y)$. Learning will thus be performed while recording each set $\{ (v_1, v_2, v_3, v_4), (x, y) \}_i$ as a learning exemplar. In such a sensorimotor space, studying the behavior of SAGG-RIAC is relevant according to the flexible aspect of the line, which makes this system very difficult to model analytically, because it is highly redundant and highly sensitive to small variations of inputs. In the following experiment, the task space will consist of a limited area of the water surface. We will consider the basis of the arm as fixed on the coordinates $(0,0)$, the limits of the task space will be fixed to $[-3;3] \times [-3;3]$ while the reachable region corresponds to a disk whose radius is $1$, and can be contained in $[-1;1] \times [-1;1]$ (see Fig. \ref{Fishing}).

\subsection{Qualitative Results}
\begin{figure}[h!]
\center
\includegraphics[width=\linewidth]{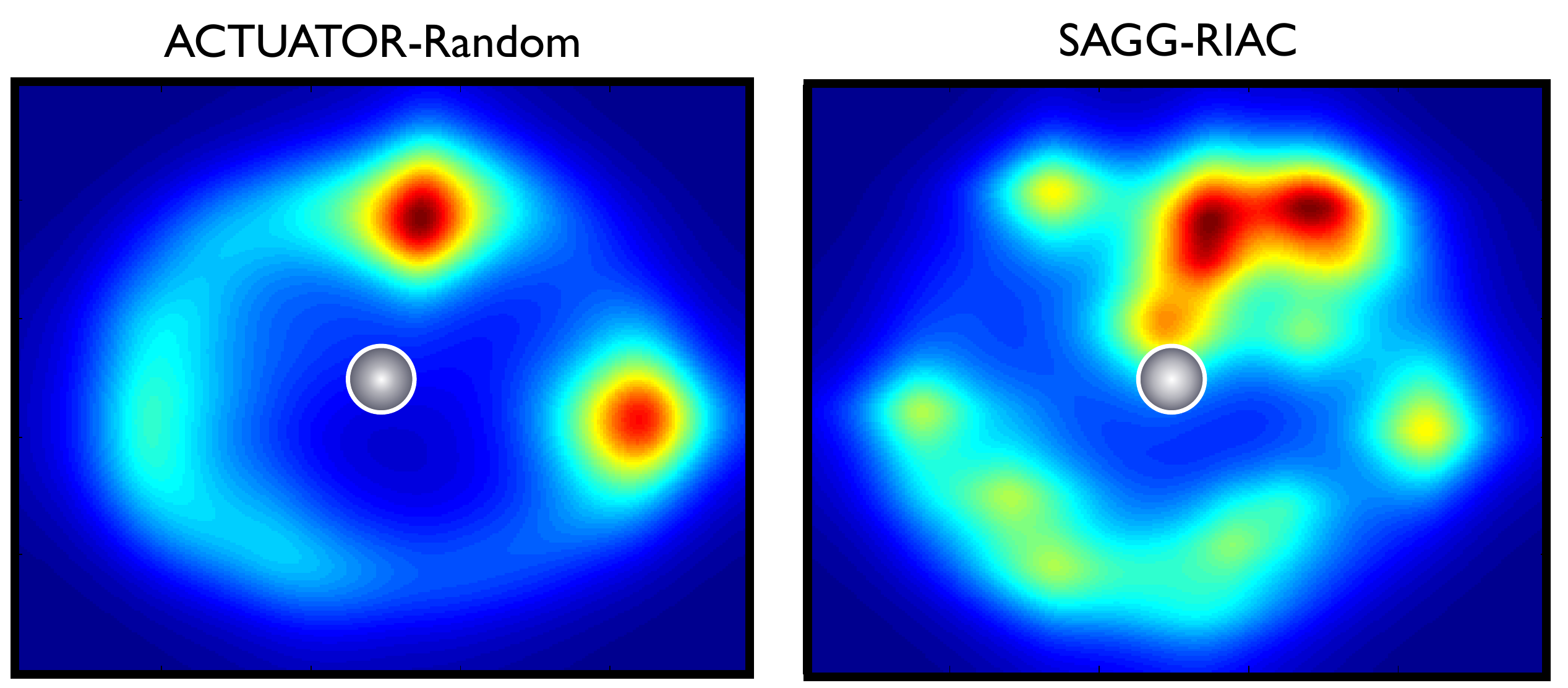}
\caption{Histograms of positions reached by the float when entering in contact with the water in the fishing experiment, after 10000 contact float/water, using ACTUATOR-RANDOM and SAGG-RIAC exploration methods.}
\label{ExplorationFishing}
\end{figure}

Fig. \ref{ExplorationFishing} shows histograms of the repartition of positions reached by the float on the water surface computed after 10000 "water touched" trials (a "water touched" trial corresponds to a reaching attempt where the float effectively touches the surface), after running \textbf{ACTUATOR-RANDOM} and SAGG-RIAC exploration processes. The point situated at the center corresponds to the base to which the arm handling the fishing rod is situated (see Fig. \ref{Fishing}). While observing the two figures, we can note a repartition of positions situated inside a disk, which radius delimits position reached when the line is maximally slack. Yet, the distribution of reached (and reachable) positions within this disk is both asymetrical among and between the two exploration processes. The asymetries on each figure are in fact reflecting the asymetries of the robot setup (see Fig. \ref{Fishing}): the geometry of the robot is not symmetric and its starting/rest configuration is also not symmetric. Coupled with the structure of motor primitives, this makes that the structure of the reachable positions is complex and asymetric, and this can be observed especially in the \textbf{ACTUATOR-RANDOM} sub-figure, since it shows the asymetric distribution of float position reached when the parameters of the action primitives are sampled uniformly (and thus symmetrically). Comparing the two histograms, we note that SAGG-RIAC drives the exploration toward positions of the float not explored by \textbf{ACTUATOR-RANDOM}, such as the large part situated at the bottom of the reachable area. Thus, SAGG-RIAC drives here the exploration toward more diverse regions of the space. SAGG-RIAC is therefore able to avoid spending large amounts of time exclusively guiding the exploration toward the same areas, as \textbf{ACTUATOR-RANDOM} does. Extended experimentation with this setup showed that the distribution of reached points with SAGG-RIAC (right sub-figure) corresponds closely to the actual whole reachable space. Eventually, these qualitative results emphasize that SAGG-RIAC is able to drive the exploration process efficiently when carried out with highly redundant and complex robots with compliant/soft parts.

\subsection{Quantitative Results}
\begin{figure}
\center
\includegraphics[width=0.6\linewidth]{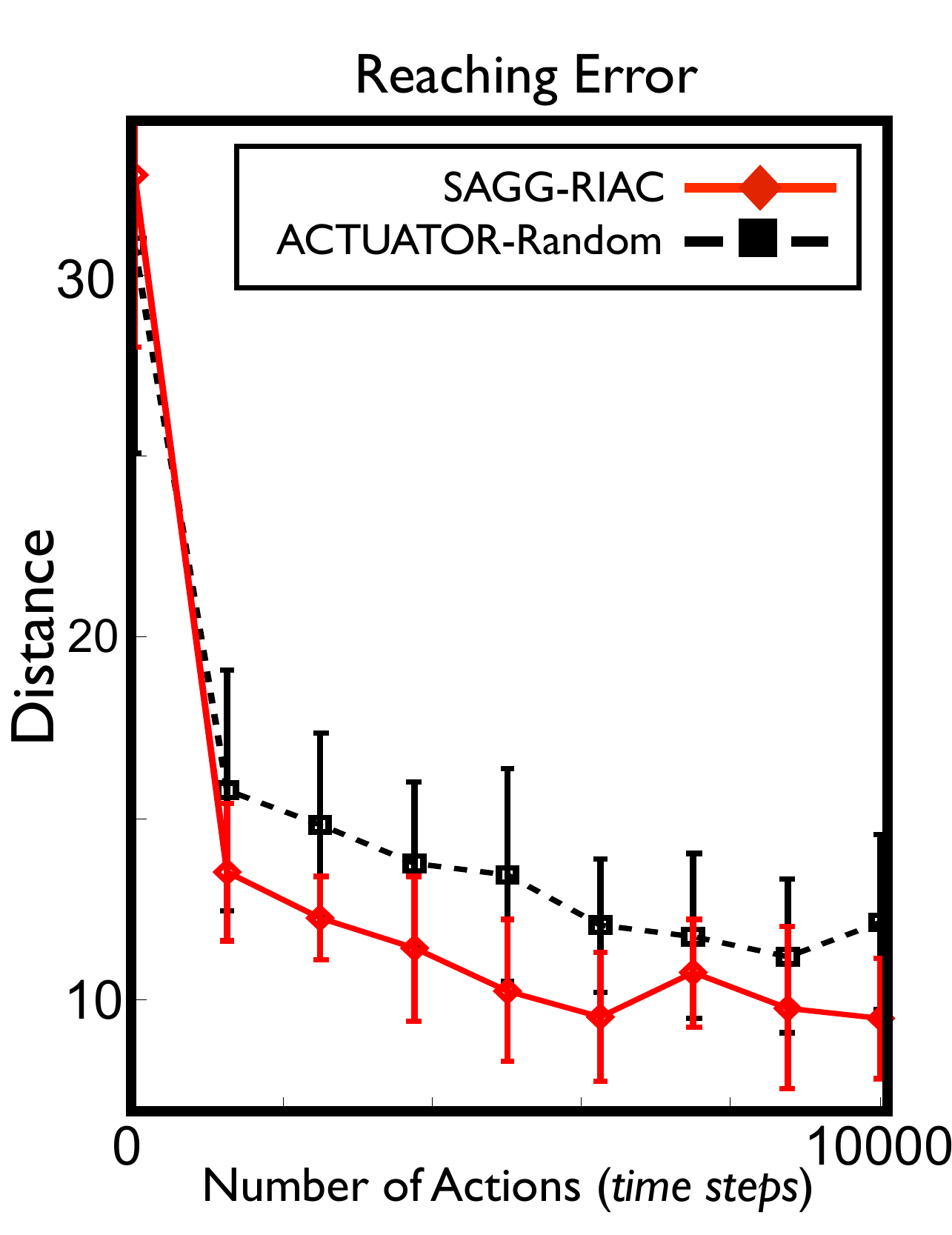}
\caption{Quantitative results for the fishing experiment measured using the reaching error over the number of experimentations.}
\label{ResultsFish}
\end{figure}

Fig. \ref{ResultsFish} shows the mean reaching errors obtained using ACTUATOR-RANDOM and SAGG-RIAC, statistically computed after 10 experiments with different random seeds. Here, the comparison of these two methods shows that SAGG-RIAC led to significantly more efficient results after 1000 successful trials. Also, after 6000 trials, we can observe a small increase in reaching errors of SAGG-RIAC. This phenomenon is due to the discovery of new motor synergies which led to already mastered goal positions. This discovered redundancy reduces the generalization capability for computing the inverse model for a small amount of time until these new parameters of motor synergy have been explored enough to disambiguate the invert model (i.e. two distinct local inverse models are well encoded and do not interfere). 

\section{Conclusion and Future Work}
This paper introduced the Self-Adaptive Goal Generation architecture, SAGG-RIAC, for active learning of inverse models in robotics through intrinsically motivated goal exploration.
First, we demonstrated the high efficiency of learning inverse models by performing an exploration driven by the active self-generation of high-level goals in the parameterized task space instead of traditional motor babbling specified inside a low-level control space. Active exploration in the task space leverages the redundancy often characterizing sensorimotor robotic spaces: this strategy drives robots to learn a maximal amount of tasks (i.e. learn to generate in a controlled manner a maximal number of effects in the task space), instead of numerous ways to perform the same tasks (i.e. learn many action policies to achieve the same effect in the task space). Coupling goal babbling and sophisticated intrinsically motivated active learning also allows a robot to perform efficient autonomous learning of its limits of reachability, and of inverse models with unknown high-dimensional body schemas of different architectures. Intrinsically motivated active learning was here driven by the active stochastic search of areas in the task space where competence progress is maximal. This also allowed emerging developmental trajectories by driving the robot to progressively focus and learn tasks of increasing complexities, while discovering its own limits of reachability, avoiding to spend much exploration time trying to perform impossible tasks.

While we showed that such an approach could allow efficient learning when the action space was continuous and high-dimensional, the experiments performed here were assuming that a low-dimensional task space was initially provided. It is frequent to have such low-dimensional task spaces for useful engineering problems in robotics, where one can assume that an engineer helps the robot learner by designing by hand the task space (including the choice of the variables and parameters specifying the task space). On the other hand, if one would like to use an architecture like SAGG-RIAC in a developmental framework, where one would not assume low-dimensional task spaces pre-specified to the robot, some additional mechanisms should be added to equip the robot with the following two related capabilities:
\begin{itemize}
\item Find autonomously low-dimensional task spaces. Indeed, a too high dimension of a task space would make the evaluation of ``competence progress'' suffer from the curse of dimensionality;
\item Explore actively multiple task spaces (potentially an open-ended number of task space), thus opening the possibility to learn fields of skills which may be of different kinds;
\end{itemize}
There are several potential approaches that could be used to address these issues that include:

\begin{itemize}

\item \textbf{Mechanisms for higher-level stochastic generation of task spaces}, and their active selection through global measures of competence progress, forming an architecture with three levels of active learning (active choice of a task space inside a space of tasks spaces, active choice of goals inside the chosen task space, and active choice of actions to learn to reach the chosen goal) would be a natural extension of the work presented in this article.

\item \textbf{Social guidance} and learning by interaction: social guidance mechanisms allowing a non-engineer human to drive the attention of a robot toward particular task spaces, through physical guidance \cite{Calinon09, Nguyen11} or human-robot interfaces allowing the robot to be attracted toward particular dimensions of the environment \cite{Rouanet09}, may be introduced. 
Inverse reinforcement learning mechanisms, which are able to extract reward functions thanks to examples of action policies could also be seen as a mean to infer interesting task spaces from human demonstrations \cite{Toussaint11}. Social guidance may also be used as a mechanism to bootstrap the evaluation of competence progress, and the identification of zones of reachability, in very large or high-dimensional spaces such as shown in \cite{Nguyen11}, which presents an approach to combine intrinsically motivated learning like SAGG-RIAC with techniques for learning by demonstration.

\item \textbf{Maturational constraints}: Although SAGG-RIAC highly simplifies the learning process by using goal babbling and drives it efficiently thanks to intrinsic motivations, learning still have to begin by a period of random exploration in order to discriminate unreachable areas as well as areas of differing interests. This becomes a problem when the volume of reachable areas in the task space is a lot smaller than the task space itself or when the task space becomes itself high-dimensional. An important direction for future work is to take inspiration from the maturational processes of infants which are constrained in their learning and development by numerous physiological and cognitive mechanisms such as the limitation of their sensorimotor apparatus, as well as the evolving capabilities of their brain \cite{Bernstein67, Turkewitz85, Bongard10, Pfeifer10}. For instance, infants have a reduced visual acuity which prevents them from accessing high visual frequencies as well as distinguishing distant objects. This acuity then progressively grows as the maturation process evolves. Using such constraints in synergy with goal babbling and intrinsic motivation, such as explored in \cite{Baranes11a}, would potentially allow to constrain and simplify further learning since the first actions of the robot \cite{Elman93, French02}, and could be crucial when considering life-long learning in unbounded task spaces. 
\end{itemize}

\section*{Acknowledgment}
We thank everyone who gave us their feedback on the paper.
This research was partially funded by ERC Grant EXPLORERS 240007

\bibliographystyle{plain}

\begin{thebibliography}{100}

\bibitem{Abbeel04}
P.~Abbeel and A.~Ng.
\newblock Apprenticeship learning via inverse reinforcement learning.
\newblock In {\em Proceedings of the 21st International Conference on Machine
  Learning (2004)}, pages 1--8. ACM Press, New York, 2004.

\bibitem{Albert72}
A.~Albert.
\newblock Regression and the moore-penrose pseudo inverse.
\newblock In {\em Mathematics in science and engineering}. Academic Press,
  Inc., 1972.

\bibitem{Arkes82}
H.~R. Arkes and J.~P. Garske.
\newblock Optimal level theories.
\newblock In {\em Psychological theories of motivation}, volume~2, pages
  172--195. 1982.

\bibitem{Arya98}
S.~Arya, D.~M. Mount, N.S. Netanyahu, R.~Silverman, and A.Y. Wu.
\newblock An optimal algorithm for approximate nearest neighbor searching fixed
  dimensions.
\newblock {\em Journal of the ACM (JACM)}, 45(6):891--923, November 1998.

\bibitem{Asada09}
M.~Asada, K.~Hosoda, Y.~Kuniyoshi, H.~Ishiguro, T.~Inui, Y.~Yoshikawa,
  M.~Ogino, and C.~Yoshida.
\newblock Cognitive developmental robotics: A survey.
\newblock {\em IEEE Trans. Autonomous Mental Development}, 1(1), 2009.

\bibitem{Bakker04}
B.~Bakker and J.~Schmidhuber.
\newblock Hierarchical reinforcement learning based on subgoal discovery and
  subpolicy specialization.
\newblock In {\em Proc. 8th Conf. on Intelligent Autonomous Systems (2004)},
  2004.

\bibitem{Baldassare11}
G.~Baldassare.
\newblock What are intrinsic motivations? a biological perspective.
\newblock In {\em Proceeding of the IEEE ICDL-EpiRob Joint Conference}, 2011.

\bibitem{Baranes09}
A.~Baranes and P-Y. Oudeyer.
\newblock Riac: Robust intrinsically motivated exploration and active learning.
\newblock {\em IEEE Trans. on Auto. Ment. Dev.}, 1(3):155--169, 2009.

\bibitem{Baranes10b}
A.~Baranes and P.~Y. Oudeyer.
\newblock Intrinsically motivated goal exploration for active motor learning in
  robots: A case study.
\newblock In {\em Proceedings of the IEEE/RSJ International Conference on
  Intelligent Robots and Systems (IROS)}, Taipei, Taiwan, 2010.

\bibitem{Baranes11a}
A.~Baranes and P-Y. Oudeyer.
\newblock The interaction of maturational constraints and intrinsic motivations
  in active motor development.
\newblock In {\em Proceedings of ICDL-EpiRob 2011}, 2011.

\bibitem{Barto05}
A.~Barto and O.~Simsek.
\newblock Intrinsic motivation for reinforcement learning systems.
\newblock In CT~New~Haven, editor, {\em Proceedings of the Thirteenth Yale
  Workshop on Adaptive and Learning Systems (2005)}, 2005.

\bibitem{Barto04}
A~Barto, S~Singh, and N.~Chenatez.
\newblock Intrinsically motivated learning of hierarchical collections of
  skills.
\newblock In {\em Proc. 3rd Int. Conf. Dvp. Learn.}, pages 112--119, San Diego,
  CA, 2004.

\bibitem{Berlyne60}
D.~Berlyne.
\newblock {\em Conflict, Arousal and Curiosity}.
\newblock McGraw-Hill, 1960.

\bibitem{Berlyne66}
D.~Berlyne.
\newblock Curiosity and exploration.
\newblock {\em Science}, 153(3731):25--33, 1966.

\bibitem{Berniker09}
M.~Berniker, A.~Jarc, E.~Bizzi, and M.C. Tresch.
\newblock Simplified and effective motor control based on muscle synergies to
  exploit musculoskeletal dynamics.
\newblock {\em Proceedings of the National Academy of Sciences of the United
  States of America (PNAS)}, 106(18):7601--7606, May 2009.

\bibitem{Bernstein67}
N~Bernstein.
\newblock {\em The Coordination and Regulation of Movements}.
\newblock Pergamon, 1967.

\bibitem{Berthier99}
N.~E. Berthier, R.K. Clifton, D.D. McCall, and D.J. Robin.
\newblock Proximodistal structure of early reaching in human infants.
\newblock {\em Exp Brain Res}, 1999.

\bibitem{Billard08}
Aude Billard, Sylvain Calinon, Rudiger Dillmann, and Stefan Schaal.
\newblock Robot programming by demonstration.
\newblock In B.~Siciliano and O.~Khatib, editors, {\em Handbook of Robotics},
  pages 1371--1394. Springer, 2008.

\bibitem{Bishop07}
C.M. Bishop.
\newblock Pattern recognition and machine learning.
\newblock In {\em Information Science and Statistics}. Springer, 2007.

\bibitem{Bitzer09}
S.~Bitzer and S.~Vijayakumar.
\newblock Latent spaces for dynamic movement primitives.
\newblock In {\em Proceedings of IEEE/RAS International Conference on Humanoid
  Robots}, 2009.

\bibitem{Blank05}
D.~Blank, D.~Kumar, L.~Meeden, and J.~Marshall.
\newblock Bringing up robot: Fundamental mechanisms for creating a
  self-motivated, self-organizing architecture.
\newblock {\em Cybernetics and Systems}, 36(2), 2005.

\bibitem{Bongard10}
Josh~C. Bongard.
\newblock Morphological change in machines accelerates the evolution of robust
  behavior.
\newblock {\em Proceedigns of the National Academy of Sciences of the United
  States of America (PNAS)}, January 2010.

\bibitem{Bullock93}
D.~Bullock, S.~Grossberg, and F.H. Guenther.
\newblock A self-organizing neural model of motor equivalent reaching and tool
  use by a multijoint arm.
\newblock {\em Journal of Cognitive Neuroscience}, 5(4):408--435, 1993.

\bibitem{Calinon09}
S.~Calinon and A.~Billard.
\newblock Statistial learning by imitation of competing constraints in joint
  space and task space.
\newblock {\em Advanced Robotics}, 23(15):2059--2076, 2009.

\bibitem{Lopes10b}
R.~Cantin-Martinez, M.~Lopes, and L.~Montesano.
\newblock Body schema acquisition through active learning.
\newblock In {\em EEE - International Conference on Robotics and Automation
  (ICRA)}, Anchorage, Alaska, USA, 2010.

\bibitem{Cederborg10}
T.~Cederborg, M.~Li, A.~Baranes, and P-Y. Oudeyer.
\newblock Incremental local inline gaussian mixture regression for imitation
  learning of multiple tasks.
\newblock In {\em Proceedings of the IEEE/RSJ International Conference on
  Intelligent Robots and Systems (2010)}, Taipei, Taiwan, 2010.

\bibitem{Chernova09}
S.~Chernova and M.~Veloso.
\newblock Interactive policy learning through confidence-based autonomy.
\newblock {\em J. Artificial Intelligence Research}, 34:1--25, 2009.

\bibitem{Chiaverini97}
S~Chiaverini.
\newblock Singularity-robust task-priority redundancy resolution for real-time
  kinematic control of robot manipulators.
\newblock {\em IEEE Transactions on Robotics and Automation}, 13(3):398--410,
  1997.

\bibitem{Cohn94}
D.~Cohn, L.~Atlas, and R.~Ladner.
\newblock Improving generalization with active learning.
\newblock {\em Mach. Learn.}, 15(2):201--221, 1994.

\bibitem{Cohn96}
David~A. Cohn, Zoubin Ghahramani, and Michael~I. Jordan.
\newblock Active learning with statistical models.
\newblock {\em J Artificial Intelligence Research}, 4:129--145, 1996.

\bibitem{Csik96}
M.~Csikszentmihalyi.
\newblock {\em Creativity-Flow and the Psychology of Discovery and Invention}.
\newblock Harper Perennial, New York, 1996.

\bibitem{Dasgupta04}
S.~Dasgupta.
\newblock Analysis of a greedy active learning strategy.
\newblock {\em Adv. Neural Inform. Process. Systems}, 17, 2004.

\bibitem{Dattorro11}
J.~Dattorro.
\newblock {\em Convex Optimization and Euclidean Distance Geometry}.
\newblock Meboo Publishing USA, 2011.

\bibitem{Avella03}
A.~D'Avella, P.~Saltiel, and E.~Bizzi.
\newblock Combinations of muscle synergies in the construction of a natural
  motor behavior.
\newblock {\em Nature neuroscience}, 6(3):300--308, march 2003.

\bibitem{Deci85}
E.L. Deci and M.~Ryan.
\newblock {\em Intrinsic Motivation and self-determination in human behavior}.
\newblock Plenum Press, New York, 1985.

\bibitem{Delcomyn80}
F.~Delcomyn.
\newblock Neural basis for rhythmic behaviour in animals.
\newblock {\em Science}, 210:492--498, 1980.

\bibitem{Dichter97}
T.~B. Dichter, N.~A. Busch, and D.~E. Knauf.
\newblock Mastery motivation: Appropriate tasks for toddlers.
\newblock {\em Infant Behavior and Development}, 20(4):545--548, 1997.

\bibitem{Drescher91}
G.L. Drescher.
\newblock {\em Made-Up Minds: A Constructivist Approach to Artificial
  Intelligence}.
\newblock MIT Press, 1991.

\bibitem{Elman93}
J.~Elman.
\newblock Learning and development in neural networks: The importance of
  starting small.
\newblock {\em Cognition}, 48:71--99, 1993.

\bibitem{Fasel10}
I.~Fasel, A.~Wilt, N.~Mafi, and C.~T. Morrison.
\newblock Intrinsically motivated information foraging.
\newblock In {\em Proceedings of the IEEE 9th International Conference on
  Development and Learning (2010)}, 2010.

\bibitem{Fedorov72}
V.~Fedorov.
\newblock {\em Theory of Optimal Experiment}.
\newblock Academic Press, Inc., New York, NY, 1972.

\bibitem{French02}
R.~M. French, M.~Mermillod, P.~C. Quinn, A.~Chauvin, and D.~Mareschal.
\newblock The importance of starting blurry: Simulating improved basic-level
  category learning in infants due to weak visual acuity.
\newblock In LEA, editor, {\em Proceedings of the 24th Annual Conference of the
  Cognitive Science Society (2002)}, pages 322--327, New Jersey, 2002.

\bibitem{Freund97}
Y.~Freund, H.S. Seung, E.~Shamir, and N.~Tishby.
\newblock Selective sampling using the query by committee algorithm.
\newblock {\em Machine Learning}, 28(2-3):133--168, 1997.

\bibitem{Grillner85}
S.~Grillner and P.~Wallen.
\newblock Central pattern generators for locomotion, with special reference to
  vertebrates.
\newblock {\em Annual Review of Neuroscience}, 8:233--261, march 1985.

\bibitem{Hansen01}
N.~Hansen and A.~Ostermeier.
\newblock Completely derandomized self- adaptation in evolution strategies.
\newblock {\em Evolutionary Computation}, 9(2):159--195, 2001.

\bibitem{Hart08b}
S.~Hart and R.~Grupen.
\newblock Intrinsically motivated hierarchical manipulation.
\newblock In {\em Proceedings of the IEEE Conference on Robots and Automation
  (2008)}, 2008.

\bibitem{Hart08}
S.~Hart, S.~Sen, and R.~Grupen.
\newblock Generalization and transfer in robot control.
\newblock In Lund Univeristy~Cognitive Studies, editor, {\em Proc. Of the 8th
  International Conference On Epigenetic Robotics (2008)}, University of
  Sussex, 2008.

\bibitem{Huang02}
X.~Huang and J.~Weng.
\newblock Novelty and reinforcement learning in the value system of
  developmental robots.
\newblock In C.~Prince, Y.~Demiris, Y.~Marom, H.~Kozima, and C.~Balkenius,
  editors, {\em Proc 2nd Int. Workshop Epigenetic Robotics: Modeling Cognitive
  Development in Robotic Systems}, volume~94, pages 47--55. Lund University
  Cognitive Studies, 2002.

\bibitem{Ijspeert08}
A.J. Ijspeert.
\newblock Central pattern generators for locomotion control in animals and
  robots: A review.
\newblock {\em Neural Networks}, 21(4):642--653, 2008.

\bibitem{Toussaint11}
N.~Jetchev and M.~Toussaint.
\newblock Task space retrieval using inverse feedback control.
\newblock In L.~Getoor and T.~Scheffer, editors, {\em International Conference
  on Machine Learning (ICML-11)}, volume~28, pages 449--456, New York, NY, USA,
  2011.

\bibitem{Jones98}
D.~Jones, M.~Schonlau, and W.~Welch.
\newblock Efficient global optimization of expensive black-box functions.
\newblock {\em Global Optimization}, 13(4):455--492, 1998.

\bibitem{Kakade02}
S.~Kakade and P.~Dayan.
\newblock Dopamine: Generalization and bonuses.
\newblock {\em Neural Networks}, 15(4-6):549--59, 2002.

\bibitem{Kapoor07}
A.~Kapoor, K.~Grauman, R.~Urtasun, and T.~Darrell.
\newblock Active learning with gaussian processes for object categorization.
\newblock In {\em Proceeding of the IEEE 11th Int. Conf. Comput. Vis. (2007)},
  Crete, Greece, 2007.

\bibitem{Breve}
J.~Klein.
\newblock Breve: a 3d environment for the simulation of decentralized systems
  and artificial life.
\newblock In MIT Press, editor, {\em Proceeding of the eighth international
  conference on artificial life (2003)}, 2003.

\bibitem{Kober-RSS-10}
J.~Kober, E.~Oztop, and J.~Peters.
\newblock Reinforcement learning to adjust robot movements to new situations.
\newblock In {\em Proceedings of Robotics: Science and Systems}, 2010.

\bibitem{Krause07}
A.~Krause and C.~Guestrin.
\newblock Nonmyopic active learning of gaussian processes: an
  exploration-exploitation approach.
\newblock In {\em 24th international conference on Machine learning}, 2007.

\bibitem{Krause08}
A.~Krause, A.~Singh, and C.~Guestrin.
\newblock Near-optimal sensor placements in gaussian processes: Theory,
  efficientalgorithms and empirical studies.
\newblock {\em Journal of Machine Learning Research}, 9:235--284, 2008.

\bibitem{Lee84}
W.A. Lee.
\newblock Neuromotor synergies as a basis for coordinated intentional action.
\newblock {\em J. Mot. Behav.}, 16:135--170, 1984.

\bibitem{Lopes09}
M.~Lopes, F.~Melo, B.~Kenward, and J.~Santos-Victor.
\newblock A computational model of social-learning mechanisms.
\newblock {\em Adaptive Behavior}, 467(17), 2009.

\bibitem{Lopes10}
M.~Lopes, f.~Melo, and J.~Santos-Victor.
\newblock Abstraction levels for robotic imitation: Overview and computational
  approaches.
\newblock In {\em From Motor Learning to Interaction Learning in Robots}.
  SpringerLink, 2010.

\bibitem{Lopes09b}
M.~Lopes, F.~S. Melo, and L.~Montesano.
\newblock Active learning for reward estimation in inverse reinforcement
  learning.
\newblock In {\em European Conference on Machine Learning (ECML/PKDD)}, Bled,
  Slovenia, 2009.

\bibitem{Oudeyer10b}
M.~Lopes and P-Y. Oudeyer.
\newblock Active learning and intrinsically motivated exploration in robots:
  Advances and challenges (guest editorial).
\newblock {\em IEEE Transactions on Autonomous Mental Development},
  2(2):65--69, 2010.

\bibitem{Lopes07}
M.~Lopes and J.~Santos-Victor.
\newblock A developmental roadmap for learning by imitation in robots.
\newblock {\em IEEE Transactions in Systems Man and Cybernetic, Part B:
  Cybernetics}, 37(2), 2007.

\bibitem{Luciw11}
M.~Luciw, V.~Graziano, M.~Ring, and J.~Schmidhuber.
\newblock Artificial curiosity with planning for autonomous perceptual and
  cognitive development.
\newblock In {\em Proceeding of the First IEEE ICDL-EpiRob Joint Conference
  (2011)}, 2011.

\bibitem{Marshall04}
J.~Marshall, D.~Blank, and L.~Meeden.
\newblock An emergent framework for self-motivation in developmental robotics.
\newblock In {\em Proc. 3rd Int. Conf. Development Learn. (2004)}, pages
  104--111, San Diego, CA, 2004.

\bibitem{Pfeifer10}
Harold Martinez, Max Lungarella, and Rolf Pfeifer.
\newblock On the influence of sensor morphology on eye motion coordination.
\newblock In {\em Proc. of the IEEE 9th International Conference on Development
  and Learning (2010)}, pages 238 --243, August 2010.

\bibitem{Martinez09}
R.~Martinez-Cantin, N.~de~Freitas, E.~Brochu, J.~Castellanos, and A.~Doucet.
\newblock A {Bayesian} exploration-exploitation approach for optimal online
  sensing and planning with a visually guided mobile robot.
\newblock {\em Autonomous Robots - Special Issue on Robot Learning, Part B},
  2009.

\bibitem{Merrick09}
Kathryn Merrick and Mary~Lou Maher.
\newblock Motivated learning from interesting events: Adaptive, multitask
  learning agents for complex environments.
\newblock {\em Adaptive Behavior - Animals, Animats, Software Agents, Robots,
  Adaptive Systems}, 17(1):7--27, 2009.

\bibitem{Mugan09}
J.~Mugan and B.~Kuipers.
\newblock Autonomously learning an action hierarchy using a learned qualitative
  state representation.
\newblock In {\em Proceedings of the International Joint Conference on
  Artificial Intelligence (2009)}, 2009.

\bibitem{Muja09}
M.~Muja and D.G. Lowe.
\newblock Fast approximate nearest neighbors with automatic algorithm.
\newblock In {\em International Conference on Computer Vision Theory and
  Applications (2009)}, 2009.

\bibitem{Nguyen11}
M.~Nguyen, A.~Baranes, and P-Y. Oudeyer.
\newblock Bootstrapping intrinsically motivated learning with human
  demonstrations.
\newblock In {\em proceedings of the IEEE International Conference on
  Development and Learning}, Frankfurt, Germany, 2011.

\bibitem{Peters11}
D.~Nguyen-Tuong and J.~Peters.
\newblock Model learning for robot control: A survey.
\newblock {\em Cognitive Processing}, 12(4):319--340, 2011.

\bibitem{Nishii94}
J.~Nishii, Y.~Uno, and R.~Suzuki.
\newblock Mathematical models for the swimming pattern of a lamprey.
\newblock In {\em Biological Cybernetics}, volume~72. Springer, 1994.

\bibitem{Oudeyer08}
P.~Oudeyer and F.~Kaplan.
\newblock How can we define intrinsic motivations ?
\newblock In {\em Proc. Of the 8th Conf. On Epigenetic Robotics (2008)}, 2008.

\bibitem{Oudeyer05}
P.~Oudeyer, F.~Kaplan, V.~Hafner, and A.~Whyte.
\newblock The playground experiment: Task-independent development of a curious
  robot.
\newblock In {\em Proceedings of the AAAI Spring Symposium on Developmental
  Robotics}, pages 42--47, 2005.

\bibitem{Oudeyer07b}
P.-Y. Oudeyer and F.~Kaplan.
\newblock What is intrinsic motivation? a typology of computational approaches.
\newblock {\em Frontiers of Neurorobotics}, page 1:6, 2007.

\bibitem{Oudeyer07}
P-Y. Oudeyer, F.~Kaplan, and V.~Hafner.
\newblock Intrinsic motivation systems for autonomous mental development.
\newblock {\em IEEE Transactions on Evolutionary Computation}, 11(2):pp.
  265--286, 2007.

\bibitem{Peters08d}
J.~Peters and S.~Schaal.
\newblock Natural actor critic.
\newblock {\em Neurocomputing}, (7-9):1180--1190, 2008.

\bibitem{Peters08c}
J.~Peters and S.~Schaal.
\newblock reinforcement learning of motor skills with policy gradients.
\newblock (4):682--97, 2008.

\bibitem{Peters03}
J.~Peters, S.~Vijayakumar, and S.~Schaal.
\newblock Reinforcement learning for humanoid robotics.
\newblock In {\em Third IEEE-RAS International Conference on Humanoid Robots
  (2003)}, 2003.

\bibitem{Artoolkit00}
I.~Poupyrev, H.~Kato, and M.~Billinghurst.
\newblock Artoolkit user manual, version 2.33.
\newblock Technical report, University of Washington, 2000.

\bibitem{Redding88}
R.~E. Redding, G.~A. Morgan, and R.~J. Harmon.
\newblock Mastery motivation in infants and toddlers: Is it greatest when tasks
  are moderately challenging?
\newblock {\em Infant Behavior and Development}, 11(4):419--430, 1988.

\bibitem{Redgrave06}
P.~Redgrave and K.~Gurney.
\newblock The short-latency dopamine signal: A role in discovering novel
  actions?
\newblock {\em Nat. Rev. Neurosci.}, 7(12):967--75, Nov 2006.

\bibitem{Riedmiller09}
M.~Riedmiller, T.~Gabel, R.~Hafner, and S.~Lange.
\newblock Reinforcement learning for robot soccer.
\newblock {\em Autonomous Robot}, 27:55--73, 2009.

\bibitem{Rolf10a}
M.~Rolf, J.~Steil, and M.~Gienger.
\newblock Goal babbling permits direct learning of inverse kinematics.
\newblock {\em IEEE Trans. Autonomous Mental Development}, 2(3):216--229,
  09/2010 2010.

\bibitem{Rolf11}
M.~Rolf, J.~Steil, and M.~Gienger.
\newblock Online goal babbling for rapid bootstrapping of inverse models in
  high dimensions.
\newblock In {\em Proceeding of the IEEE ICDL-EpiRob Joint Conference (2011)},
  2011.

\bibitem{Hofsten94}
L.~Ronnquist and C.~von Hofsten.
\newblock Neonatal finger and arm movements as determined by a social and an
  object context.
\newblock {\em Early Develop. Parent.}, 3(2):81--94, 1994.

\bibitem{Rouanet09}
P.~Rouanet, P-Y. Oudeyer, and D.~Filliat.
\newblock An integrated system for teaching new visually grounded words to a
  robot for non-expert users using a mobile device.
\newblock In {\em Proceedings of IEEE-RAS International Conference on Humanoid
  Robots (HUMANOIDS 2010)}, Paris, France, 2009.

\bibitem{Roy01}
N.~Roy and A.~McCallum.
\newblock Towards optimal active learning through sampling estimation of error
  reduction.
\newblock In {\em Proc. 18th Int. Conf. Mach. Learn. (2001)}, volume~1, pages
  143--160, 2001.

\bibitem{Ryan00}
Richard~M. Ryan and Edward~L. Deci.
\newblock Intrinsic and extrinsic motivations: Classic definitions and new
  directions.
\newblock {\em Contemporary Educational Psychology}, 25(1):54 -- 67, 2000.

\bibitem{Salaun10}
C.~Salaun, V.~Padois, and O.~Sigaud.
\newblock Learning forward models for the operational space control of
  redundant robots.
\newblock In {\em From Motor Learning to Interaction Learning in Robots},
  volume 264, pages 169--192. Springer, 2010.

\bibitem{Schaal94}
S.~Schaal and C.~G. Atkeson.
\newblock robot juggling: an implementation of memory-based learning.
\newblock {\em Control systems magazine}, pages 57--71, 1994.

\bibitem{Schein07}
A.~Schein and L.H. Ungar.
\newblock Active learning for logistic regression: An evaluation.
\newblock {\em Machine Learning}, 68:235--265, 2007.

\bibitem{Schembri07b}
M.~Schembri, M.~Mirolli, and Baldassarre G.
\newblock Evolution and learning in an intrinsically motivated reinforcement
  learning robot.
\newblock In Springer, editor, {\em Proceedings of the 9th European Conference
  on Artificial Life (2007)}, pages 294--333, Berlin, 2007.

\bibitem{Schembri07}
M.~Schembri, M.~Mirolli, and Baldassarre G.
\newblock Evolving childhood's length and learning parameters in an
  intrinsically motivated reinforcement learning robot.
\newblock In {\em Proceedings of the Seventh International Conference on
  Epigenetic Robotics}. Lund University Cognitive Studies, 2007.

\bibitem{Schmidhuber91}
J.~Schmidhuber.
\newblock Curious model-building control systems.
\newblock In {\em Proc. Int. Joint Conf. Neural Netw. (1991)}, volume~2, pages
  1458--1463, 1991.

\bibitem{Schmidhuber91c}
J.~Schmidhuber.
\newblock A possibility for implementing curiosity and boredom in
  model-building neural controllers.
\newblock In J.~A. Meyer and S.~W. Wilson, editors, {\em Proc. SAB'91}, pages
  222--227, 1991.

\bibitem{Schmidhuber99}
J.~Schmidhuber.
\newblock Artificial curiosity based on discovering novel algorithmic
  predictability through coevolution.
\newblock In P.~Angeline, Z.~Michalewicz, M.~Schoenauer, X.~Yao, and
  Z.~Zalzala, editors, {\em Congress on Evolutionary Computation}, pages
  1612--1618, Piscataway, NJ, 1999. IEEE Press.

\bibitem{Schmidhuber02}
J.~Schmidhuber.
\newblock {\em Exploring the Predictable}, pages 579--612.
\newblock Springer, 2002.

\bibitem{Schmidhuber06}
J.~Schmidhuber.
\newblock Optimal artificial curiosity, developmental robotics, creativity,
  music, and the fine arts.
\newblock {\em Connection Science}, 18(2), 2006.

\bibitem{Schmidhuber10}
J.~Schmidhuber.
\newblock Formal theory of creativity, fun, and intrinsic motivation.
\newblock {\em IEEE Transaction on Autonomous Mental Development},
  2(3):230--247, 2010.

\bibitem{Schmidhuber11}
J.~Schmidhuber.
\newblock Powerplay: Training an increasingly general problem solver by
  continually searching for the simplest still unsolvable problem.
\newblock In {\em Report arXiv:1112.5309}, 2011.

\bibitem{Schohn00}
G.~Schohn and D.~Cohn.
\newblock Less is more: Active learning with support vector machines.
\newblock {\em Proceedings of the Seventeenth International Conference on
  Machine Learning}, 2000.

\bibitem{Schultz97}
W.~Schultz, P.~Dayan, and P.~Montague.
\newblock A neural substrate of prediction and reward.
\newblock {\em Science}, 275:1593--1599, 1997.

\bibitem{Settles09}
B.~Settles.
\newblock Active learning literature survey.
\newblock CS Tech. Rep. 1648, Univ. Wisconsin-Madison, Madison, WI, 2009.

\bibitem{Handbook08}
Siciliano and Khatib.
\newblock {\em Handbook of Robotics}.
\newblock Springer, 2008.

\bibitem{Sigaud11}
O.~Sigaud, C.~Salaun, and V.~Padois.
\newblock On-line regression algorithms for learning mechanical models of
  robots: a survey.
\newblock {\em Robotics and Autonomous System}, 59(12):1115--1129, July 2011.

\bibitem{Barto12}
B.~Castro~Da Silva, G.~Konidaris, and A.~Barto.
\newblock Learning parameterized skills.
\newblock In {\em Proceedings of International Conference of Machine Learning},
  2012.

\bibitem{Barto06}
{\"O}.~{\c S}im{\c s}ek and A.G. Barto.
\newblock An intrinsic reward mechanism for efficient exploration.
\newblock In {\em Proceedings of the Twenty-Third International Conference on
  Machine Learning (2006)}, 2006.

\bibitem{Barto10}
S.~Singh, R.L. Lewis, A.G. Barto, and J.~Sorg.
\newblock Instrinsically motivated reinforcement learning: An evolutionary
  perspective.
\newblock {\em IEEE Transactions on Autonomous Mental Development (IEEE TAMD)},
  2(2):70--82, 2010.

\bibitem{Stout10}
A.~Stout and A~Barto.
\newblock Competence progress intrinsic motivation.
\newblock In {\em Proceedings of the International Conference on Development
  and Learning (2010)}, 2010.

\bibitem{Stulp12}
F.~Stulp and O.~Sigaud.
\newblock Path integral policy improvement with covariance matrix adaptation.
\newblock In {\em Proceedings of International Conference of Machine Learning},
  2012.

\bibitem{Stulp11}
F.~Stulp, E.~Theodorou, M.~Kalakrishnan, P.~Pastor, L.~Righetti, and S.~Schaal.
\newblock Learning motion primitive goals for robust manipulation.
\newblock In {\em Int. Conference on Intelligent Robots and Systems (IROS),},
  2011.

\bibitem{Sutton98}
Richard~S. Sutton and Andrew~G. Barto.
\newblock Reinforcement learning: An introduction, 1998.

\bibitem{Sutton90}
R.S. Sutton.
\newblock Integrated architectures for learning, planning, and reacting based
  on approximating integrated architectures for learning, planning, and
  reacting based on approximating dynamic programming.
\newblock In {\em Proceedings of the International Machine Learning
  Conference}, pages 212--218, 1990.

\bibitem{Sutton99}
R.S. Sutton, D.~Precup, and S.~Singh.
\newblock Between mdps and semi-mdps: A framework for temporal abstraction in
  reinforcement learning.
\newblock {\em Artificial Intelligence}, 1123(181-211), 1999.

\bibitem{Theodorou07}
E~Theodorou, J~Peters, and S.~Schaal.
\newblock Reinforcement learning for optimal control of arm movements.
\newblock In {\em Abstracts of the 37st meeting of the society of neuroscience
  (2007)}, 2007.

\bibitem{Thrun95}
S.~Thrun.
\newblock Exploration in active learning.
\newblock In M~Arbib, editor, {\em Handbook of Brain Science and Neural
  Networks}, Cambridge, MA, 1995. MIT Press.

\bibitem{Thrun92c}
S.~Thrun and K.~Moller.
\newblock Active exploration in dynamic environments.
\newblock In {\em Proceedings of Advances of Neural Information Processing
  Systems}, 1992.

\bibitem{Ting07}
L.~Ting and J.~McKay.
\newblock Neuromechanics of muscle synergies for posture and movement.
\newblock {\em Curr. Opin. Neubiol.}, 7:622--628, 2007.

\bibitem{Toussaint06}
Marc Toussaint and Amos Storkey.
\newblock Probabilistic inference for solving discrete and continuous state
  markov decision processes.
\newblock In {\em Proceedings of the 23rd international conference on Machine
  learning}, ICML '06, pages 945--952, New York, NY, USA, 2006. ACM.

\bibitem{Turkewitz85}
G.~Turkewitz and P.A. Kenny.
\newblock The role of developmental limitations of sensory input on
  sensory/perceptual organization.
\newblock {\em J Dev Behav. Pediatr.}, 6(5):302--6, 1985.

\bibitem{Meer97}
A.~van~der Meer.
\newblock Keeping the arm in the limelight: Advanced visual control of arm
  movements in neonates.
\newblock {\em Eur. J. Paediatric Neurol}, 1(4):103--108, 1997.

\bibitem{Meer95}
A.~van~der Meer, F.~van~der Weel, and D.~Lee.
\newblock The functional significance of arm movements in neonates.
\newblock {\em Science}, 267(5198):693--695, 1995.

\bibitem{Vijayakumar05}
S.~Vijayakumar, A.~D'Souza, and S.~Schaal.
\newblock Incremental online learning in high dimensions.
\newblock {\em Neural Computation}, 17(12):2602--2634, 2005.

\bibitem{Hofsten04}
C.~von Hofsen.
\newblock An action perspective on motor an action perspective on motor
  development.
\newblock {\em TRENDS in Cognitive Science}, 8(6), 2004.

\bibitem{Vygotsky78}
L.S. Vygotsky.
\newblock {\em Mind and society: The development of higher mental processes}.
\newblock Cambridge, MA: Harvard University Press, 1978.

\bibitem{Weng04}
J.~Weng.
\newblock Developmental robotics: Theory and experiments.
\newblock {\em Int. J. Humanoid Robotics}, 1(2):199--236, 2004.

\bibitem{Weng01}
J.~Weng, J.~McClelland, A.~Pentland, O.~Sporns, I.~Stockman, M.~Sur, and
  E.~Thelen.
\newblock Autonomous mental development by robots and animals.
\newblock {\em Science}, 291(599-600), 2001.

\bibitem{White59}
R.~White.
\newblock Motivation reconsidered: The concept of competence.
\newblock {\em Psychol. Rev.}, 66:297--333, 1959.

\bibitem{Whitehead91}
S.~Whitehead.
\newblock A study of cooperative mechanisms for faster reinfocement learning.
\newblock Technical Report 365, Univ. Rochester, Rochester, NY, 1991.

\end{thebibliography}

\section{Biographies}

\begin{figure}[h!]
\center
\includegraphics[width=0.4\linewidth]{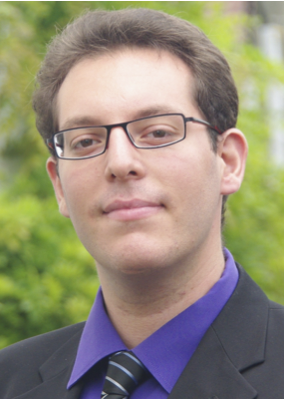}

\label{AdrienPhoto}
\end{figure}
\textbf{Adrien Baranes} received the M.S. degree in artificial intelligence and robotics from the University Paris VI, France, in 2008 and the a Ph.D. degree in artificial intelligence from the French National Institute of Computer Sciences and Control (INRIA)/University Bordeaux 1, France, in 2011. During his PhD, he studied developmental mechanisms allowing to constrain and drive the exploration process of robots in order to allow them to progressively learn high quantities of knowledge and know-how in unprepared open-ended spaces. Since January 2012, he has been studying intrinsic motivations with a biological/neurological point of view as a Post-Doctoral Fellow at Columbia University Medical Center, New-York, thanks to a Fulbright grant, and will pursue his research thanks to an HFSP Cross-Disciplinary Fellowship.

\begin{figure}[h!]
\center
\includegraphics[width=0.4\linewidth]{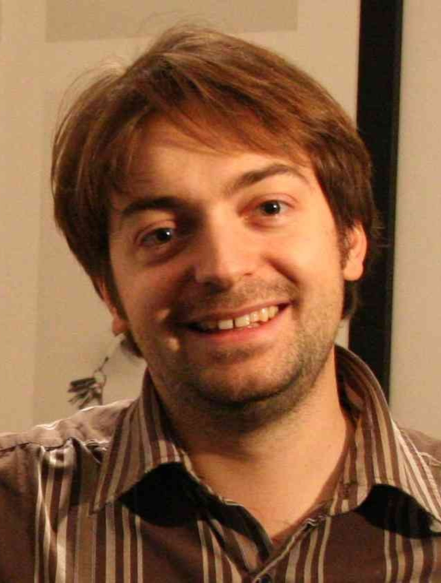}

\label{Pierre-YvesPhoto}
\end{figure}
\textbf{Dr. Pierre-Yves Oudeyer} is permanent researcher at Inria and responsible of the FLOWERS team at Inria and Ensta-ParisTech. Before, he has been a permanent researcher in Sony Computer Science Laboratory for 8 years (1999-2007). He studied theoretical computer science at Ecole Normale Supérieure in Lyon, and received his Ph.D. degree in artificial intelligence from the University Paris VI, France. After having worked on computational models of language evolution, he is now working on developmental  and social robotics. He has published a book, more than 80 papers in international journals and conferences, holds 8 patents, gave several invited keynote lectures in international conferences, and received several prizes for his work. In particular, he is a laureate of the ERC Starting Grant EXPLORERS. He is editor of the IEEE CIS Newsletter on Autonomous Mental Development, and an associate editor of IEEE Transactions on Autonomous Mental Development, Frontiers in Neurorobotics, and of the International Journal of Social Robotics.

\end{document}